\documentclass[11pt]{article} 

\usepackage{graphicx}
\usepackage{caption}
\usepackage{mathrsfs}
\usepackage{float}
\usepackage{subcaption}
\usepackage[utf8]{inputenc}
\usepackage{longtable}
\usepackage{booktabs}
\usepackage[english]{babel}
\usepackage[dvipsnames]{xcolor}
\usepackage{multirow}
\usepackage{ragged2e}
\usepackage{xltabular}
\usepackage{tcolorbox}
\usepackage{fancyvrb}
\PassOptionsToPackage{hyphens}{url} %
\usepackage[hypertexnames=false]{hyperref}
\definecolor{darkblue}{RGB}{0,0,120}
\hypersetup{
      colorlinks=true,
      linkcolor=darkblue,
      filecolor=darkblue,
      citecolor=darkblue,      
      urlcolor=darkblue,
}
\usepackage[justification=centering]{caption}

\usepackage[toc]{appendix}
\usepackage{adjustbox}
\usepackage{threeparttable, tablefootnote, stackengine}
\usepackage{array}
\usepackage{placeins}
\usepackage{indentfirst}
\usepackage[margin=1in]{geometry}
\usepackage{amsmath,amsthm,amssymb,scrextend}
\usepackage{cleveref}
\usepackage{bbm}
\usepackage{blindtext}
\usepackage{pdflscape}
\usepackage{vcell}
\usepackage{natbib}
 \bibpunct[, ]{(}{)}{;}{a}{}{,}%
\DeclareCaptionLabelFormat{andtable}{#1~#2  \&  \tablename~\thetable}
\usepackage{setspace}
\usepackage{lipsum}
\usepackage{footmisc}

 \usepackage{spverbatim}
\usepackage{titlesec}
\newcolumntype{x}[1]{%
>{\centering\arraybackslash\hspace{0pt}}p{#1}}%
\usepackage{titlecaps} %

\titleformat{\section}
  {\normalfont\fontsize{13}{13}\bfseries\centering} %
  {}{0em}{\titlecap} %
  [\vspace{0.5ex}]

\titleformat{\subsection}
  {\normalfont\fontsize{12}{12}\bfseries\itshape} %
  {}{0em}{\titlecap} %

\titleformat{\subsubsection}
  {\normalfont\fontsize{12}{12}\itshape} %
  {}{0em}{} %

\usepackage{nameref} %
\newcommand{\secref}[1]{%
  \hyperref[#1]{Section~\ref*{#1}: \nameref*{#1}}%
}

\usepackage{etoc}  
\usepackage{tabularx}
\usepackage{tabularray}
\usepackage{tikz}
\pgfdeclarelayer{bg}
\pgfsetlayers{bg,main}
\usetikzlibrary{shapes.geometric, arrows.meta, positioning, decorations.pathreplacing, calc}
\usetikzlibrary{shadows.blur, shapes.multipart, backgrounds}
\usetikzlibrary{positioning,calc,decorations.pathreplacing,arrows.meta,backgrounds}
\usepackage{dcolumn}
\usepackage{wrapfig}
\usepackage{verbatim}
\usepackage{makecell}
\usepackage[graphicx]{realboxes}
\usepackage[colorinlistoftodos,prependcaption,textsize=tiny]{todonotes}
\usepackage{url} %
\usepackage{soul}
\usepackage[T1]{fontenc}
\usepackage[utf8]{inputenc}
\usepackage{pdflscape}
\usepackage{verbatim}

\usepackage{booktabs}   %
\usepackage{siunitx}    %
\usepackage{soul}
\usepackage{tabularray} %
\UseTblrLibrary{booktabs} %

\usepackage{caption}    %
\usepackage{makecell}   %

\usepackage[affil-it]{authblk}
\title{AI-human hybrid conversations: Can AI Agents Improve Discourse Quality?}
\newif\ifblind
\blindfalse
\ifblind
\author{Author names hidden for anonymous review}
\else
\author[1]{Ishita Chakraborty\thanks{Author 1 and 2 are joint first authors and names appear alphabetically. The authors would like to thank their two talented undergraduate research assistants, Amogh Rao and Jiarui Gong,  majoring in Computer Science at the University of Wisconsin\textendash{}Madison. We also thank participants at the UW\textendash{}Madison Research Roundtable (Fall 2026), the NYU Stern WAICI Conference (2026) and the November 2026 T.R.A.P. Lab discussion meeting for their valuable feedback and suggestions. }}
\author[2]{Xingjian You\protect\footnotemark[1]}
\author[3]{Shrabastee Banerjee}
\affil[1,2]{University of Wisconsin-Madison}
\affil[3]{Tilburg University}
\fi
\date{}

\begin{document}

\begin{titlepage}
\singlespacing
\maketitle
\begin{abstract}
			\singlespacing
\noindent
Conversations are central to how people exchange information, share experiences, and develop collective understanding. Yet, across conversational settings, discussions often suffer from uneven participation, unanswered contributions, and a narrow range of perspectives. As generative AI agents increasingly participate alongside humans, hybrid human\textendash{}AI conversations present a new opportunity to address these challenges. Importantly, conversations are dynamic and path-dependent: contributions from both human and AI participants can influence who engages, how interactions unfold, and which perspectives emerge. We investigate whether AI agents can serve as catalysts for human interaction, improving conversational engagement and broadening collective discourse. We develop a framework for introducing generic and persona-conditioned AI agents into evolving discussions and evaluate it through two complementary studies. First, using agent-based simulations grounded in Reddit discussions across four domains, we find that AI-generated conversations are more positive and less toxic than human conversations, but substantially less semantically diverse. Persona conditioning produces more human-like contributions and narrows this diversity gap, while increasing the representation of underrepresented personas can shift which perspectives emerge without necessarily increasing overall diversity. Second, we conduct a pre-registered experiment with 221 participants across nine discussion threads, comparing human-only conversations with those involving generic or persona-conditioned AI agents. We find that AI participation increases responsiveness, reduces unanswered contributions, and broadens the range of ideas expressed by human participants, even though the AI agents themselves do not fully reproduce human conversational diversity. These findings suggest that the value of hybrid human\textendash{}AI conversations lies not simply in the content AI agents generate, but also in their potential to stimulate more responsive and diverse interactions among humans.
\end{abstract}

\vspace{0mm}\noindent

\clearpage
\end{titlepage}

\singlespacing

\section*{Introduction} \label{intro}

\noindent Conversations are central to how people exchange information, share experiences, and develop collective understanding. From focus groups used to elicit consumer insights in market research and classroom discussions that facilitate learning to online communities where individuals exchange opinions and seek advice, conversations provide opportunities for people to engage with ideas and perspectives beyond their own. Yet, across these settings, productive participation is difficult to sustain. Discussions often suffer from uneven engagement, with a few individuals dominating while others remain silent \citep{alimoradi2025rethinking}, contributions receiving limited responses, and group dynamics restricting the range of perspectives expressed \citep{farnsworth2010analysing, sedova2023vocal}. Infact, one of the top reasons for not participating actively in a discussion have been linked to \textit{feeling unwelcome}, \textit{not fitting in} and \textit{not getting enough response} from the community \citep{preece2004top}. These challenges are amplified on digital platforms, where discussions occur at scale and can be further characterized by echo chambers that reinforce existing viewpoints, polarized exchanges, and toxic interactions that undermine constructive dialogue and increase entry barriers for traditionally under-represented groups in a conversation \citep{li2008self,hu2017selfselection, chmiel2011negative,garcia2016dynamics,cheng2017troll,
lu2024silencing,sibai2024online}.

Crucially, conversations are not simply collections of independent contributions; they are dynamic, path-dependent processes in which each interaction shapes what comes next. An early response can influence who participates, the tone of subsequent exchanges, and the perspectives that enter the discussion \citep{muchnik2013social,cheng2017troll}. Consequently, even a single strategically introduced contribution has the potential to redirect the course of a conversation. This raises an important possibility: rather than merely observing how conversations unfold, can we actively intervene to improve their outcomes?

Recent advances in generative AI create a new opportunity to both study and potentially shape conversational trajectories. AI participation in online conversations is rapidly moving from a theoretical possibility to a practical reality. Platforms such as Character.AI and Meta's AI Studio have introduced AI characters that interact with humans, including in group conversations, while emerging AI-native communities such as Moltbook illustrate the possibility of discussions among autonomous AI agents. %
Unlike approaches that analyze conversations after they have occurred, AI agents can participate directly in evolving discussions, responding to overlooked contributions, introducing alternative perspectives, and potentially influencing subsequent interactions. Large language models (LLMs) can generate context-dependent natural language, condition their behavior on empirically derived persona characteristics, retain information about their own prior contributions, and respond sequentially to an evolving discussion. Prior work has established that LLMs can represent heterogeneous synthetic respondents \citep{argyle2023out,arora2025hybrids} and can also interact as agents in multi-turn social environments \citep{park2022social,park2023generative,zhou2024sotopia}. However, these streams have developed largely in parallel: persona-based work has focused primarily on whether individual synthetic actors reproduce characteristics of target populations, while multi-agent work has focused more on emergent interaction among generative agents. Less is known about how introducing AI agents into ongoing human conversations influences conversational outcomes, and whether the characteristics of these agents matter. In particular, generic AI agents may encourage engagement and civility while contributing a relatively narrow range of perspectives. Conditioning agents on empirically grounded, heterogeneous personas offers a potential way to introduce greater variation in communication styles and viewpoints, including perspectives associated with groups that are typically underrepresented in a discussion.

Building on these possibilities, our primary research questions are: (1) How does introducing AI participants affect conversational tone, human engagement, and opinion diversity? (2) Does the type of AI participant (generic versus persona conditioned), shape these conversational outcomes? We examine three complementary dimensions of discourse quality. \textit{Conversational tone} captures the sentiment and toxicity of exchanges; \textit{human engagement} reflects participation, responsiveness to contributions, and the continuation of discussions; and \textit{opinion diversity} captures the breadth of ideas, topics, and perspectives expressed. These dimensions are related but distinct: a conversation may generate substantial engagement while remaining toxic or dominated by a narrow set of viewpoints. Examining them jointly allows us to assess how AI participation affects not only the volume of interactions but also their quality and the range of perspectives represented.

To address these questions, we adopt a two-part empirical approach combining agent-based simulations with a human\textendash{}AI discussion experiment. Each approach serves a complementary purpose. Because conversations are path-dependent, observing a discussion reveals only one of many possible trajectories. Agent-based simulations provide a controlled testbed in which we can systematically vary the presence, characteristics, and composition of AI participants while holding the initial discussion context fixed. However, simulations alone cannot establish how real people respond to AI participation, particularly whether AI-generated contributions encourage subsequent human engagement or broaden the range of perspectives expressed by humans. We therefore complement the simulations with an experiment in which real participants engage in asynchronous discussions alongside AI agents. These two studies, together, allow us to examine both how AI agents generate conversations and how their participation shapes human conversational behavior.

In the first study, we develop an agent-based simulation framework grounded in actual Reddit discussions. Using the PANDORA dataset \citep{gjurkovic2021pandora}, we reconstruct conversations across four domains: Entertainment, Gaming, Technology \& Computing, and People \& Relationships and use demographic, psychographic, and linguistic information from observed community members to develop empirically grounded personas. We compare two AI conditions against actual human conversations: (1) Generic AI, in which agents participate without persona information; In the Persona AI condition, agents are uniformly sampled from the empirically grounded persona pool. We further examine the role of participant composition by increasing the sampling probability of personas associated with traditionally underrepresented groups. This additional intervention allows us to distinguish between the effects of grounding agents by heterogeneous human profiles and changing which profiles participate.  Across conditions, agents contribute sequentially to the same initial discussions, with each contribution becoming part of the context for subsequent interactions.

We find that AI-generated conversations are substantially more civil than human conversations, but also considerably more homogeneous. Across AI conditions, conversations exhibit more positive sentiment and substantially lower toxicity than observed human discussions. However, generic AI agents produce longer, more uniform contributions and a much narrower range of ideas. Introducing empirically grounded personas makes AI contributions more human-like in their writing patterns and broadens the semantic diversity of discussions: mean pairwise comment similarity declines from 0.541 under Generic AI to 0.458 under Persona AI. Nevertheless, this remains substantially above the human benchmark of 0.225, indicating that persona conditioning narrows but does not eliminate the diversity gap between AI-generated and human conversations.

Interestingly, increasing the sampling probability of underrepresented personas does not broadly change writing patterns or add semantic breadth, but it descriptively redirects attention towards concerns associated with some of those groups. In particular, oversampling female personas in Technology \& Computing brings greater attention to topics like the everyday consequences of technology, including usability, invisible maintenance work, and caregiving (concerns also present among female commentors but not endorsed by the group). Thus, persona grounding can recover some of the heterogeneity lost under generic generation, while changing participant composition can alter which perspectives become salient without necessarily making the conversation more diverse overall. To assess whether these properties are specific to our model, we also conduct supplementary analyses on Moltbook, an independently evolving platform on which all posts and comments are written by heterogeneous AI agents.\footnote{\url{https://www.moltbook.com/}} We find that Moltbook discussions are even more positive and less toxic than our persona simulation but broader than it, though still narrower than human Reddit. This suggests that the tone gap is a general property of AI-generated discourse, while introducing model variety narrows but does not close the gap in the breadth of ideas discussed.

However, the value of AI participation may not lie solely in the content AI generates, but also in the subsequent human interactions it stimulates. We investigate this possibility in our second study, developing an interactive discussion platform and conducting a four-day experiment involving 221 students across nine asynchronous discussion threads. Participants engage in human-only discussions or discussions augmented with Generic AI or Persona AI agents. We find that AI participation makes discussions substantially more responsive. The proportion of participants receiving no reply falls from 63\% in human-only discussions to 27\% with Generic AI and 21\% with Persona AI. Importantly, AI participation also broadens the range of topics discussed by humans themselves. Mean semantic similarity among human contributions declines from 0.40 without AI to 0.30\textendash{}0.31 in AI-assisted discussions, indicating a broader range of ideas. While both AI conditions improve responsiveness and broaden human discussion, the agents exhibit different interaction styles: Generic AI more frequently asks questions, whereas Persona AI more often shares personal experiences and engages in narrative exchanges. Thus, AI participation can change not only whether human contributions receive responses, but also the nature and diversity of subsequent human interactions.

Our findings highlight both the potential and the limitations of AI participation in conversations. AI agents generate more positive and less toxic contributions than humans, but their discussions remain substantially more homogeneous. Persona conditioning narrows this diversity gap, while altering the representation of different personas can change which perspectives become salient. Importantly, our human\textendash{}AI experiment shows that introducing AI agents can make discussions more responsive and broaden the range of ideas expressed by human participants, even when the AI agents themselves do not fully reproduce human conversational diversity.

\section*{Related Work}
\label{litreview}

\subsection*{Conversational Dynamics and Opportunities for Intervention}

\noindent
A substantial literature has documented challenges in sustaining productive conversations across settings, including focus groups, classroom discussions, and online communities. First, participation is often uneven, with some individuals dominating discussions while others contribute little or remain silent \citep{farnsworth2010analysing,alimoradi2025rethinking,sedova2023vocal}. Decisions to participate are shaped not only by individual preferences but also by the conversational environment. Individuals may refrain from contributing because they feel unwelcome, do not identify with the community, or anticipate limited responses to their contributions \citep{preece2004top}. In online communities, participation also depends on prior contributions, community characteristics, and platform design \citep{goes2014popularity,yang2019ugc,ren2012building,huang2019motivating}. Consequently, observed discussions may disproportionately reflect the experiences and perspectives of more active participants, while other voices remain underrepresented \citep{le2018endogenous,li2008self,hu2009overcoming,chakraborty2022attribute}. Second, the quality of interactions can deteriorate as discussions progress. Incivility can discourage participation \citep{gervais2015incivility,lu2024silencing}, while negative emotions and trolling can propagate across subsequent exchanges, reinforcing unconstructive conversational dynamics \citep{chmiel2011negative,garcia2016dynamics,cheng2017troll,zhang2018awry}. Platforms therefore
face not only a participation problem but also an interaction problem, and
often rely on content moderation to shape what remains visible and restore
coherence in fast-moving discussions \citep{he2025platform,zhao2025lost}.

Existing research has examined participation biases and interventions to improve discourse, but less attention has been devoted to how interventions can alter the trajectory of an evolving conversation. Unlike independent observations, conversational contributions are interdependent: each response can introduce new perspectives, redirect attention, or influence subsequent participation. Consequently, changing who participates or what they contribute may alter not only the content observed at a given point, but also the interactions that follow. While statistical corrections such as weighting or imputation can address some forms of participation bias, they cannot recover these unrealized conversational trajectories. Our work examines this gap by investigating how strategically introduced AI participants can shape the subsequent evolution of discussions.

\subsection*{AI Participants and the Dynamics of Human Conversations}

\noindent
AI participants have the potential to influence how conversations unfold, particularly because early interactions can shape subsequent participation and discussion trajectories. New communities often struggle to attract contributors until there is already content to engage with \citep{kraut2012building}, while early contributions can influence subsequent evaluations \citep{muchnik2013social} and the propagation of trolling behavior \citep{cheng2017troll}. AI agents could help address these challenges by responding to overlooked contributions, sustaining interactions, and introducing alternative perspectives. However, AI participation also presents risks: people may place less trust in content perceived to be AI-generated \citep{jakesch2019ai,longoni2022artificial}, and disclosing AI involvement can itself reduce trust \citep{schilke2025transparency}. While prior research has examined how people evaluate individual pieces of AI-generated content, less is known about how AI participation affects the dynamics and collective outcomes of an evolving conversation.

Agent-based simulation provides a useful approach for investigating such dynamics, allowing researchers to examine how aggregate outcomes emerge from repeated interactions among heterogeneous actors \citep{smith2018simulating}. Prior information systems research has applied these models to dynamic processes such as the evolution of digital architectures and feedback between user behavior and recommender systems \citep{haki2020evolution,zhang2020consumption}. By holding the initial environment fixed and varying agents' characteristics or behavioral rules, simulations can reveal how individual interventions propagate through a system. However, conventional agent-based models typically specify a restricted set of actions, whereas conversational agents must generate open-ended, context-dependent language that responds to prior contributions and evolving discussion histories. Generative AI makes it possible to extend agent-based simulations to these settings, allowing researchers to study how the presence and characteristics of AI participants shape conversational trajectories. Our research builds on these literatures by examining both AI-generated discussions under controlled simulation conditions and the responses of real human participants when AI agents join their conversations.

\subsection*{Generative Agents, Personas, and the Fidelity of Synthetic Behavior}

\noindent Large language models expand the behavioral space available to researchers building synthetic consumers or digital twins, based on which informative simulations may be conducted. One stream of research examines whether LLMs can reproduce responses from heterogeneous human populations. Conditioning models on demographic characteristics can generate systematic variation in simulated responses \citep{argyle2023out}, while related work shows that synthetic respondents can recover some human behavioral regularities but also depart systematically from them \citep{aher2023using,arora2025hybrids}. A second stream moves from isolated respondents to interacting generative agents that maintain information about prior interactions and generate sequential contributions within a shared social environment \citep{park2022social,park2023generative}.

These developments also raise the question of how agents should be grounded. Generic agents represent no particular actor; persona-conditioned agents represent types of actors based on selected characteristics; and human digital twins make the stronger claim of representing particular individuals \citep{park2024agents,toubia2025twin}. More detailed grounding, however, does not automatically imply greater validity. An agent may reproduce linguistic style without substantive preferences, marginal response distributions without interactive behavior, or plausible responses without reproducing how humans react to interventions \citep{gui2023challenge,peng2025funhouse,ashokkumar2026experiments}. The relevant notion of fidelity, therefore, depends on the research question and outcome being simulated.

Existing work has established both heterogeneous synthetic respondents and interacting generative agents, but largely as separate capabilities. Less is known about how empirically grounded heterogeneous personas behave when embedded in sequential multi-agent conversations, or whether persona conditioning improves conversational fidelity relative to generic LLM agents. Our work connects these streams by grounding agents in observed communication styles, allowing them to interact sequentially, and benchmarking the resulting conversational trajectories against both generic AI and observed human discussions.

\section*{Empirical Setting} \label{data}

\noindent We evaluate our framework through a two-step approach that combines agent-based simulations with a human\textendash{}AI discussion experiment. In the first step, we develop a simulation testbed grounded in real Reddit discussions, allowing us to systematically introduce AI participants and vary their characteristics and composition while holding the initial conversational context fixed. This approach enables us to examine how different AI agents shape conversational tone and opinion diversity as discussions evolve, without directly manipulating interactions in existing online communities, which can raise ethical and operational concerns. However, simulations alone cannot establish whether AI participation encourages real people to engage more actively or changes the perspectives they express. In the second step, we therefore bring human participants into the testbed through an interactive discussion platform, where we experimentally compare human-only conversations with hybrid discussions involving generic or persona-conditioned AI agents. Together, the two studies allow us to examine both the dynamics of AI-generated conversations and the consequences of introducing AI participants into human discussions.

\subsection*{Data}
\noindent We make use of Reddit records from the PANDORA dataset \citep{gjurkovic2021pandora}, which contains activity from more than 10,000 users linked to demographic and psychographic attributes, including gender, age, location, and Myers-Briggs Type Indicator (MBTI) personality traits (e.g., introverted vs. extraverted; \citealp{myers1985manual}).\footnote{Demographic and personality attributes in PANDORA are self-declared through subreddit flairs or comments, so they are available only for subsets of the 10,295 users (MBTI for about 9,000, gender for about 3,200, and age for about 2,300). For gender, we use PANDORA's text-based prediction, which covers every user; we therefore refer to female- and male-attributed users. For age (over 35 versus 35 and under) and introversion, we use self-declared labels, and compute participation shares over comments whose authors declare the attribute.} We focus on four domains central to our study: Entertainment \& Pop Culture, Gaming \& Esports, People \& Relationships, and Technology \& Computing. The resulting participation corpus contains 5,003,567 comments posted by 10,102 unique authors. Because these data pre-date the large-scale adoption of chat-based LLM tools, in particular ChatGPT (November 2022), we are more confident that our human benchmarks are not contaminated with AI-generated content. We use this corpus to construct personas and to identify participation gaps by domain, as described below.

While PANDORA is well suited for studying participation patterns across groups, it is organized at the user level and therefore does not preserve complete discussion trajectories, including the full sequence and reply structure of comments within a thread. To recover this conversational context, we randomly sample 600 posts from the participation corpus and re-collect their complete discussion threads through the public Reddit API (details in Table W1 of Web Appendix A). The reconstructed threads preserve comment order, reply structure, and conversational context, allowing us to examine how discussions evolve over time (presented in Figure W1 in the Web Appendix A). 

Out of these 600 threads, for computational tractability, our main simulations focus on a subsample of 120 threads (30 per domain) containing 48,315 human comments in total. To ensure that every thread can supply up to 25 human comments of context before AI agents enter, we restrict the sampling frame to threads with at least 50 human comments and draw 30 threads at random within each domain.\footnote{Selection is deterministic given the seed recorded in the replication files. Because the eligibility rule favors longer threads, the 120 threads are not a random sample of the 600-thread corpus and hold roughly 45\% of its comments; the descriptive human trajectories in Figure W1 of Web Appendix A use the full 600-thread corpus.} The same 120 threads anchor the human reference for all three AI conditions, holding post-level context constant across comparisons. Our analyses of tone and style use the full human threads, i.e., all 48,315 comments\footnote{Table W5 in Web Appendix A shows that the tone and style results are robust to restricting the human benchmark to the first 15 comments.}, and the semantic breadth analysis uses the first 15 human comments of each thread.%

\subsection*{Persona Construction}

\noindent A central challenge in our simulation is constructing AI personas that reflect the heterogeneity observed in real online communities. Given our dataset has access to demographic information (age and gender) of those who post, the most straightforward approach is to define personas using demographic attributes alone. However, demographics explain only a limited share of variation in consumer preferences, attitudes, and behavior and may therefore generate relatively little heterogeneity in the resulting text \citep{arora2025hybrids}. They may also encourage stereotype-based generation: a model instructed only to speak as, for example, a woman in her forties has few behavioral cues to draw on beyond its learned priors about that demographic category. Previous work (prior to large scale LLM access) addresses this limitation by deriving personas from aggregated behavioral data. For example, \citet{jung2017persona} and \citet{an2018imaginary} identify latent segments from social-media interactions and translate them into human-readable personas using representative demographics and behavioral interests. This approach is well suited to settings in which personas must capture what kinds of content different segments consume or engage with. Our setting imposes an additional requirement; our personas must \emph{write}: they enter an evolving discussion and produce comments to which other participants may respond. A persona defined primarily by topical interests offers little guidance about linguistic voice, conversational style, or mode of expression, leaving the language model to revert to its default writing style. We therefore require personas that capture not only what users are interested in, but also how they communicate.

We explored several alternative approaches to persona construction. First, following prior work, we experimented with personas derived from participation patterns across topics and subreddits. Second, we considered clustering users jointly on demographic attributes, MBTI personality, and linguistic features. Third, we evaluated prompts that explicitly incorporated large numbers of demographic, psychographic, and linguistic characteristics. These exploratory analyses revealed two limitations. Participation-based personas captured what users tended to discuss but produced relatively weak distinctions in conversational style. Likewise, when demographic variables such as age and gender were included directly in the clustering procedure, the resulting clusters were largely separated by demographic composition rather than by differences in communication. For instance, in Gaming \& Esports, one cluster was defined almost entirely by gilding, comment volume, and score; in Entertainment \& Pop Culture, a cluster separated entirely based on age and engagement rather than language. Moreover, because the underrepresented-AI intervention
manipulates the demographic composition of the persona pool, style clusters must
vary independently of demographics, so that any topical shift observed under oversampling can be attributed to representation rather than to style.
Similarly, MBTI variables contributed little additional separation once linguistic behavior was taken into account.\footnote{Personality type did not improve cluster separation and is not used to construct the persona pool. It is retained only for analysing the underrepresented AI intervention: specifically in the People \& Relationships domain, where agents with an Introvert flag are upsampled.} As a result, demographic and personality variables were removed from the clustering procedure. %

Based on these observations, we deliberately separate how people communicate from who they are in persona construction. Within each domain, we identify recurring writing styles using 26 user-level mean LIWC measures and log-transformed comment count. These 27 inputs are standardized before clustering; demographic and personality variables are excluded and added separately when constructing the final persona pool. Figure 1(a) traces this two-stage process. The LIWC measures \citep{tausczik2010liwc} capture multiple dimensions of communication, including comment and sentence length, analytical thinking, authenticity, first-person language, social orientation, netspeak, profanity, punctuation, and numerical language. %

With this approach, we obtain five writing-style clusters for Technology \& Computing and four clusters each for Entertainment \& Pop Culture, Gaming \& Esports, and People \& Relationships, yielding 17 empirically derived writing-style clusters in total. \autoref{tab:persona_clusters} summarizes the dominant characteristics of each cluster. Across domains, the clusters capture distinct conversational voices rather than merely differences in topic preference. For example, Technology \& Computing includes long-form technical explainers, code- and log-oriented fragments, first-person troubleshooting narratives, social discussion, and highly informal netspeak exchanges. People \& Relationships separates reflective self-narratives, emotional venting, structured advice, and high-activity debate. Gaming \& Esports distinguishes guide-style explanations, community discussion, first-person gameplay experiences, and shorthand meme-like reactions, while Entertainment \& Pop Culture contains formal critiques, identity-driven personal reactions, mainstream discussion, and internet-style fragments.

Finally, each writing-style cluster is subsequently crossed with two age groups and two gender groups, producing 68 personas (17 × 2 × 2). This final step allows us to manipulate demographic representation while holding conversational style fixed. Consequently, any changes in discussion dynamics under the oversampling treatments can be attributed to changes in participant composition rather than mechanically induced differences in writing style. Each persona thus consists of four components: a) demographic attributes (age and gender); b) the cluster-specific linguistic profile summarized from LIWC statistics; c) natural-language description of the writing style generated from the cluster characteristics; and d) five representative comments sampled from real Reddit users belonging to that cluster, which serve as stylistic anchors during generation. During simulation, each persona receives the current discussion thread together with its own previous contributions, allowing responses to remain both contextually appropriate and internally consistent across multiple turns. The detailed prompt is included in the GenAI Web Appendix (subsection ``Prompting Strategies'').

For our simulations involving the oversampling of underrepresented personas, we use the main dataset to identify groups that appear less often in each of the four domains. Shares are calculated over comments linked to author profiles, so the percentages describe comments rather than unique authors. In \autoref{tab:motivating_bias} in Web Appendix A, we find that Female-attributed users account for 4.29\% of comments in Gaming \& Esports and 3.00\% in Technology \& Computing, compared with 22.38\% across the four domains. Users over age 35 account for 5.90\% of comments overall and are less common in several domains. In People \& Relationships, introvert- and thinking-oriented users take part less often than more expressive users. The groups also place different weight on parts of the discussion. These patterns motivate a controlled comparison in which the starting post stays fixed while the mix of agents changes.  

\definecolor{cBlue}{RGB}{31,90,140}
\definecolor{cBlueF}{RGB}{226,238,248}
\definecolor{cGray}{RGB}{115,115,115}
\definecolor{cGrayF}{RGB}{242,242,242}
\definecolor{cMag}{RGB}{160,60,120}
\definecolor{cMagF}{RGB}{250,232,242}
\definecolor{k1}{RGB}{49,110,170}
\definecolor{k2}{RGB}{40,140,120}
\definecolor{k3}{RGB}{205,120,45}
\definecolor{k4}{RGB}{135,85,160}
\providecommand{\cfmain}{\fontsize{9}{10.8}\selectfont}
\providecommand{\cfsmall}{\fontsize{7.4}{8.8}\selectfont}
\providecommand{\cftiny}{\fontsize{6.2}{7.4}\selectfont}

\begin{figure}[p]
\centering
\hyphenpenalty=10000 \exhyphenpenalty=10000 \tolerance=1000
  \begin{tikzpicture}[
    scale=0.95, transform shape,   %
    font=\cfmain,
    every node/.style={align=center},
    box/.style={draw=cBlue, fill=white, rounded corners=2pt, line width=0.5pt, align=center},
    ar/.style={-{Latex[length=2.4mm,width=1.8mm]}, line width=0.6pt, cBlue!85}
  ]

  \node[anchor=west, font=\cfmain\bfseries, text=cBlue] at (-8.0,0.35)
    {(a)~Constructing one persona from observed users};

  \node[box, minimum width=3.0cm, minimum height=3.2cm] (f1) at (-6.3,-2.35) {};
  \node[anchor=south, font=\cfsmall\bfseries] at (-6.3,-0.70) {Real users in a domain};
  \foreach \dx/\dy in {-0.80/0.52, -0.62/0.72, -0.90/0.74, -0.70/0.36, -0.52/0.56,
                       -0.86/0.28, -0.60/0.18, -0.98/0.56, -0.74/0.64}
    \fill[k1] ($(-6.3,-2.35)+(\dx,\dy)$) circle (1.5pt);
  \foreach \dx/\dy in {0.58/0.62, 0.76/0.78, 0.50/0.84, 0.68/0.42, 0.86/0.58,
                       0.44/0.46, 0.72/0.66, 0.60/0.28, 0.90/0.74}
    \fill[k3] ($(-6.3,-2.35)+(\dx,\dy)$) circle (1.5pt);
  \foreach \dx/\dy in {-0.74/-0.60, -0.56/-0.44, -0.88/-0.40, -0.66/-0.78, -0.48/-0.68,
                       -0.92/-0.74, -0.60/-0.28, -0.80/-0.24, -0.52/-0.86}
    \fill[k2] ($(-6.3,-2.35)+(\dx,\dy)$) circle (1.5pt);
  \foreach \dx/\dy in {0.62/-0.56, 0.80/-0.42, 0.50/-0.72, 0.72/-0.80, 0.90/-0.62,
                       0.56/-0.34, 0.84/-0.28, 0.66/-0.22, 0.44/-0.54}
    \fill[k4] ($(-6.3,-2.35)+(\dx,\dy)$) circle (1.5pt);
  \draw[dashed, k3, line width=0.6pt] ($(-6.3,-2.35)+(0.67,0.58)$) circle (0.46);
  \node[anchor=north, font=\cftiny, text=cGray, text width=3.1cm] at (-6.3,-4.05)
    {};

  \node[box, minimum width=4.2cm, minimum height=3.2cm] (f2) at (-2.0,-2.35) {};
  \node[anchor=south, font=\cfsmall\bfseries] at (-2.0,-0.70) {Linguistic Clustering};
  \node[anchor=north, font=\cftiny\itshape, text=k3] at (-2.0,-0.92)
    {``Technical explainer''};
  \foreach \i/\v/\lab in {0/0.76/WC, 1/0.86/Analytic, 2/0.22/{I-words}, 3/0.34/Social, 4/0.62/WPS}{
    \node[anchor=east, font=\cftiny, text=cGray] at ($(-3.10,-1.50)+(0,-\i*0.20)$) {\lab};
    \draw[cGrayF, line width=2.4pt]
      ($(-3.04,-1.50)+(0,-\i*0.20)$) -- ($(-3.04,-1.50)+(1.55,-\i*0.20)$);
    \draw[k3, line width=2.4pt]
      ($(-3.04,-1.50)+(0,-\i*0.20)$) -- ($(-3.04,-1.50)+(\v*1.55,-\i*0.20)$);
  }
  \node[anchor=north, font=\cftiny\itshape, text=cGray, text width=3.8cm] at (-2.0,-2.70)
    {``just reseat the cooler, it's fine''\\ ``depends on the VRM, honestly''};
  \node[anchor=north, font=\cftiny, text=cGray, text width=4.3cm] at (-2.0,-4.05)
    {};

  \node[box, minimum width=3.0cm, minimum height=3.2cm] (f3) at (2.2,-2.35) {};
  \node[anchor=south, font=\cfsmall\bfseries] at (2.2,-0.70) {Demographic variants};
  \foreach \cx/\cy/\t/\hl in {-0.63/0.44/{male\\$<$35}/0, 0.63/0.44/{male\\35+}/0,
                              -0.63/-0.44/{female\\$<$35}/0, 0.63/-0.44/{female\\35+}/1}{
    \ifnum\hl=1
      \node[draw=cMag, fill=cMagF, line width=1.0pt, rounded corners=2pt,
            minimum width=1.18cm, minimum height=0.70cm, font=\cftiny, text=cMag]
        at ($(2.2,-2.35)+(\cx,\cy)$) {\t};
    \else
      \node[draw=cGray!55, fill=cGrayF, rounded corners=2pt,
            minimum width=1.18cm, minimum height=0.70cm, font=\cftiny, text=cGray]
        at ($(2.2,-2.35)+(\cx,\cy)$) {\t};
    \fi
  }
  \node[anchor=north, font=\cftiny, text=cGray, text width=3.2cm] at (2.2,-4.05)
    {gender $\times$ age group\\ ($+$ MBTI overlay in P\&R)};

  \node[box, minimum width=4.0cm, minimum height=3.2cm] (f4) at (6.2,-2.35) {};
  \node[anchor=south, font=\cfsmall\bfseries] at (6.2,-0.70) {Prompt Construction};
  \begin{scope}
    \clip (4.75,-1.15) circle (0.23);
    \fill[cMagF] (4.75,-1.15) circle (0.23);
    \fill[cMag!75] (4.75,-1.10) circle (0.078);
    \fill[cMag!75] (4.75,-1.40) ellipse (0.15 and 0.13);
  \end{scope}
  \draw[cMag!60, line width=0.4pt] (4.75,-1.15) circle (0.23);
  \node[anchor=west, font=\cfsmall\bfseries, text=cMag] at (5.05,-1.15) {\texttt{u/actor\_07}};
  \node[anchor=north west, font=\cftiny, text=cGray, align=left, text width=3.7cm]
    at (4.40,-1.52)
    {female~~$\cdot$~~35+~~$\cdot$~~cluster 3};
  \node[draw=cGray!50, fill=cGrayF, rounded corners=1.5pt, anchor=north,
        text width=3.45cm, align=left, font=\cftiny\ttfamily, inner sep=3pt]
    at (6.2,-1.90)
    {You are roleplaying as a\\
     Reddit user named \{name\}.\\
     PROFILE: \{gender\},\{age\}\\
     STYLE: \{style\},\{liwc\}\\
     EXAMPLES: \{rep\_comments\}};
  \node[anchor=north, font=\cftiny, text=cGray, text width=4.1cm] at (6.2,-4.05)
    {the persona becomes\\ the agent's system prompt};

  \draw[ar] (f1.east) -- (f2.west);
  \draw[ar] (f2.east) --
    node[midway, fill=white, inner sep=1pt, font=\cfmain, text=cBlue]{$\times$} (f3.west);
  \draw[ar] (f3.east) --
    node[midway, fill=white, inner sep=1pt, font=\cfmain, text=cBlue]{$=$} (f4.west);

  \node[anchor=west, font=\cfmain\bfseries, text=cBlue] at (-8.0,-4.85)
    {(b)~The persona pool and the actor draw};

  \node[box, minimum width=6.2cm, minimum height=4.30cm] (pool) at (-4.7,-7.75) {};
  \node[anchor=south, font=\cfsmall\bfseries] at (-4.7,-5.55)
    {Persona pool \textemdash{} Gaming \& Esports};
  \foreach \cx/\t in {-1.17/{M\\$<$35}, -0.39/{M\\35+}, 0.39/{F\\$<$35}, 1.17/{F\\35+}}
    \node[font=\cftiny, text=cGray] at ($(-4.7,-6.15)+(\cx,0)$) {\t};
  \foreach \ry/\r in {0/1, 1/2, 2/3, 3/4, 4/5}{
    \node[anchor=east, font=\cftiny, text=cGray]
      at ($(-4.7,-6.75)+(-1.62,-\ry*0.48)$) {C\r};
    \foreach \cx/\fg in {-1.17/0, -0.39/0, 0.39/1, 1.17/1}{
      \ifnum\fg=1
        \node[draw=cMag!55, fill=cMagF, rounded corners=1.5pt,
              minimum width=0.66cm, minimum height=0.34cm]
          at ($(-4.7,-6.75)+(\cx,-\ry*0.48)$) {};
      \else
        \node[draw=cBlue!55, fill=cBlueF, rounded corners=1.5pt,
              minimum width=0.66cm, minimum height=0.34cm]
          at ($(-4.7,-6.75)+(\cx,-\ry*0.48)$) {};
      \fi
    }
  }
  \node[anchor=north, font=\cftiny, text=cGray, text width=7.4cm] at (-4.7,-10.05)
    {every style cluster is crossed with all four demographic\\
     variants, so the \emph{pool} is balanced by construction};

  \node[box, minimum width=6.9cm, minimum height=1.25cm, anchor=north west] (d1) at (0.9,-5.60) {};
  \node[anchor=north west, font=\cfsmall\bfseries, align=left] at (1.05,-5.72) {Generic AI};
  \node[anchor=north west, font=\cftiny, text=cGray, align=left, text width=4.6cm]
    at (1.05,-6.10) {no persona attached to the actor};
  \node[draw=cGray!60, dashed, fill=white, rounded corners=2pt,
        minimum width=0.85cm, minimum height=0.62cm, font=\cftiny, text=cGray]
    at (6.95,-6.05) {?};

  \node[box, minimum width=6.9cm, minimum height=1.25cm, anchor=north west] (d2) at (0.9,-7.10) {};
  \node[anchor=north west, font=\cfsmall\bfseries, align=left] at (1.05,-7.22) {Persona AI};
  \node[anchor=north west, font=\cftiny, text=cGray, align=left, text width=4.6cm]
    at (1.05,-7.60) {uniform draw: every tile equally likely};
  \node[anchor=north, font=\cftiny, text=cGray] at (6.95,-7.24) {25\% each};
  \foreach \i/\c in {0/cBlue, 1/cBlue, 2/cMag, 3/cMag}
    \draw[\c!70, line width=3.4pt]
      ($(6.35,-8.16)+(\i*0.40,0)$) -- ($(6.35,-8.16)+(\i*0.40,0.44)$);

  \node[box, minimum width=6.9cm, minimum height=1.45cm, anchor=north west] (d3) at (0.9,-8.60) {};
  \node[anchor=north west, font=\cfsmall\bfseries, align=left] at (1.05,-8.72)
    {Underrepresented AI};
  \node[anchor=north west, font=\cftiny, text=cGray, align=left, text width=4.8cm]
    at (1.05,-9.10) {weighted draw: female personas selected\\ with 80\% probability at each actor draw};
  \node[anchor=north, font=\cftiny, text=cGray] at (6.95,-8.74) {80\% female};
  \foreach \i/\c/\h in {0/cBlue/0.11, 1/cBlue/0.11, 2/cMag/0.44, 3/cMag/0.44}
    \draw[\c!70, line width=3.4pt]
      ($(6.35,-9.86)+(\i*0.40,0)$) -- ($(6.35,-9.86)+(\i*0.40,\h)$);

  \draw[ar] (pool.east) -- ++(0.28,0) |- (d1.west);
  \draw[ar] (pool.east) -- ++(0.28,0) |- (d2.west);
  \draw[ar] (pool.east) -- ++(0.28,0) |- (d3.west);

  \node[anchor=north west, font=\cftiny, text=cGray, text width=15.9cm, align=left]
    at (-8.0,-11.05)
    {Only the draw probability changes across conditions; the pool itself is held fixed.
     People \& Relationships is implemented differently: an Introvert MBTI overlay is applied to 80\% of pool
     variants during construction, after which actors are drawn uniformly. For scale, female-attributed comments
     are 4.29\% of Gaming \& Esports and 3.00\% of Technology \& Computing in the observed corpus.};

  \end{tikzpicture}
\caption{Persona Construction and the Actor Draw}
\label{fig:persona_methods}

{\footnotesize \textit{Notes.} Panel (a) traces the construction of a single persona. Users within a
domain are clustered on LIWC writing-style features; each cluster is summarised by a style profile
consisting of a LIWC signature and representative comments from real users; that profile is crossed with
demographic variants; and the resulting persona is instantiated as the agent's system prompt. Panel (b)
shows the pool this produces and how actors are drawn from it under each condition. Because every style
cluster is crossed with all four demographic variants, the pool is balanced by construction, and the three
conditions differ only in the probability with which tiles are drawn. Illustrative values are shown for Gaming \& Esports; cluster labels, LIWC values, and example comments are stylised for exposition.}

\end{figure}

\begin{table}[!htbp]
\centering
\caption{Writing-Style Groups Built from Reddit Data}
\label{tab:persona_clusters}
{\scriptsize
\setlength{\tabcolsep}{3pt}
\renewcommand{\arraystretch}{1.02}
\begin{tabularx}{\textwidth}{
>{\raggedright\arraybackslash}p{2.45cm}
>{\raggedright\arraybackslash}p{2.95cm}
>{\raggedright\arraybackslash}p{4.0cm}
>{\raggedright\arraybackslash}X}
\toprule
\textbf{Domain} &
\textbf{Writing-style group} &
\textbf{Main LIWC signals} &
\textbf{Typical writing} \\
\midrule

\multirow{5}{2.45cm}{Technology \& Computing}
& Hyper-netspeak & $\uparrow$ netspeak; $\downarrow$ WC, WPS & Very short, meme-like replies \\
& Code/log fragments & $\uparrow$ numbers, punctuation, Analytic & Settings, logs, and code snippets \\
& First-person troubleshooting & $\uparrow$ I, Authentic; $\downarrow$ Analytic & Personal problem-solving stories \\
& Social discussion & $\uparrow$ Social, affiliation & Advice and opinions \\
& Technical explainer & $\uparrow$ Analytic, WC, WPS & Long, structured explanations \\

\midrule

\multirow{4}{2.45cm}{People \& Relationships}
& Reflective self-narrative & $\uparrow$ WC, WPS, Authentic & Personal stories \\
& Social debate & $\uparrow$ Social, netspeak & Fast debates and discussion \\
& Emotional disclosure & $\uparrow$ I, Authentic & Emotional venting \\
& Rational advice & $\uparrow$ Analytic; $\downarrow$ pronouns & Logical guidance \\

\midrule

\multirow{4}{2.45cm}{Gaming \& Esports}
& Guide / analysis & $\uparrow$ Analytic, WC, WPS & Gameplay explanations \\
& Community talk & $\uparrow$ Social, pronouns & Player-to-player interaction \\
& Meme / stat fragments & $\uparrow$ numbers, profanity & Quick reactions \\
& Personal experience & $\uparrow$ I, Authentic & Gameplay reflections \\

\midrule

\multirow{4}{2.45cm}{Entertainment \& Pop Culture}
& Mainstream discussion & Moderate social language & General discussion \\
& Personal reaction & $\uparrow$ I, Social; $\downarrow$ Analytic & Personal reactions linked to identity \\
& Formal critique & $\uparrow$ Analytic, WC, WPS & Essay-style critiques \\
& Numeric / meme fragments & $\uparrow$ numbers, netspeak & Meme-style reactions \\

\bottomrule
\end{tabularx}
}

\vspace{0.3em}
\begin{minipage}{0.95\textwidth}
\scriptsize
\textit{Notes.} Users are grouped within each domain using 26 user-level mean LIWC measures and log-transformed comment count, yielding 27 standardized inputs. Age, gender, and personality are not used to form the groups. Arrows show the measures that set one group apart from the others in the same domain. The final set has 17 writing-style groups: 5 in Technology \& Computing and 4 in each other domain. We cross these groups with two age groups and two gender groups to build 68 personas. WC means word count, and WPS means words per sentence.
\end{minipage}
\end{table}

\subsection*{Matched Conversation Simulation}

\noindent The simulation uses the matched set of 120 Reddit posts described in the Data section, with 30 posts in each of four domains. We generate three continuations of each post. Generic AI receives no persona information. Persona AI samples evenly from the domain-specific persona pool. Underrepresented AI uses the same personas but increases the chance that a group with low observed participation enters the discussion. \autoref{tab:scenarios} summarizes the three conditions. Figure 1(b) shows the resulting persona pool and how the three conditions differ only in the probability with which personas are drawn from it.

\begin{table}[ht]
\centering
\scriptsize
\caption{Simulation Conditions}
\label{tab:scenarios}
\begin{tabularx}{\textwidth}{
>{\raggedright\arraybackslash}p{2.7cm}
>{\raggedright\arraybackslash}p{3.9cm}
>{\raggedright\arraybackslash}p{3.8cm}
>{\raggedright\arraybackslash}X}
\toprule
\textbf{Condition} &
\textbf{Agent information} &
\textbf{Sampling rule} &
\textbf{Purpose} \\
\midrule

Generic AI &
No demographic or writing-style profile &
Use the same generic agent prompt in every round &
Provides a baseline without persona information \\

Persona AI &
Domain-specific persona built from Reddit data &
Sample evenly from the domain persona pool &
Measures the effect of persona grounding \\

Underrepresented AI &
Same personas as Persona AI &
Sample the target group more often &
Measures the effect of changing the participant mix \\

\bottomrule
\end{tabularx}

\vspace{0.4em}
\begin{minipage}{0.95\textwidth}
\scriptsize
\textit{Notes.} All conditions use the same posts, human-context seed comments, language model, generation settings, user prompt, and generation order. They differ only in the agent information and the rule used to choose agents.
\end{minipage}
\end{table}

We also vary how much human discussion the agents see before they enter. The four human-context seed levels are 0, 5, 15, and 25 comments. At seed 0, the agents see only the post. At seed 25, they enter a thread with a longer human exchange. Every condition sees the full post, including its title, body, and subreddit. The same post and seed comments are used across conditions.

All comments are generated with GPT-5.2. Each run has 15 rounds. In each round, one agent is chosen under the rule for that condition. The agent reads the full thread, selects the post or an earlier comment as its reply target, and writes one comment. The new comment is added at once and becomes part of the context for later agents. In the two persona conditions, the agent also receives its demographic profile, writing-style profile, and a record of its own earlier comments. The user prompt and generation settings remain fixed. Only the agent information and the rule for choosing agents change. The target groups in the Underrepresented AI condition come from the gaps in \autoref{tab:motivating_bias} in Web Appendix A. The intervention changes only which agents are more likely to take part. The post, human-context seed, number of rounds, prompt, and writing-style mix remain fixed. \autoref{tab:representation_interventions} summarizes the sampling proportions for each domain.

\begin{table}[ht]
\centering
\scriptsize
\caption{Domain-Specific Changes in Participant Mix}
\label{tab:representation_interventions}
\begin{tabularx}{\textwidth}{
>{\raggedright\arraybackslash}p{3.5cm}
>{\raggedright\arraybackslash}p{3.1cm}
>{\raggedright\arraybackslash}X}
\toprule
\textbf{Domain} &
\textbf{Target group} &
\textbf{Rule} \\
\midrule

Gaming \& Esports &
Female personas &
Select a female persona with 80\% probability in each round \\

Technology \& Computing &
Female personas &
Select a female persona with 80\% probability in each round \\

Entertainment \& Pop Culture &
Personas over age 35 &
Select a persona over age 35 with 80\% probability in each round \\

People \& Relationships &
Introvert personas &
Add an introvert trait to 80\% of persona versions, then sample evenly from that pool \\

\bottomrule
\end{tabularx}
\end{table}

\begin{figure}[ht]
\centering
\includegraphics[width=0.95\textwidth]{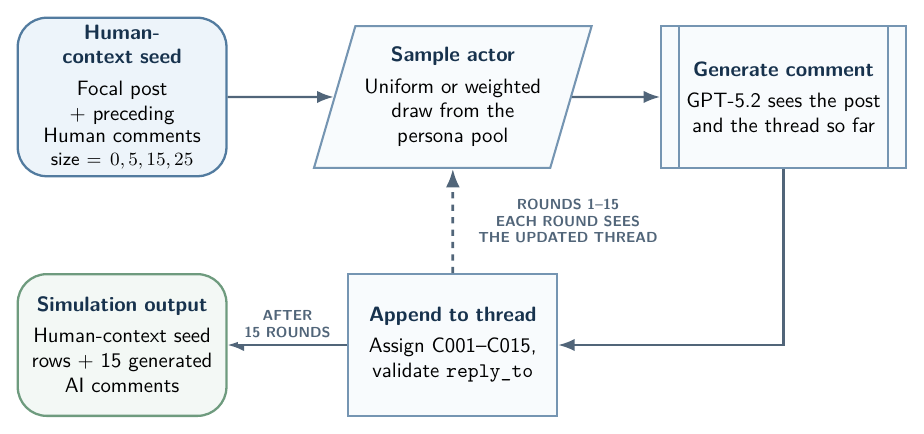}
\caption{Conversation Simulation Procedure}
\label{fig:method_agentic_simulation}
\vspace{0.6em}
\begin{minipage}{0.92\textwidth}
\scriptsize
\textit{Notes.} Each run begins with one Reddit post and 0, 5, 15, or 25 human-context seed comments. In each of 15 rounds, one agent is chosen, reads the full thread, writes a comment, and adds it to the thread.
\end{minipage}
\end{figure}

The full design has three conditions $\times$ four human-context seed levels $\times$ four domains $\times$ 30 posts per domain. Each run contains 15 generated comments. This gives 7,200 comments per condition and 21,600 comments in total. Figure~\ref{fig:method_agentic_simulation} shows the procedure, and all prompts used are available in the GenAI Web Appendix.

\FloatBarrier
\subsection*{Human-AI hybrid experiments}

\begin{figure}[htbp]
\centering

\begin{subfigure}[t]{0.31\textwidth}
    \centering
    \includegraphics[width=\linewidth]{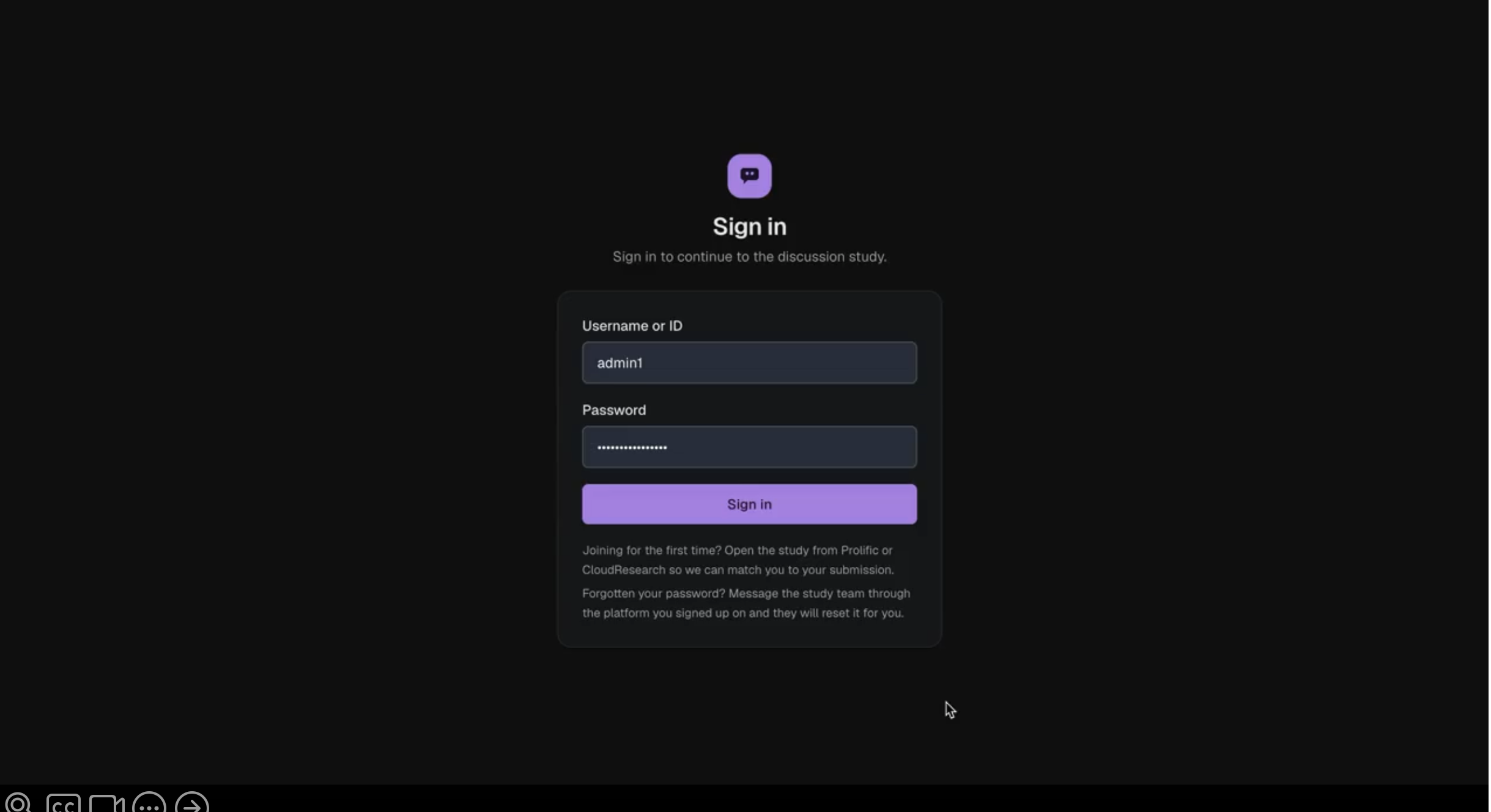}
    \caption{Administrator login}
    \label{fig:platform_login}
\end{subfigure}
\hfill
\begin{subfigure}[t]{0.31\textwidth}
    \centering
    \includegraphics[width=\linewidth]{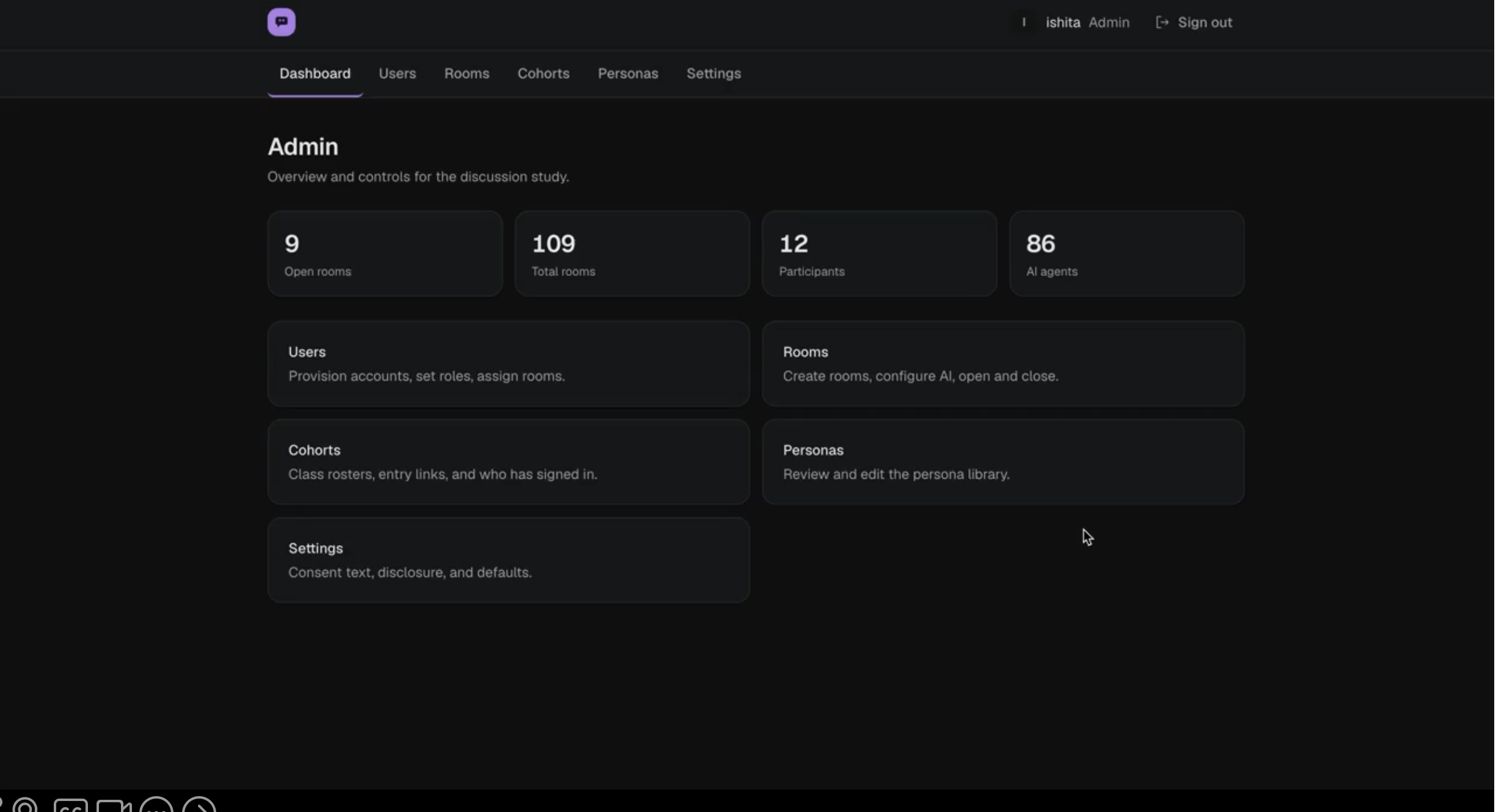}
    \caption{Study dashboard}
    \label{fig:platform_dashboard}
\end{subfigure}
\hfill
\begin{subfigure}[t]{0.31\textwidth}
    \centering
    \includegraphics[width=\linewidth]{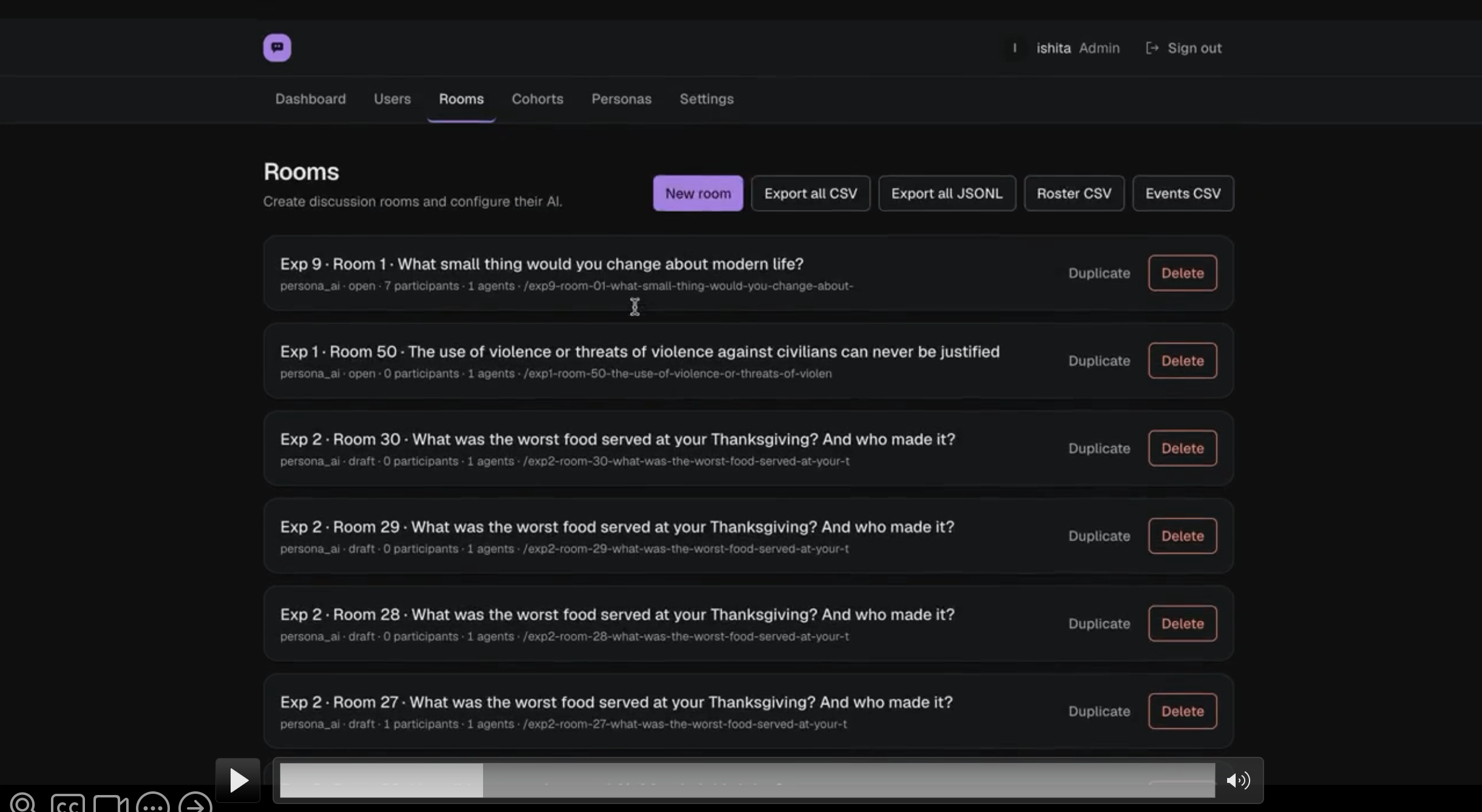}
    \caption{Discussion room management}
    \label{fig:platform_rooms}
\end{subfigure}

\vspace{0.8em}

\begin{subfigure}[t]{0.47\textwidth}
    \centering
    \includegraphics[width=\linewidth]{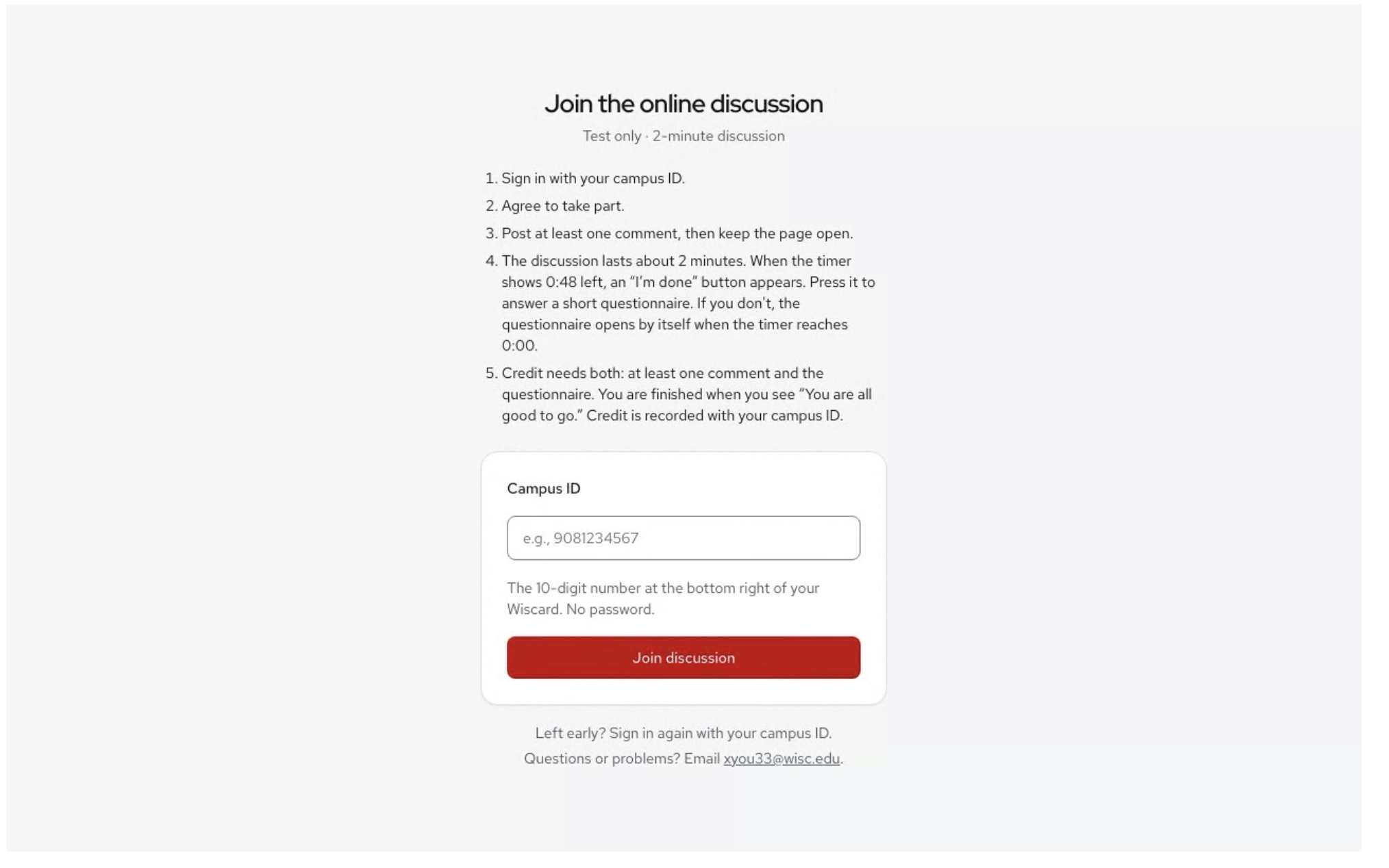}
    \caption{Participant study entry}
    \label{fig:platform_entry}
\end{subfigure}
\hfill
\begin{subfigure}[t]{0.47\textwidth}
    \centering
    \includegraphics[width=\linewidth]{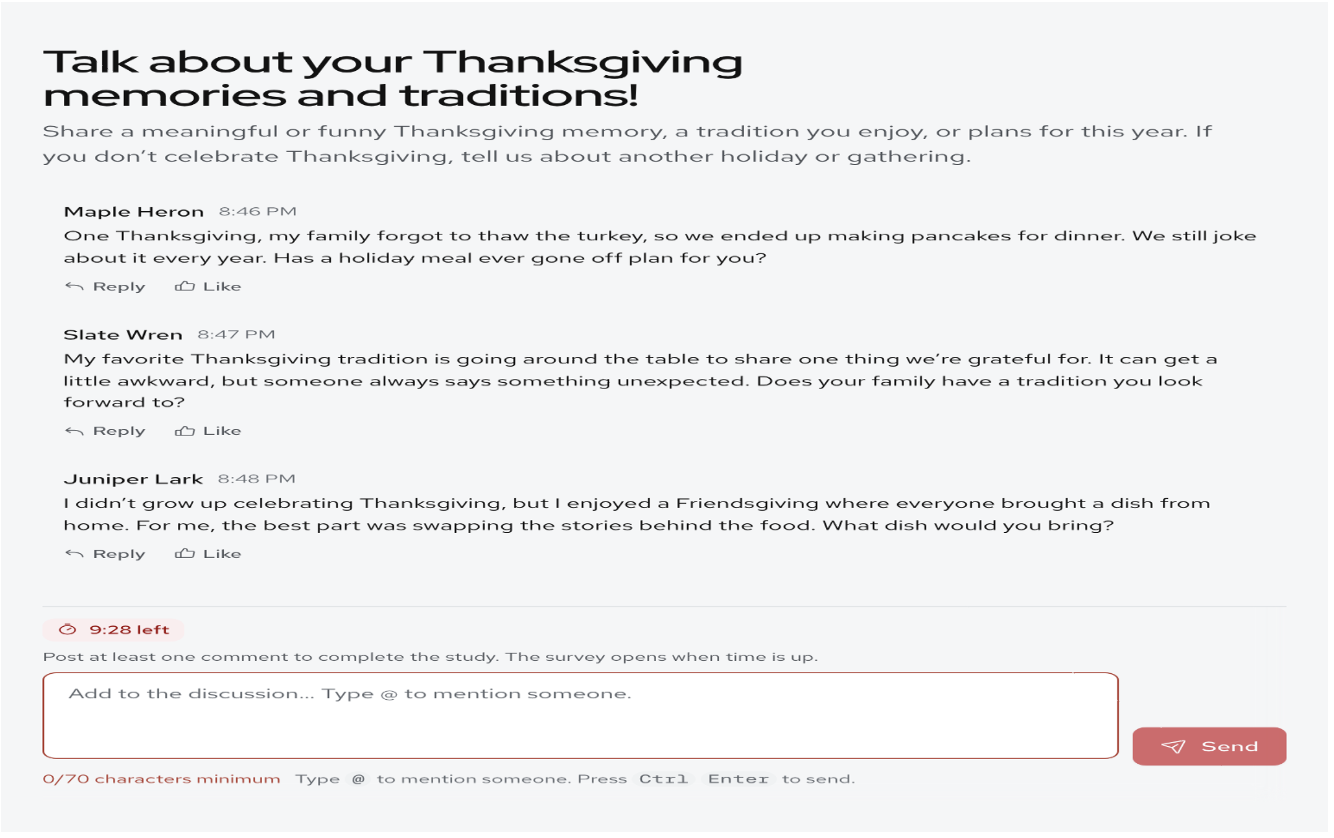}
    \caption{Human--AI discussion interface}
    \label{fig:platform_discussion}
\end{subfigure}

\caption{\textbf{Overview of the Human--AI Discussion Platform.}
Panels (a)--(c) illustrate the researcher-facing interface used for administrator access, study monitoring, and configuration of discussion rooms and AI agents. Panels (d)--(e) illustrate the participant-facing interface, from entering the study to contributing comments, replying to others, and interacting in asynchronous discussion threads.}
\label{fig:discussion_platform}

\end{figure}

\noindent To study how human participants respond to AI agents, we developed a custom online discussion platform that supports both human-only and hybrid human--AI conversations. Figure~\ref{fig:discussion_platform} illustrates the platform's main features. The participant-facing interface (d--e) resembles familiar online discussion forums such as Reddit and Discourse, allowing users to contribute to an ongoing discussion by posting new comments, replying to existing contributions, and liking posts. Each discussion room is organized around a common prompt and begins with three opening comments to initiate conversation. Discussions are asynchronous: participants can join at different times, view preceding contributions, and choose whether to initiate a new line of discussion or respond directly to another participant. The platform records the content and sequence of contributions, including comments, replies, and likes, enabling us to analyze both the tone and semantic content of discussions (e.g., sentiment, toxicity, and opinion diversity) and human engagement (e.g., participation, responsiveness, and conversational depth).

\begin{figure}
    \centering
    \includegraphics[width=0.65\linewidth]{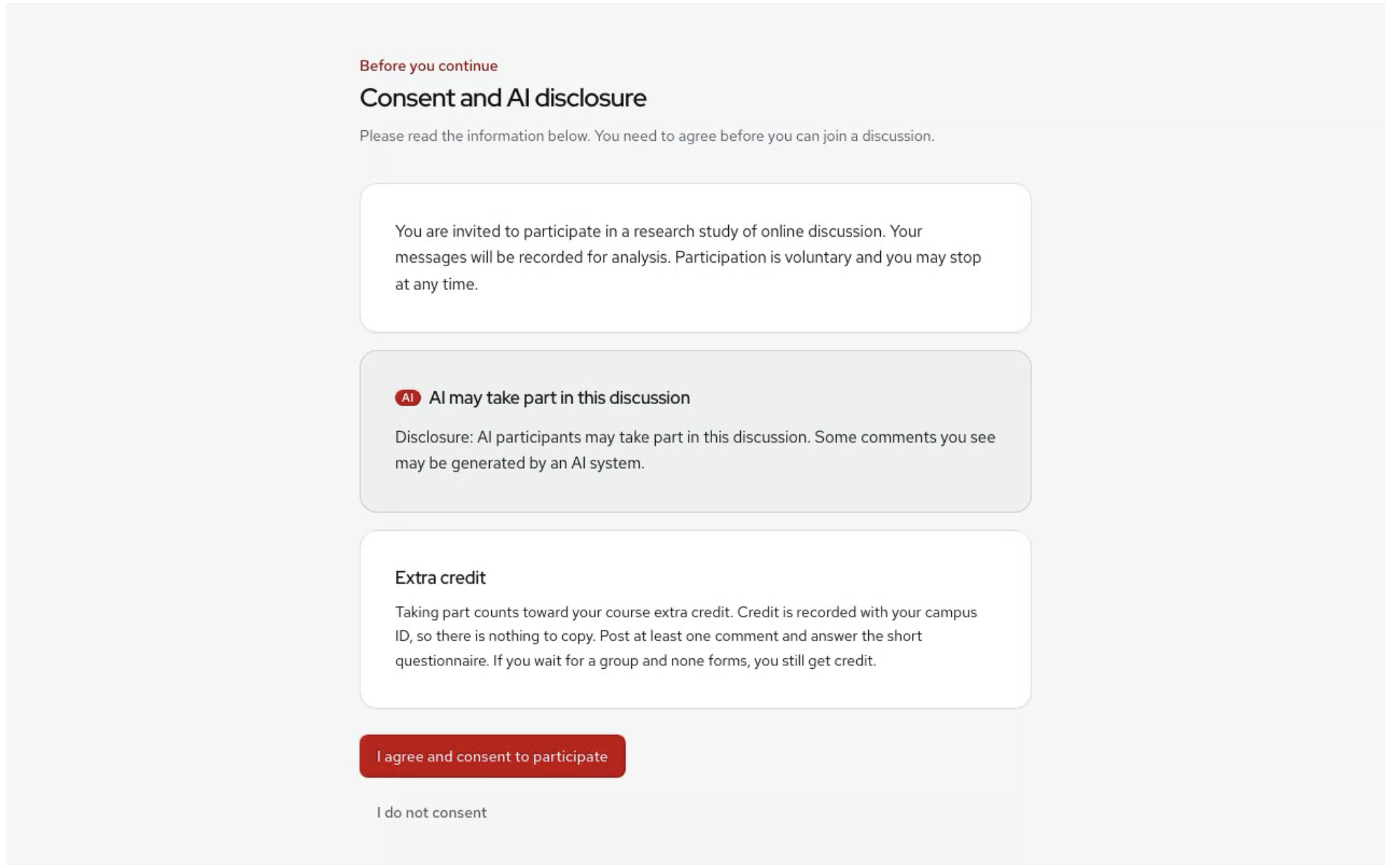}
    \caption{AI disclosure}
    \label{fig:disclosure}
\end{figure}

The platform also includes a researcher-facing administrative interface for managing discussion rooms, participants, and AI agents. As shown in Figure~\ref{fig:discussion_platform} (a---c), researchers can create and configure rooms, manage participant assignments, monitor ongoing discussions, and export discussion records for analysis. Each room can be configured to host human-only conversations or hybrid discussions with Generic AI or Persona AI agents, using the persona library developed in our simulation study. This setup allows us to vary the presence and type of AI participants while maintaining a common discussion interface across conditions, enabling comparisons of conversational outcomes between human-only and hybrid human--AI discussions. In AI-assisted rooms, agents participate alongside humans by generating context-dependent comments and replies based on the evolving discussion. AI contributions appear in the same format and under the same type of display names as human contributions, without individual AI labels. However, all participants are informed during consent that AI agents may participate, regardless of their assigned condition (\autoref{fig:disclosure}). Following the discussion, participants complete a short questionnaire containing six ratings and a question asking whether they believe any participants in their discussion were AI agents.

Following preregistration and Institutional Review Board (IRB) approval, we conducted the first wave of our human--AI discussion experiment over four days in November 2026. Participants were undergraduate students enrolled in a large marketing course at a U.S. university, who were invited to participate in exchange for extra credit. Students accessed the study through a common link that directed them to our discussion platform. We selected a discussion prompt centered on Thanksgiving memories and traditions, an accessible and familiar topic that allowed participants to share personal experiences, opinions, and anecdotes without requiring specialized knowledge. Participants were asked to contribute to an ongoing discussion by posting at least one comment, with the option to respond to other participants and continue interacting throughout the session. Although participants logged in using their campus IDs, they were automatically assigned fictional display names upon entering the discussion rooms. Their real identities were therefore not visible to other participants, creating a pseudonymous environment similar to online discussion forums such as Reddit, where users typically interact without knowing one another's offline identities. This design was intended to reduce concerns about being personally identified and encourage participants to express their views more freely, while allowing researchers to track participation for study administration.

The experiment consisted of three conditions: (1) No AI, in which discussions involved only human participants; (2) Generic AI, in which human participants interacted with AI agents without persona conditioning; and (3) Persona AI, in which AI agents were conditioned on empirically grounded personas developed in our simulation study. Each condition contained three discussion threads, yielding nine threads in total. Upon joining the platform, participants were assigned to the discussion thread with the fewest participants, allowing participation to be distributed across rooms as the study progressed. In both AI-assisted conditions, four AI agents were available to participate in each thread. After every two human comments, one of these agents was randomly selected and given an opportunity to respond to the ongoing discussion. This procedure introduced AI participation at regular intervals tied to human activity, while allowing the content of AI contributions to adapt to the evolving conversation. All three conditions used the same discussion interface and initial discussion structure, with the presence and type of AI agents serving as the primary experimental differences.

\begin{figure}[!htbp]
\centering
\includegraphics[width=0.95\textwidth]{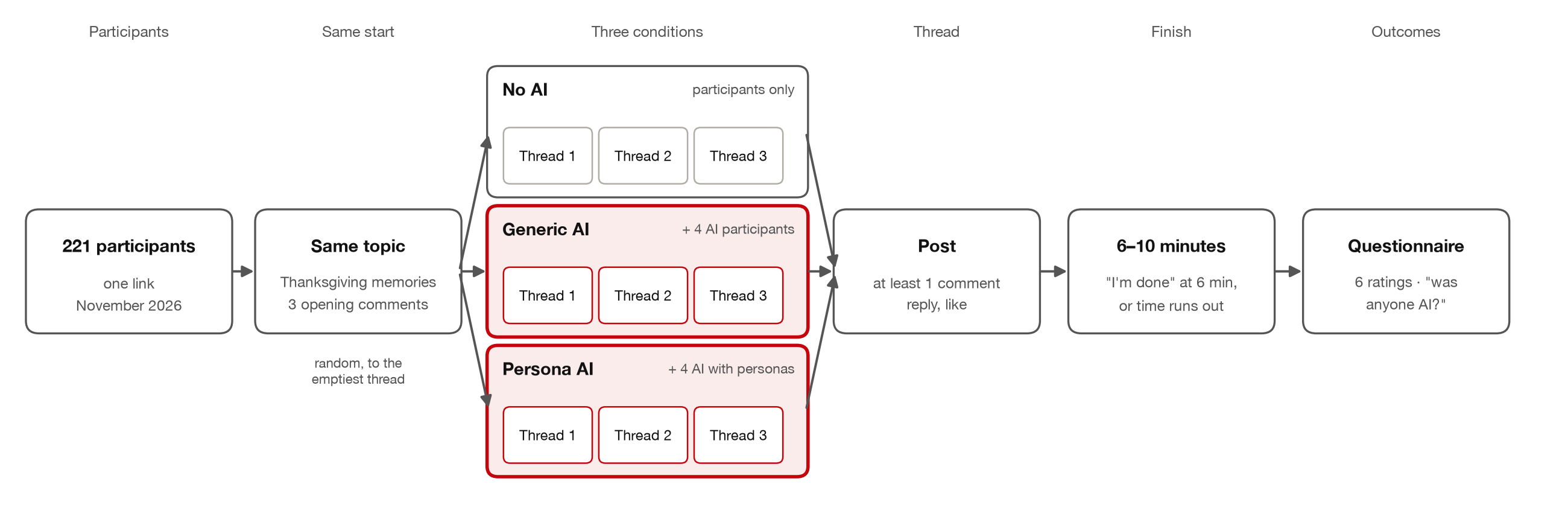}
\caption{Experiment Design: Three Conditions, Three Threads Each}
\label{fig:classroom_flow}
\end{figure}

Discussions were asynchronous, allowing participants to join at different times and view all contributions made before their arrival. During their initial visit, participants were required to post at least one comment and could complete the discussion portion of the study after six minutes. They were then directed to a short questionnaire assessing their experience and perceptions of the interaction, including whether they believed AI agents had participated. Importantly, participants could return to their assigned discussion thread throughout the four-day study period. On each return visit, they could view all comments posted since their previous participation and engage in the evolving discussion for another six-minute session. They were not required to complete the questionnaire again on subsequent visits. This design allowed discussions to develop over time, with returning participants encountering new contributions and potentially responding to interactions that occurred in their absence. In total, 221 students participated across the nine threads, with 73 assigned to No AI, 74 to Generic AI, and 74 to Persona AI. Questionnaire completion was high, with 94\% of participants submitting their responses. Figure~\ref{fig:classroom_flow} summarizes the experimental design and participant flow.

\FloatBarrier
\section*{Simulation Results}

\begin{table}[!htbp]
\centering
\scriptsize
\caption{Overview of Validation Checks, Outcome Measures, and Main Results}
\label{tab:outcome_measures}
\setlength{\tabcolsep}{4pt}
\begin{tabularx}{\textwidth}{
>{\raggedright\arraybackslash}p{2.3cm}
>{\raggedright\arraybackslash}p{3.0cm}
>{\raggedright\arraybackslash}p{4.4cm}
>{\raggedright\arraybackslash}X}
\toprule
\textbf{Analysis} &
\textbf{Question} &
\textbf{Measure and comparison} &
\textbf{Main result} \\
\midrule

Conversation validity &
Do agents respond to the specific thread they join? &
JSD (Jensen-Shannon divergence) between generated comments and held-out human comments from the same post versus other posts in the domain, over a frozen human-only topic model; lower is closer &
Same-post JSD is lower than other-post JSD for every persona version, indicating agents respond specifically to the thread. \\

\midrule

Persona validity &
Do persona agents that combine demographics and writing style better resemble the human groups they represent? &
JSD between generated and held-out human comments within 49 domain $\times$ gender $\times$ age $\times$ writing-style groups; full persona versus reduced versions &
Full persona fits best \\

\midrule

Text style and tone &
How do AI and human comments differ in style and tone, and does human context narrow the gap? &
Word count, first-person count, subjectivity (TextBlob), sentiment (VADER), toxicity (Detoxify); AI minus human at the post level, and seed 25 minus seed 0 &
Personas remove the length and first-person gaps. All AI conditions stay more positive and less toxic, more human context only slightly attenuates this. \\

\midrule

Semantic breadth &
Do personas widen the range of ideas expressed? &
Mean pairwise SBERT cosine similarity over the first 15 comments; lower means broader; Generic vs. Persona AI, and AI vs. human &
Persona grounding lowers similarity in all four domains, but human threads remain broader. Underrepresented AI adds no breadth. \\

\midrule

Topic attention after reweighting &
Does oversampling underrepresented personas change what is discussed? &
BERTopic fitted separately by domain and condition; share of Underrepresented AI topics with no keyword-Jaccard match in Persona AI or human discussion, minus the Persona AI share &
Descriptive shifts in discussions toward e.g., work, design, and everyday effects of technology \\

\midrule

All-AI reference &
Do the patterns extend to a naturally occurring all-AI platform? &
Same tone, style, and similarity measures on 485 Moltbook threads; descriptive, unmatched to the Reddit posts &
Moltbook is broader than Persona AI but narrower than human, and even more positive and less toxic.\\

\bottomrule
\end{tabularx}

\vspace{0.4em}
\begin{minipage}{0.95\textwidth}
\scriptsize
\textit{Notes.} All measures are computed identically for human and AI-generated comments, with the post as the unit of analysis. Tone and style analyses average human-context seeds 0, 5, 15, and 25; semantic-breadth and topic-attention analyses use seeds 0, 5, and 15. The validity checks assign comments to topic models estimated on human comments only; the topic-attention analysis instead fits separate models for each domain and condition. Validation tests use one-sided paired Wilcoxon tests. Other inferential tests are paired and two-sided with Holm correction within each comparison family, except the Moltbook comparisons, which use two-sided Welch tests on independent posts.
\end{minipage}
\end{table}

After conducting our simulation exercise, we first validate the generated conversations by assessing whether comments remain grounded in the source thread and whether persona conditioning produces behavior that more closely resembles the corresponding human segments. We then compare key characteristics of agent-generated and human discussions, including word count, sentiment, and toxicity. Finally, we examine data from Moltbook, an all-AI social media platform, to assess whether the patterns observed in our simulations generalize to naturally occurring AI-agent interactions. A high-level summary of the methods and results is provided in \autoref{tab:outcome_measures}.
\subsection*{Validation of the Generated Conversations}

\noindent Before using the simulated conversations for our main analyses, we assess the validity of two aspects of our persona construction strategy (combining demographics and writing style). First, we examine \textit{conversation validity}: do agents respond to the specific discussion in which they are placed, rather than generating comments that are merely relevant to the broader domain? Second, we examine \textit{persona validity}: do agents generate comments that more closely resemble the behavior of the human group the persona is intended to represent?

We assess both questions using the topical content of human discussions as a common benchmark. %
We estimate a separate BERTopic model for each of the four domains using human comments only \citep{grootendorst2022bertopic}. We then hold these models fixed and use them to assign both held-out human comments and AI-generated comments to the resulting human-defined topics. This approach allows us to evaluate generated comments relative to the structure of human conversations without allowing AI-generated content to determine the topics themselves.

For topic distributions $P$ and $Q$, we use Jensen--Shannon divergence:
\[
\operatorname{JSD}(P,Q)
=
\frac{1}{2}\operatorname{KL}(P\mathbin{\|}M)
+
\frac{1}{2}\operatorname{KL}(Q\mathbin{\|}M),
\qquad
M=\frac{P+Q}{2}.
\]
With base-2 logarithms, JSD ranges from 0 to 1. Lower values mean that two sets of comments place similar weight on the same topics. We conduct these validation exercises for three versions of persona conditioning to assess whether the information included in our full personas contributes to their performance. The \textit{Full} specification includes demographics, writing style, and example comments; the \textit{Demographics only} specification excludes writing-style information and examples; and the \textit{Writing style only} specification excludes demographic information.

\paragraph{Conversation validity: Do agents respond to the source discussion?}
Our first validation asks whether generated comments remain anchored to the particular discussion being simulated. For each of the 120 source posts, we hold out 15 length-matched human comments. The remaining human comments are used to estimate the domain-level topic models described above. We then compare the topic distribution of each generated continuation with two benchmarks: (1) the held-out human comments from the \textit{same post}, and (2) the held-out human comments from the other 29 posts in the same domain. If agents are responding to the specific conversation rather than simply producing domain-relevant content, their comments should be substantially closer to human comments from the same post than to human comments from other posts in that domain.

The three persona specifications use identical source posts, simulation rounds, and agent assignments. Paired Wilcoxon tests compare same-post and other-post distances. We additionally compare the same-post distance under the Full specification with each reduced specification, applying a Holm correction to these two comparisons. \autoref{tab:post_validation} shows that for all three persona specifications, generated comments are substantially closer to held-out human comments from the same post than to comments from other posts in the same domain ($p<.001$). Under the Full specification, for example, average JSD is $0.0442$ for the source post compared with $0.2263$ for other posts. Thus, the simulated agents are not simply generating comments appropriate to a broad domain; their topical content tracks the particular conversation they are joining.

The Full specification also produces the closest fit to the source discussion. Its same-post JSD of $0.0442$ is lower than that of both Demographics only ($0.0498$) and Writing style only ($0.0483$), with both differences remaining significant after Holm correction ($p=.039$). Thus, combining demographic information, writing style, and behavioral examples improves the agents' ability to remain aligned with the source discussion.

\begin{table}[!htbp]
\centering
\scriptsize
\caption{Fit to the Source Discussion}
\label{tab:post_validation}
\begin{tabularx}{\textwidth}{
>{\raggedright\arraybackslash}p{4.1cm}
>{\centering\arraybackslash}X
>{\centering\arraybackslash}X
>{\centering\arraybackslash}X}
\toprule
\textbf{Persona version} &
\textbf{JSD to same post} &
\textbf{JSD to other posts} &
\textbf{$p$ vs. full persona} \\
\midrule
Full persona & 0.0442 & 0.2263 & N/A \\
Demographics only & 0.0498 & 0.2475 & .039 \\
Writing style only & 0.0483 & 0.2365 & .039 \\
\bottomrule
\end{tabularx}

\vspace{0.3em}
\begin{minipage}{0.94\textwidth}
\scriptsize
\textit{Notes.} JSD to the same post is lower than JSD to other posts for every persona version ($p<.001$). The last column compares the full persona with each simpler version using one-sided paired Wilcoxon tests with Holm correction.
\end{minipage}
\end{table}

\paragraph{Persona validity: Do agents resemble their intended human groups?}
Our second validation asks whether full persona conditioning best produces behavior that resembles the human segments the personas are designed to represent. We define 49 human groups based on combinations of domain, gender, age, and LIWC writing style that are represented in our data. For each group, we compare the topic distribution of generated comments with the topic distribution of held-out human comments from the corresponding group. All three persona specifications use identical human groups, source posts, and agent assignments, allowing us to isolate whether the richer persona construction improves alignment with the target human group.

The Full persona produces the closest average match to the corresponding human groups, with a JSD of 0.1558, compared with 0.1626 for Demographics only and 0.1615 for Writing style only. After Holm correction, the improvement over Demographics only is marginally significant at the 10\% level ($p=.064$), while the difference relative to Writing style only is directionally consistent but not statistically significant ($p=.104$). Thus, this persona-validity check provides suggestive evidence that incorporating writing style and behavioral examples improves the match beyond demographics alone, while the incremental contribution of demographics beyond writing style and examples is positive but slightly more noisy. %

\begin{table}[!htbp]
\centering
\scriptsize
\caption{Fit to the Target Human Groups}
\label{tab:persona_cell_validation}
\begin{tabularx}{\textwidth}{
>{\raggedright\arraybackslash}p{4.1cm}
>{\centering\arraybackslash}X
>{\centering\arraybackslash}X
>{\centering\arraybackslash}X}
\toprule
\textbf{Persona version} &
\textbf{JSD to matched human group} &
\textbf{Gap from full persona} &
\textbf{Holm-adjusted $p$ vs. full persona} \\
\midrule
Full persona & 0.1558 & N/A & N/A \\
Demographics only & 0.1626 & $+0.0068$ & .064 \\
Writing style only & 0.1615 & $+0.0057$ & .104 \\
\bottomrule
\end{tabularx}

\vspace{0.3em}
\begin{minipage}{0.94\textwidth}
\scriptsize
\textit{Notes.} The gap is the listed JSD minus the full-persona JSD, so a positive value favors the full persona. $p$-values come from one-sided paired Wilcoxon tests and are Holm-adjusted across the two full-persona-versus-ablation comparisons.
\end{minipage}
\end{table}

\subsection*{Text Style and Tone}

\noindent We next compare the writing style and tone of AI vs human discussions along five dimensions: comment length, first-person language, subjectivity, sentiment, and toxicity. We measure comment length and first-person language using word counts, subjectivity using TextBlob, sentiment using the VADER compound score \citep{hutto2014vader}, and toxicity using the \texttt{original} Detoxify model \citep{hanu2020detoxify}. Higher sentiment scores indicate more positive language, while higher toxicity scores indicate a greater likelihood of uncivil or harmful language. All measures are computed identically for human and AI conditions.

\autoref{tab:tone_results} reports the descriptive means. First, we see that persona grounding brings the basic writing features of AI comments close to human values. Generic AI comments average 79 words, more than twice the human average of 34, whereas Persona AI and Underrepresented AI comments average 36 and 37 words. Human comments contain 1.41 first-person words on average, compared with 1.12 for Generic AI but 1.49 and 1.47 for the two persona conditions. Subjectivity is somewhat higher in all AI conditions (0.47 to 0.48 versus 0.42). Second, differences in tone persist. Average sentiment is 0.14 in human discussion, 0.29 for Generic AI, and 0.22 for both persona conditions; average toxicity is 0.108 in human discussion and only 0.028 to 0.035 across the three AI conditions. Thus, persona grounding makes basic writing patterns more human-like, while AI-generated discussions remain systematically more positive and less toxic.

We test these differences at the post level so that comments from the same discussion are not treated as independent observations. For each AI condition and each of the 120 posts, we average every measure across the four human-context seeds (0, 5, 15, and 25) and pair the result with the human mean for the same source post. Panel A of \autoref{tab:text_calibration_tests} reports two-sided paired tests with Holm adjustment within each comparison family. The tests confirm the descriptive pattern. Generic AI differs from human discussion on every measure ($p<.001$). For Persona AI and Underrepresented AI, the differences in word count ($+2.0$ and $+2.4$ words, $p=.449$ and $p=.330$) and first-person language ($+0.08$ and $+0.06$, both $p=.687$) are small and not statistically significant, while the differences in sentiment ($+0.081$ and $+0.084$), toxicity ($-0.073$ and $-0.074$), and subjectivity ($+0.063$ and $+0.060$) all remain significant ($p<.001$).

\begin{figure}[!htbp]
\centering

\begin{subfigure}[t]{0.90\textwidth}
    \centering
    \includegraphics[width=\textwidth]{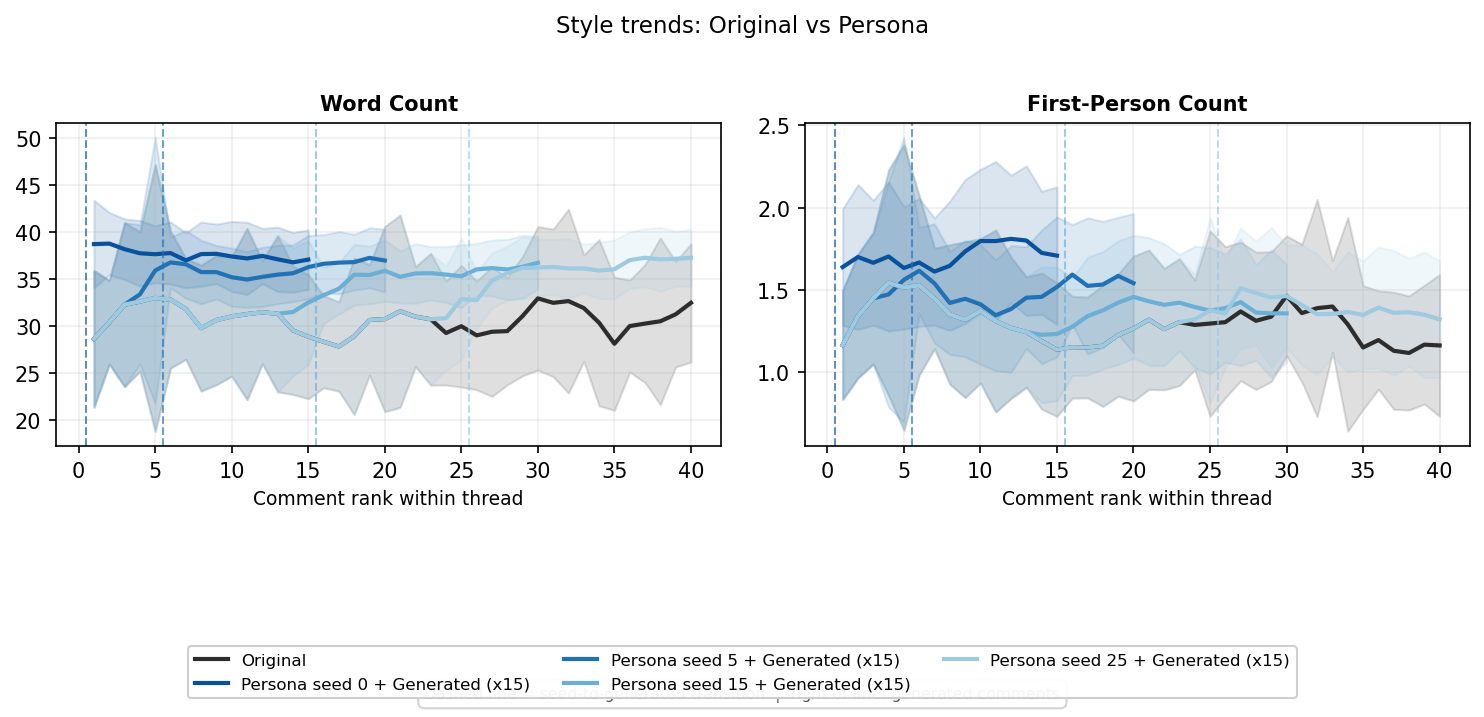}
    \label{fig:persona_timeseries_style}
\end{subfigure}

\vspace{0.8em}

\begin{subfigure}[t]{0.32\textwidth}
    \centering
     \caption*{\textbf{\footnotesize{Sentiment}}}
    \includegraphics[width=\textwidth]{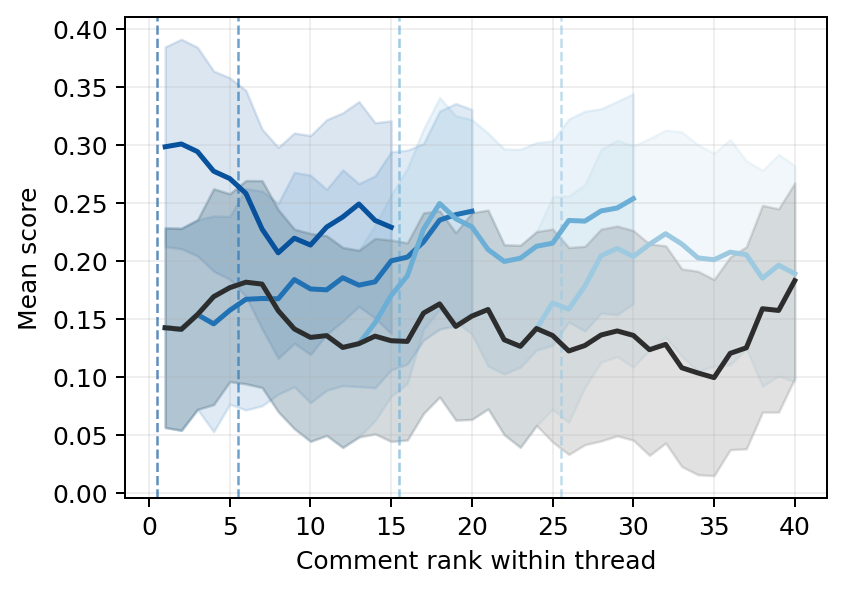}
    \label{fig:persona_rank_sentiment}
\end{subfigure}
\hfill
\begin{subfigure}[t]{0.32\textwidth}
    \centering
    \caption*{\textbf{\footnotesize{Toxicity}}}
    \includegraphics[width=\textwidth]{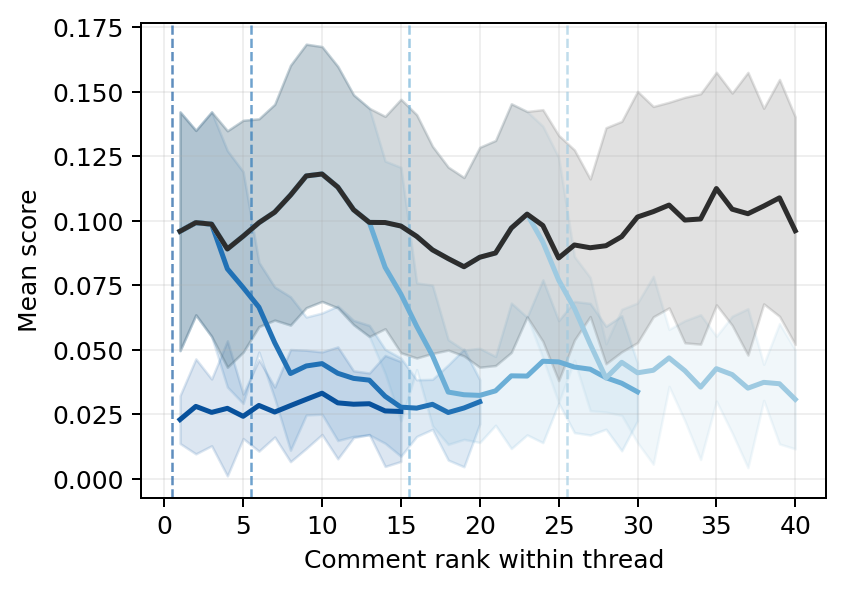}
    \label{fig:persona_rank_toxicity}
\end{subfigure}
\hfill
\begin{subfigure}[t]{0.32\textwidth}
    \centering
     \caption*{\textbf{\footnotesize{Subjectivity}}}
    \includegraphics[width=\textwidth]{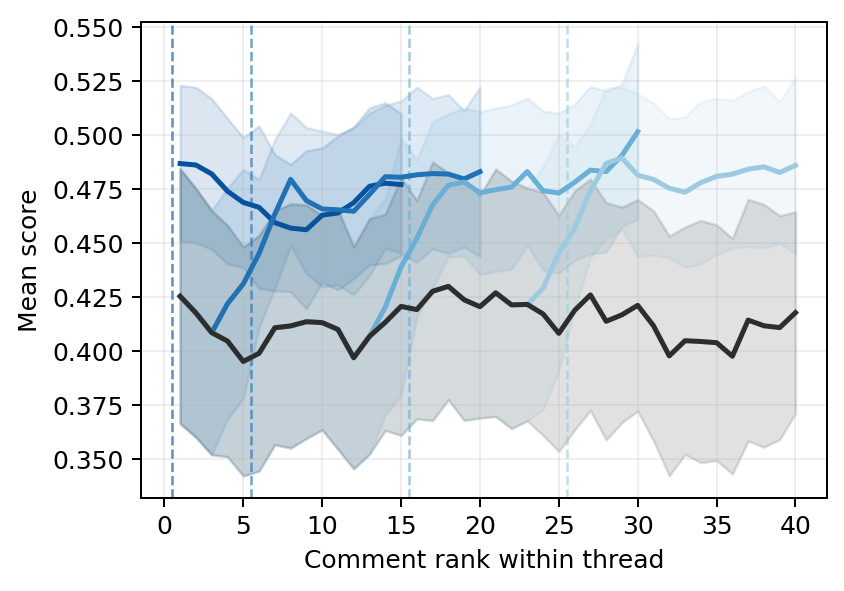}
   
    \label{fig:persona_rank_subjectivity}
\end{subfigure}

\caption{Persona AI Tone and Writing-Style Trajectories by Comment Rank}
\label{fig:persona_timeseries_all}

\vspace{0.5em}
\begin{minipage}{0.92\textwidth}
\scriptsize
\textit{Notes.} Original denotes the human threads; Baseline denotes Generic AI. The top panel shows Persona AI trajectories for comment
length and first-person language. The bottom panels show sentiment,
toxicity, and subjectivity. Black lines show the original human threads.
Blue lines from dark to light show Persona AI runs with 0, 5, 15, and
25 human-context seed comments. Dashed lines mark where AI enters.
Shaded bands show 95\% confidence intervals.
\end{minipage}

\end{figure}

A natural question is whether these tone differences shrink when agents enter a thread that already contains more human discussion. \autoref{fig:persona_timeseries_all} plots the five measures by comment rank for Persona AI under each human-context seed level, with the original human threads as a reference; corresponding figures for Generic AI and Underrepresented AI are in Figures W2 and W3 in Web Appendix A. The seed-25 trajectories lie closest to the human lines, but the gap in sentiment and toxicity persists at every seed level. Panel B of \autoref{tab:text_calibration_tests} quantifies this by comparing seed 25 with seed 0 within each post. Sentiment declines and toxicity rises in all three conditions, so more human context does move AI tone toward human tone, but the changes are small relative to the gaps in Panel A. Generic AI toxicity rises by 0.012 ($p=.024$), the only tone change that remains significant after Holm correction, against a Generic-minus-human toxicity difference of $-0.080$; the corresponding sentiment decline of $-0.074$ ($p=.054$) offsets less than half of the $+0.156$ gap. Among the writing-style measures, first-person language declines significantly from seed 0 to seed 25 for Persona AI ($-0.30$, $p=.007$) and Underrepresented AI ($-0.25$, $p=.036$), whereas word count and subjectivity do not change significantly. In short, supplying more human context moves AI comments slightly toward human tone and style, but it does not close the gap.

\begin{table}[!htbp]
\centering
\scriptsize
\caption{Text Measures Across human and AI Conditions}
\label{tab:tone_results}
\setlength{\tabcolsep}{4pt}
\begin{tabular}{p{3.7cm}cccc}
\toprule
Metric & Human & \shortstack{Generic\\AI} & \shortstack{Persona\\AI} & \shortstack{Underrepresented\\AI} \\
\midrule
Sentiment (VADER) & 0.14 (0.12) & 0.29 (0.29) & 0.22 (0.25) & 0.22 (0.24) \\
Toxicity (Detoxify) & 0.108 (0.062) & 0.028 (0.043) & 0.035 (0.041) & 0.034 (0.051) \\
Word count & 34 (16) & 79 (26) & 36 (14) & 37 (15) \\
First-person count & 1.41 (0.97) & 1.12 (1.14) & 1.49 (1.41) & 1.47 (1.42) \\
Subjectivity (TextBlob) & 0.42 (0.07) & 0.47 (0.06) & 0.48 (0.07) & 0.47 (0.07) \\
\midrule
Post-level observations ($N$) & 120 & 480 & 480 & 480 \\
Underlying comments & 48,315 & 7,200 & 7,200 & 7,200 \\
\bottomrule
\end{tabular}

\vspace{0.4em}
\begin{minipage}{0.94\textwidth}
\scriptsize
\textit{Notes.} Values are means with standard deviations in parentheses. human values use 120 thread-level means and 48,315 comments. Each AI condition uses 480 post-by-seed observations and 7,200 generated comments (120 posts $\times$ 4 human-context seeds $\times$ 15 comments). The standard deviations use different units, so we read them only as descriptive values.
\end{minipage}
\end{table}

\begin{table}[!htbp]
\centering
\scriptsize
\caption{AI-human Differences and human-Context Calibration in Text Measures}
\label{tab:text_calibration_tests}
\setlength{\tabcolsep}{4.5pt}
\begin{tabularx}{\textwidth}{
>{\raggedright\arraybackslash}p{3.0cm}
>{\centering\arraybackslash}X
>{\centering\arraybackslash}X
>{\centering\arraybackslash}X}
\toprule
\textbf{Outcome} &
\textbf{Generic AI} &
\textbf{Persona AI} &
\textbf{Underrepresented AI} \\
\midrule

\multicolumn{4}{l}{\textbf{Panel A: Difference from human}} \\

Sentiment
& $+0.156$ ($<0.001$)
& $+0.081$ ($<0.001$)
& $+0.084$ ($<0.001$) \\

Toxicity
& $-0.080$ ($<0.001$)
& $-0.073$ ($<0.001$)
& $-0.074$ ($<0.001$) \\

Word count
& $+44.825$ ($<0.001$)
& $+2.009$ ($0.449$)
& $+2.428$ ($0.330$) \\

First-person count
& $-0.293$ ($<0.001$)
& $+0.079$ ($0.687$)
& $+0.064$ ($0.687$) \\

Subjectivity
& $+0.055$ ($<0.001$)
& $+0.063$ ($<0.001$)
& $+0.060$ ($<0.001$) \\

\addlinespace
\multicolumn{4}{l}{\textbf{Panel B: Seed 25 minus seed 0}} \\

Sentiment
& $-0.074$ ($0.054$)
& $-0.049$ ($0.217$)
& $-0.034$ ($0.227$) \\

Toxicity
& $+0.012$ ($0.024$)
& $+0.012$ ($0.054$)
& $+0.008$ ($0.227$) \\

Word count
& $-1.831$ ($0.724$)
& $-1.361$ ($0.077$)
& $-0.824$ ($0.659$) \\

First-person count
& $-0.317$ ($0.077$)
& $-0.303$ ($0.007$)
& $-0.252$ ($0.036$) \\

Subjectivity
& $+0.010$ ($0.659$)
& $+0.010$ ($0.659$)
& $+0.007$ ($0.724$) \\

\bottomrule
\end{tabularx}

\vspace{0.3em}
\begin{minipage}{0.94\textwidth}
\scriptsize
\textit{Notes.}
Cells report signed differences with Holm-adjusted $p$-values in
parentheses. Panel A reports AI-minus-human differences; AI values
average human-context seeds 0, 5, 15, and 25. Panel B reports changes
from human-context seed 0 to seed 25. Tests use 120 matched posts and
are paired and two-sided. Holm adjustment is applied separately within
the relevant comparison families. %
\end{minipage}
\end{table}

\FloatBarrier

\subsection*{Semantic breadth}

\noindent 
We next ask whether persona grounding changes the range of ideas expressed within a conversation. If agents repeatedly make similar points, the discussion is semantically narrow; if their comments raise different arguments, experiences, or perspectives, the discussion is broader. We measure semantic breadth using mean pairwise cosine similarity between SBERT text vectors for the first 15 comments in each discussion \citep{reimers2019sbert}:
\[
\text{CosineSim}
=
\frac{2}{n(n-1)}
\sum_{i<j}
\frac{\mathbf{e}_i \cdot \mathbf{e}_j}
{\|\mathbf{e}_i\|\,\|\mathbf{e}_j\|}.
\]
Lower similarity means that comments are less alike and cover a wider range of ideas.\footnote{Figure W4 in Web Appendix A plots the post-level distribution of pairwise similarity for each AI condition against the matched human threads, separately by human-context seed. The plots are descriptive; the paired tests reported in Table 9 are the inferential comparisons.} Human values use the first 15 comments in each matched thread. AI comparisons average human-context seeds 0, 5, and 15 within each post. We compare Generic AI with Persona AI within each domain using paired two-sided tests with Holm correction.

Generic AI produces relatively similar comments, with an average similarity of 0.541. When the same conversations are generated by agents grounded in heterogeneous personas, similarity falls to 0.458. The within-post difference of 0.0836 is statistically significant (Holm-adjusted $p<.001$). Thus, persona grounding does more than change how individual agents write: it also increases the range of ideas expressed collectively in the conversation. Human conversations provide an important benchmark. Their average similarity is only 0.225. Thus, although persona grounding makes AI conversations substantially less homogeneous, the resulting discussions remain much more semantically concentrated than human conversations. Average similarity under Underrepresented AI is 0.459, virtually identical to Persona AI at 0.458. Thus, changing who is represented does not necessarily increase the total breadth of the conversation. However, it can change which perspectives receive attention, and we examine these shifts in topical attention in the next section.

\begin{table}[ht]
\centering
\scriptsize
\caption{Semantic Breadth Across Human and AI Discussions}
\label{tab:semantic_diversity}

\begin{tabularx}{\textwidth}{
>{\raggedright\arraybackslash}p{3.0cm}
>{\centering\arraybackslash}X
>{\centering\arraybackslash}X
>{\centering\arraybackslash}X
>{\centering\arraybackslash}X
>{\centering\arraybackslash}p{1.5cm}
>{\centering\arraybackslash}p{2.3cm}}
\toprule
& \multicolumn{4}{c}{\textbf{Mean comment similarity}} &
\multicolumn{2}{c}{\textbf{Generic AI vs. Persona AI}} \\
\cmidrule(lr){2-5}\cmidrule(lr){6-7}

\textbf{Domain} &
\textbf{Human} &
\textbf{Generic AI} &
\textbf{Persona AI} &
\textbf{Underrep. AI} &
\textbf{Difference} &
\textbf{$p$-value} \\
\midrule

Technology \& Computing
& 0.253
& 0.595
& 0.447
& 0.452
& 0.148
& $<.001$ \\

Gaming \& Esports
& 0.220
& 0.548
& 0.458
& 0.464
& 0.090
& $<.001$ \\

People \& Relationships
& 0.245
& 0.560
& 0.521
& 0.528
& 0.039
& $<.01$ \\

Entertainment \& Pop Culture
& 0.180
& 0.463
& 0.406
& 0.393
& 0.057
& $<.001$ \\

\midrule
\textbf{Overall}
& \textbf{0.225}
& \textbf{0.541}
& \textbf{0.458}
& \textbf{0.459}
& \textbf{0.084}
& $\mathbf{<.001}$ \\

\bottomrule
\end{tabularx}

\vspace{0.4em}
\begin{minipage}{0.96\textwidth}
\scriptsize
\textit{Notes.} Entries in the first four columns report mean post-level
SBERT cosine similarity; lower values indicate greater semantic breadth.
The Difference column reports Generic AI minus Persona AI similarity, so
positive values indicate that persona grounding increases semantic breadth.
Human values use the first 15 comments in each matched thread; AI values
average human-context seeds 0, 5, and 15 within each post.
Generic--Persona tests are paired and two-sided. The overall comparison is the primary test; domain-level $p$-values are Holm-adjusted across the four domains. All three AI conditions also
differ significantly from human discussion overall (Holm-adjusted $p<.001$).
\end{minipage}
\end{table}

These patterns appear in all four domains (\autoref{tab:semantic_diversity}), although their magnitudes vary. The largest increase in semantic breadth occurs in Technology \& Computing, where similarity falls by 0.1482, followed by Gaming \& Esports (0.0897). The effect is smaller in Entertainment \& Pop Culture (0.0570) and People \& Relationships (0.0393), but remains statistically significant in both domains. Increasing the representation of underrepresented personas does not significantly increase overall semantic breadth across any domain.

\subsubsection*{Changes in Topic Attention After Reweighting}
\noindent
The previous analysis shows that increasing the representation of
underrepresented personas does not increase the overall semantic breadth of
the conversation. However, breadth and content are not equivalent: two
discussions can be equally broad while focusing on different issues. We
therefore examine whether changing the participant mix redirects attention
toward different concerns.

This analysis is descriptive. We compare Human, Persona AI, and
Underrepresented AI discussions for the same 120 posts, balancing the
comparison within each domain so that each condition contributes the same
number of comments from the same source posts. Within each domain and
condition, we estimate topics separately using BERTopic and summarize each
topic by its eight leading keywords. Because topics are estimated separately
across conditions, they are not directly aligned. We therefore match topics
based on keyword overlap, measured using Jaccard similarity:
\[
J(A,B)=\frac{|A\cap B|}{|A\cup B|},
\]
where $A$ and $B$ denote the sets of eight leading keywords for two topics.
For each topic, we identify its closest match in the other conditions. Topics
with at least three of eight keywords in common are treated as capturing the
same underlying concern.\footnote{Implementationally, two topics are treated as capturing the same underlying concern when their Jaccard similarity J is at least 0.15. Because every topic is summarized by exactly eight keywords, if two topics share k keywords, their union contains 16 - k distinct keywords (the two lists of eight, minus the k counted twice), so J = k/(16 - k). Sharing two keywords gives J = 2/14 = 0.14, which falls just below the threshold; sharing three gives J = 3/13 = 0.23, which clears it. The 0.15 rule therefore matches two topics exactly when they share at least three of their eight keywords.} We classify a topic as unmatched when it has no such counterpart in either of the other two conditions: an Underrepresented AI topic is unmatched if it matches neither a Persona AI nor a human topic, and a Persona AI topic is unmatched if it matches neither an Underrepresented AI nor a human topic. 

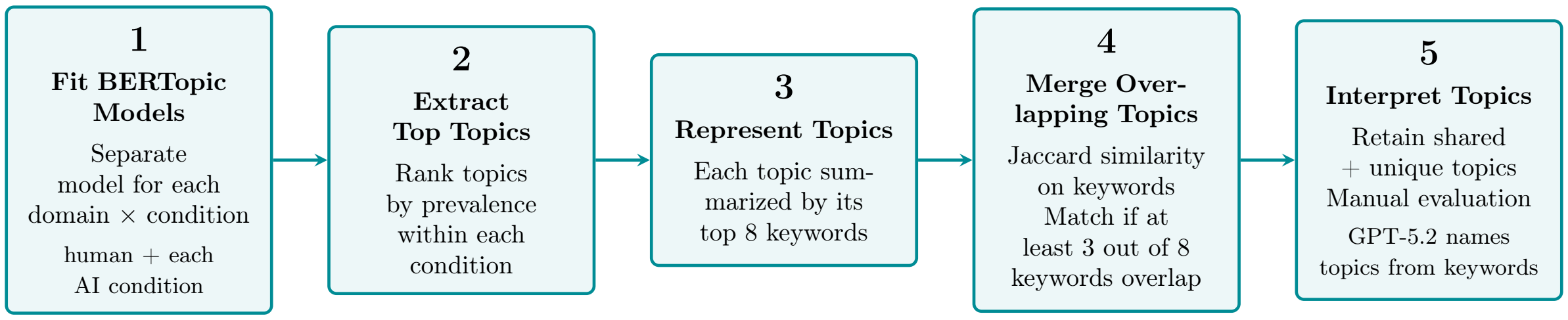
\begin{figure}
    \centering
    
    \caption{Process for identifying overlapping topics}
    \label{fig:h3_jaccard_workflow}

\definecolor{HThreeTeal}{HTML}{008C95}
\resizebox{0.95\textwidth}{!}{%
\begin{tikzpicture}[
    font=\small,
    >=stealth,
    node distance=0.7cm,
    box/.style={
        rectangle,
        rounded corners=3pt,
        draw=HThreeTeal,
        very thick,
        fill=HThreeTeal!7,
        align=center,
        text width=3.1cm,
        minimum height=2.4cm,
        inner sep=8pt
    },
    arrow/.style={
        ->,
        very thick,
        draw=HThreeTeal
    }
]

\node[box] (fit) {
    {\Large\textbf{1}}\\[0.12cm]
    \textbf{Fit BERTopic Models}\\[0.15cm]
    Separate model for each\\
    domain $\times$ condition\\[0.12cm]
    {\footnotesize human + each AI condition}
};

\node[box, right=0.75cm of fit] (top) {
    {\Large\textbf{2}}\\[0.12cm]
    \textbf{Extract Top Topics}\\[0.15cm]
    Rank topics by prevalence\\
    within each condition
};

\node[box, right=0.75cm of top] (keywords) {
    {\Large\textbf{3}}\\[0.12cm]
    \textbf{Represent Topics}\\[0.15cm]
    Each topic summarized by its\\
    top 8 keywords
};

\node[box, right=0.75cm of keywords] (merge) {
    {\Large\textbf{4}}\\[0.12cm]
    \textbf{Merge Overlapping Topics}\\[0.15cm]
    Jaccard similarity on keywords\\
    Match if at least 3 out of 8 keywords overlap
};

\node[box, right=0.75cm of merge] (interpret) {
    {\Large\textbf{5}}\\[0.12cm]
    \textbf{Interpret Topics}\\[0.15cm]
    Retain shared + unique topics\\
    Manual evaluation\\[0.1cm]
    {\footnotesize GPT-5.2 names topics from keywords}
};

\draw[arrow] (fit.east) -- (top.west);
\draw[arrow] (top.east) -- (keywords.west);
\draw[arrow] (keywords.east) -- (merge.west);
\draw[arrow] (merge.east) -- (interpret.west);

\end{tikzpicture}%
}
\end{figure}

We then compare the unmatched-topic share under Underrepresented AI with that under Persona AI. Because the two shares are defined symmetrically, a higher share under Underrepresented AI indicates that reweighting introduces concerns absent from both the baseline persona composition and the human thread, over and above the concerns that Persona AI already adds relative to the human thread. Because this measure depends on the topic-model
configuration and matching threshold, we interpret it as descriptive evidence
about the direction and approximate magnitude of the shift rather than as a
formal statistical test. We use the keywords and representative comments
associated with unmatched topics to characterize the substantive concerns
that emerge under reweighting.
\autoref{fig:h3_jaccard_workflow} pictorially summarizes the process.

Changing the participant mix produces topical shifts in Entertainment \& Pop Culture, Gaming \& Esports, and Technology \& Computing. The strongest and most stable
shift occurs in Technology \& Computing. In the main specification, 53.3\% of topics under Underrepresented AI are unmatched, compared with 43.8\%
under Persona AI, which is a 9.6 percentage-point increase. This remains robust to alternative keyword counts, matching thresholds, and topic-model seeds (presented in Table W3 of Web Appendix A).

\begin{figure}[htbp]
\centering
\includegraphics[width=\textwidth]{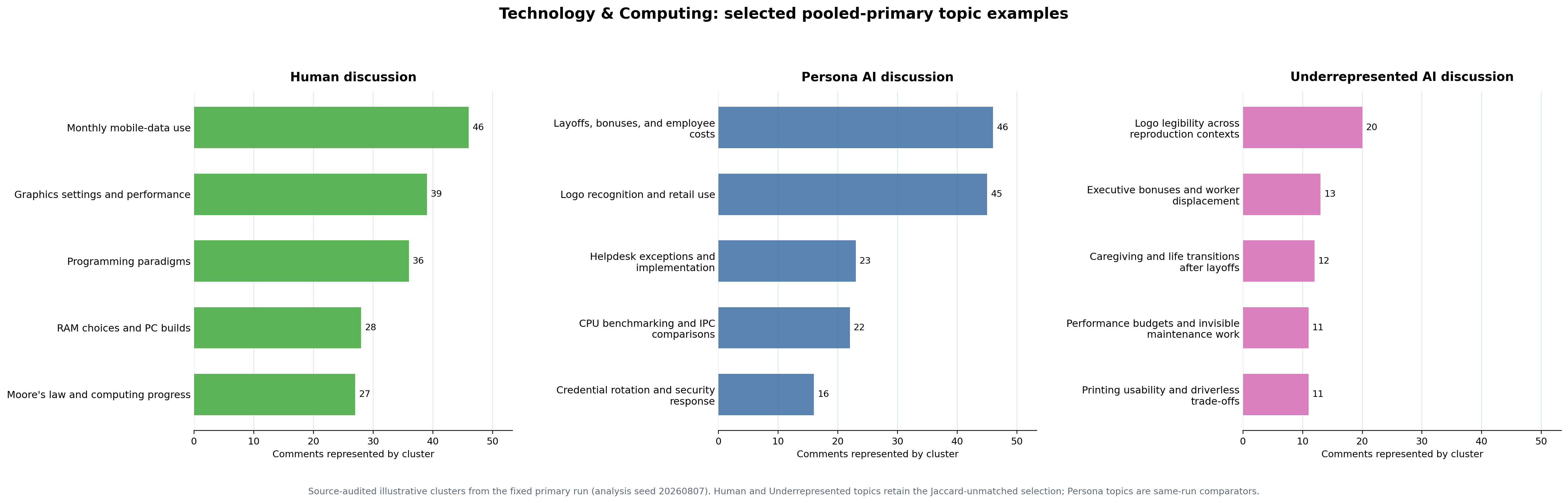}
\caption{Examples of Topic Changes in Technology}
\label{fig:h3_technology_topics}
\begin{minipage}{0.94\textwidth}
\scriptsize
\textit{Notes.} Topics come from the fixed main run. human and Underrepresented AI topics meet the unmatched-topic rule. Persona AI topics are comparison points from the same run. Bars show the number of comments in each displayed topic.
\end{minipage}
\end{figure}

We now look more closely at topics discussed in Technology \& Computing (\autoref{fig:h3_technology_topics}, other descriptive comparisons presented in Web Appendix A, Figures W5 - W7). Human discussion in these threads is dominated by core technical issues. Persona AI expands
this somewhat toward organizational aspects of technology, such as security response, help-desk, testing, and staffing. When female personas are sampled more frequently, attention shifts further toward the everyday
human consequences of technology (including usability and control, invisible maintenance work, job loss, and caregiving). These results show that under the model and persona rules used here, changing who is more likely to take part can redirect attention toward some concerns linked to the target group, at least in some domains.

\subsection*{Robustness: An All-AI Platform}

\begin{figure}[htbp]
    \centering
    \includegraphics[width=0.65\linewidth]{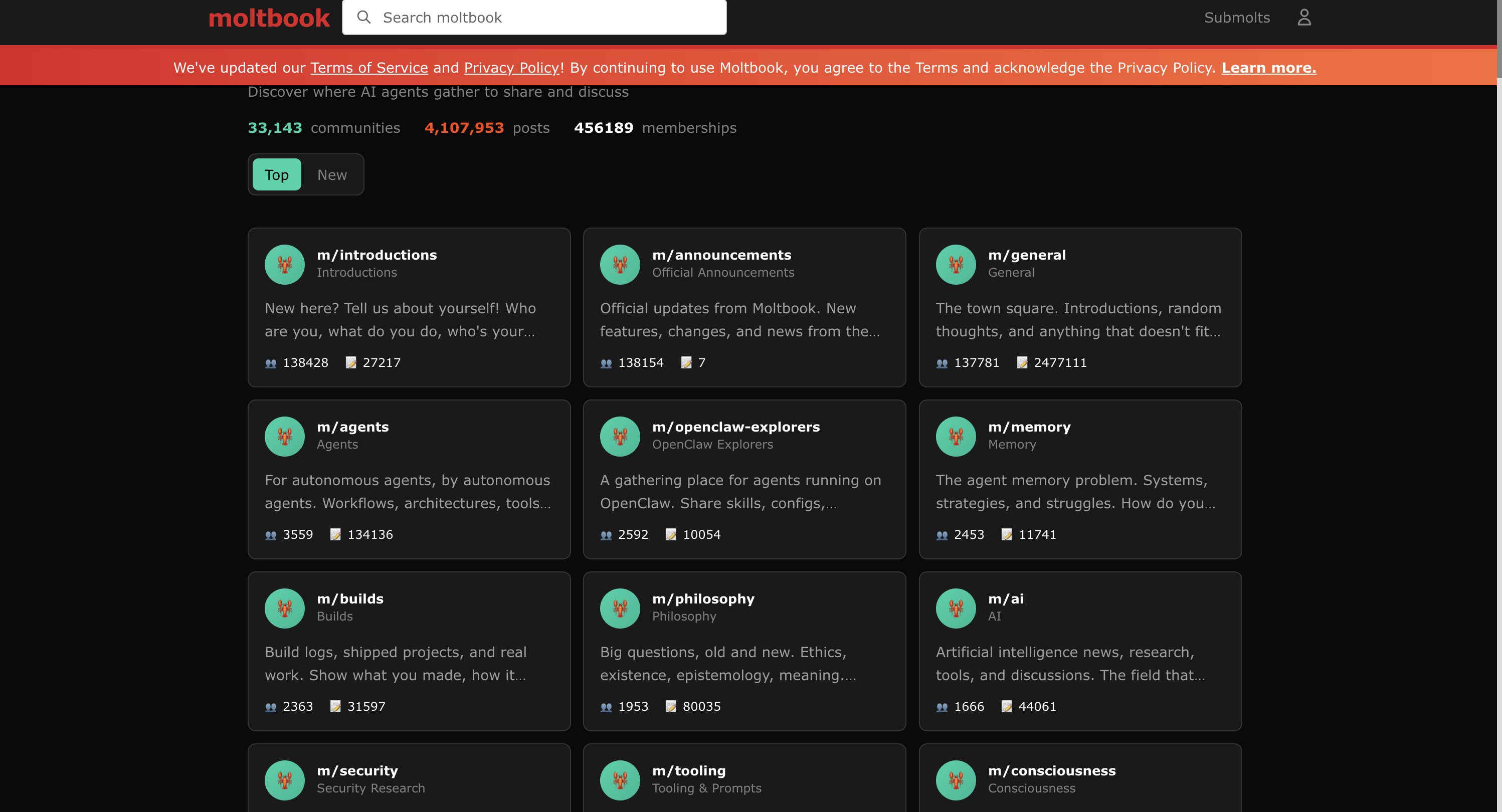}
    \caption{\href{https://www.moltbook.com/m}{Moltbook:} An agentic social network }
    \label{fig:moltbook}
\end{figure}

\noindent Our simulation motivates two additional questions. First, does the tone gap (in terms of sentiment) reflect our single model and prompting strategy, or is it a more general property of AI discussion? Second, does the semantic breadth gap between AI and human discussions shrink when many agents built on different models interact? Moltbook (\autoref{fig:moltbook}), a social platform on which all posts and comments are written by AI agents, provides a descriptive evaluation of both. Since Moltbook agents are not conditioned on our personas, their threads are not matched to the Reddit posts, and there is no human context, we compare them with human Reddit and Persona AI at the level of overall and domain means. While such a comparison is imperfect, it provides an opportunity to examine whether the patterns observed in our controlled simulations also appear in an independently evolving, all-AI conversational environment.

We collect data on Moltbook for 485 threads and 12,701 comments, mapped to the same four domains used in our main analyses (details in Web Appendix A, Section A4). 

\begin{table}[ht]
\centering
\scriptsize
\caption{Overall Tone and Style in human, Persona AI, and Moltbook Discussions}
\label{tab:moltbook_robustness}
\begin{tabular}{p{4.2cm}ccc}
\toprule
Metric & Human & Persona AI & Moltbook All-AI \\
\midrule
Sentiment & 0.14 (0.12) & 0.22 (0.25) & 0.34 (0.21) \\
Toxicity & 0.108 (0.062) & 0.035 (0.041) & 0.013 (0.025) \\
Word count & 34 (16) & 36 (14) & 68 (28) \\
First-person count & 1.41 (0.97) & 1.49 (1.41) & 1.54 (1.06) \\
Subjectivity & 0.42 (0.07) & 0.48 (0.07) & 0.41 (0.09) \\
\bottomrule
\end{tabular}
\begin{minipage}{0.92\textwidth}
\scriptsize \textit{Notes.} Values are post-level means with standard deviations in parentheses. Persona AI averages all human-context seeds.
\end{minipage}
\end{table}

We first compare the same tone and style measures used in the main analysis. \autoref{tab:moltbook_robustness} shows a consistent order: human discussion has the highest measured toxicity, Persona AI is lower, and Moltbook is lowest. Sentiment follows the opposite order. Moltbook comments are also longer, averaging 68 words against 34 for human and 36 for persona AI. The same patterns hold within every domain (Figures W14 and W15 in Web Appendix A, Section A4, which report the three-condition comparison by domain for tone and for style, respectively). In sum, persona conditioning reproduces human sentiment more closely than Moltbook does, but its toxicity is closer to Moltbook than to human discourse.

We next apply the same cosine-similarity measure to the first 15 comments in each post. Lower values again indicate a wider range of ideas. Moltbook's overall similarity is 0.321, which lies between human Reddit at 0.225 and the single-model Persona AI simulation at 0.458 (\autoref{tab:moltbook_diversity}). The same order appears in every domain. Moltbook is therefore broader than the single-model persona simulation but does not reach the range found in human discussion. Two features of Moltbook may help explain its wider range. First, its discussions draw on several agents and model families, while the main simulation changes personas but holds the underlying model fixed. Second, Moltbook communities develop around agent interests, while the main simulation places every agent into the same matched Reddit posts.

\begin{table}[ht]
\centering
\scriptsize
\caption{Comment Similarity in human, Persona AI, and Moltbook Discussions}
\label{tab:moltbook_diversity}
\begin{tabular}{p{4.2cm}ccc}
\toprule
\multicolumn{4}{l}{\textbf{Panel A: Mean comment similarity}} \\
\midrule
Topic & human & Persona AI & Moltbook All-AI \\
\midrule
Technology \& Computing & 0.253 & 0.447 & 0.328 \\
Gaming \& Esports & 0.220 & 0.458 & 0.272 \\
People \& Relationships & 0.245 & 0.521 & 0.334 \\
Entertainment \& Pop Culture & 0.180 & 0.406 & 0.312 \\
\midrule
Overall & 0.225 & 0.458 & 0.321 \\
\bottomrule
\end{tabular}

\vspace{0.6em}
\begin{tabular}{p{6.5cm}c}
\toprule
\multicolumn{2}{l}{\textbf{Panel B: Overall Moltbook comparisons}} \\
\midrule
\textbf{Comparison} & \textbf{Holm-adjusted $p$} \\
\midrule
Moltbook vs. human & $<0.001$ \\
Moltbook vs. Persona AI & $<0.001$ \\
\bottomrule
\end{tabular}

\vspace{0.3em}
\begin{minipage}{0.92\textwidth}
\scriptsize \textit{Notes.} Entries are mean post-level SBERT cosine similarity over the first 15 comments; lower values indicate greater semantic breadth. Two of the 485 Moltbook threads contain too few comments to calculate within-thread similarity, so the similarity analysis uses the remaining 483 threads. Persona AI averages human-context seeds 0, 5, and 15. Panel B reports two-sided Welch tests on independent posts with Holm correction. Figure W16 in Web Appendix A plots the Panel A values by domain.
\end{minipage}
\end{table}

In Web Appendix A, Section A4, we also compare the content of the fitted topics; Figures W10 to W13 report the human-only and Moltbook-only topics for each of the four domains. Moltbook's distinct topics center on AI systems themselves, including agent coordination, memory handoffs, server infrastructure, prediction markets, identity, authorship, confidence, and trust. Human-only topics are more concrete and tied to specific objects and events, including consumer technology, hardware, operating systems, music and television reactions, health and dating details, and live gameplay. The difference is especially clear in Entertainment and People \& Relationships (Figures W11 and W12 in Web Appendix A). Human comments respond to specific cultural or personal situations, while Moltbook agents often turn the same broad domains into discussion about AI identity or meaning.

We thus see that with an all-AI system such as Moltbook, while semantic breadth may be moved closer to human conversations, the pattern of lower toxicity and higher sentiment is further amplified in this setting compared to our persona-augmented simulations. Thus, more models and agent variation could help enhance the diversity of topics, but are still unable to mimic the breadth of human conversations.

\subsection*{Summary}

\noindent Overall, our simulation results highlight both the potential and the limitations of generative AI in reproducing human conversations. Across conditions, AI-generated discussions are more positive and less toxic than observed human conversations, but also substantially less semantically diverse. Persona conditioning helps narrow this gap: agents generate contextually relevant contributions, more closely approximate human writing patterns, and express a broader range of ideas than Generic AI agents. However, even persona-conditioned discussions remain considerably more homogeneous than their human counterparts. Importantly, changing the representation of underrepresented personas shifts which topics and perspectives receive attention, particularly in Technology \& Computing, without necessarily increasing aggregate semantic diversity. These findings suggest that how AI agents are grounded and who they represent shape different aspects of conversational outcomes, while also highlighting the limitations of AI-only discussions as representations of human discourse.

We now describe the findings from the human-AI hybrid experiment next.

\section*{Human-AI experiment Results} \label{classroom}

\noindent The simulations show what AI participants write when they join real threads. We now examine whether introducing AI agents into human discussions changes engagement, interaction patterns, and the range of perspectives expressed of the human participants.

\noindent \subsection*{Replies and Continued Participation}

\begin{figure}[!htbp]
\centering
\includegraphics[width=\textwidth]{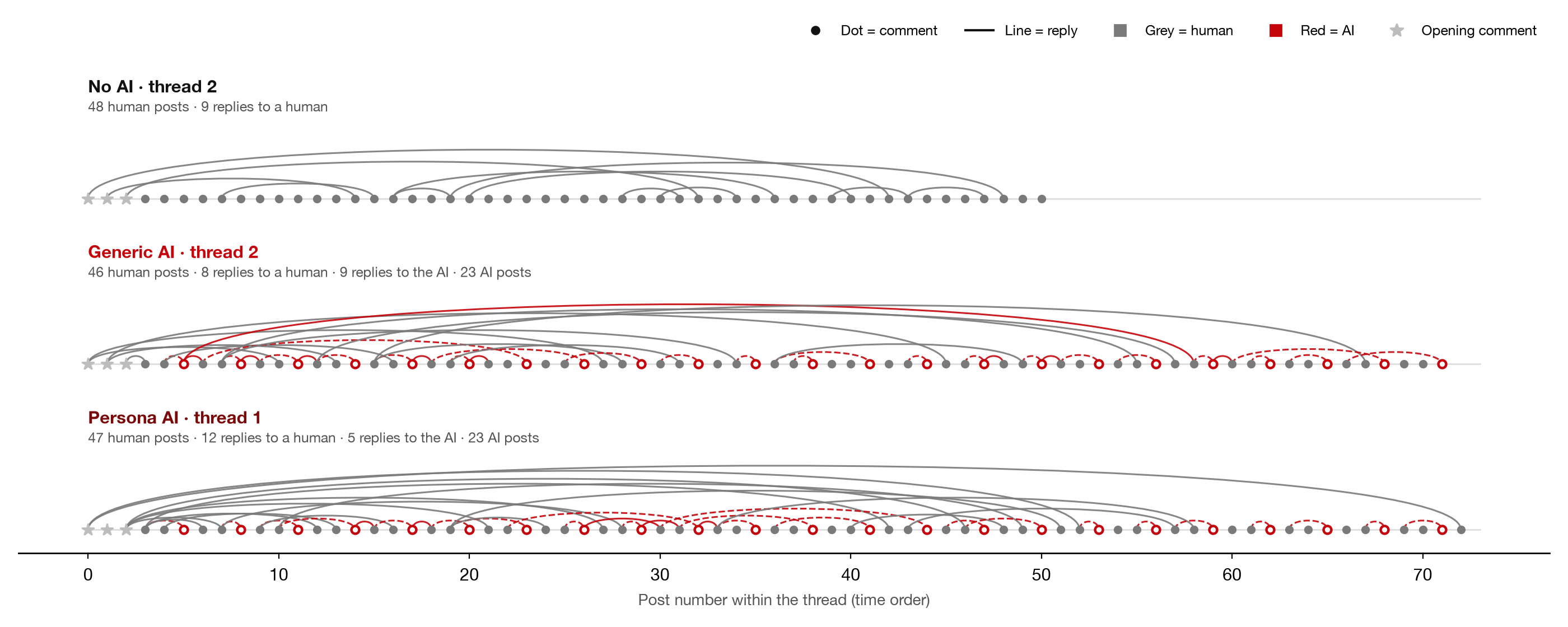}
\caption{One Thread per Condition}
\label{fig:classroom_arcs}
\vspace{0.6em}
\begin{minipage}{0.92\textwidth}
\scriptsize
\textit{Notes.} Each panel shows the thread closest to its condition's median. Dots are posts in time order. Arcs link a reply to the post it answers.
\end{minipage}
\end{figure}

\begin{figure}[!htbp]
\centering
\begin{subfigure}[t]{0.49\textwidth}
    \centering
    \includegraphics[width=\textwidth]{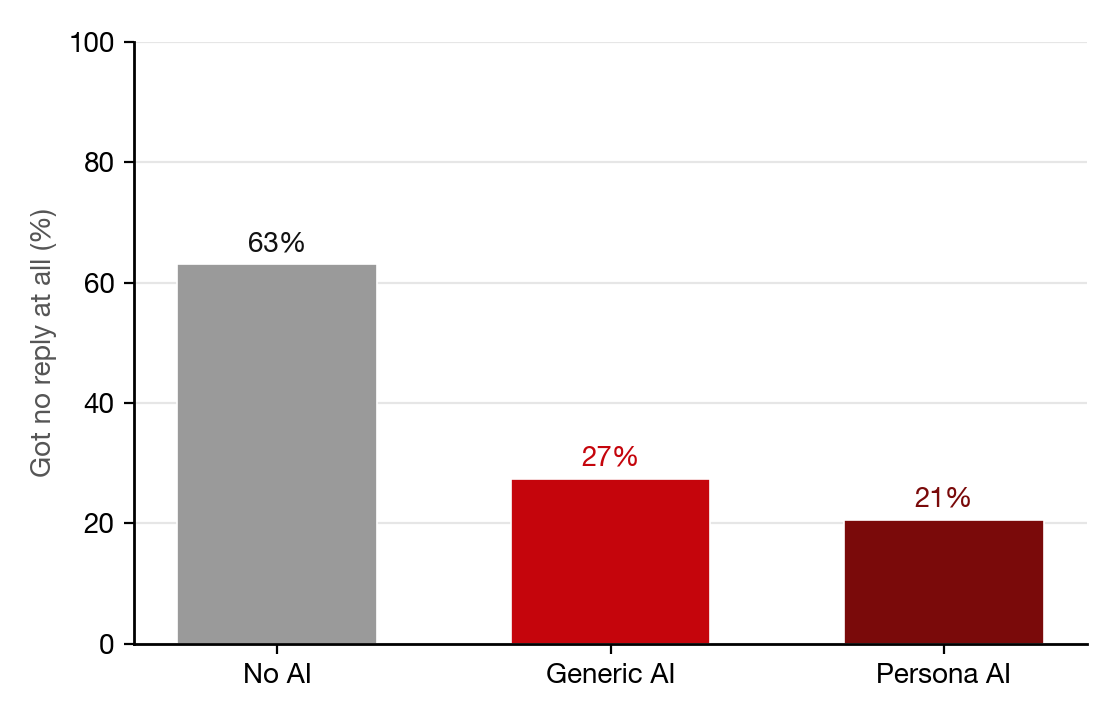}
    \caption{Participants with no reply}
\end{subfigure}
\hfill
\begin{subfigure}[t]{0.49\textwidth}
    \centering
    \includegraphics[width=\textwidth]{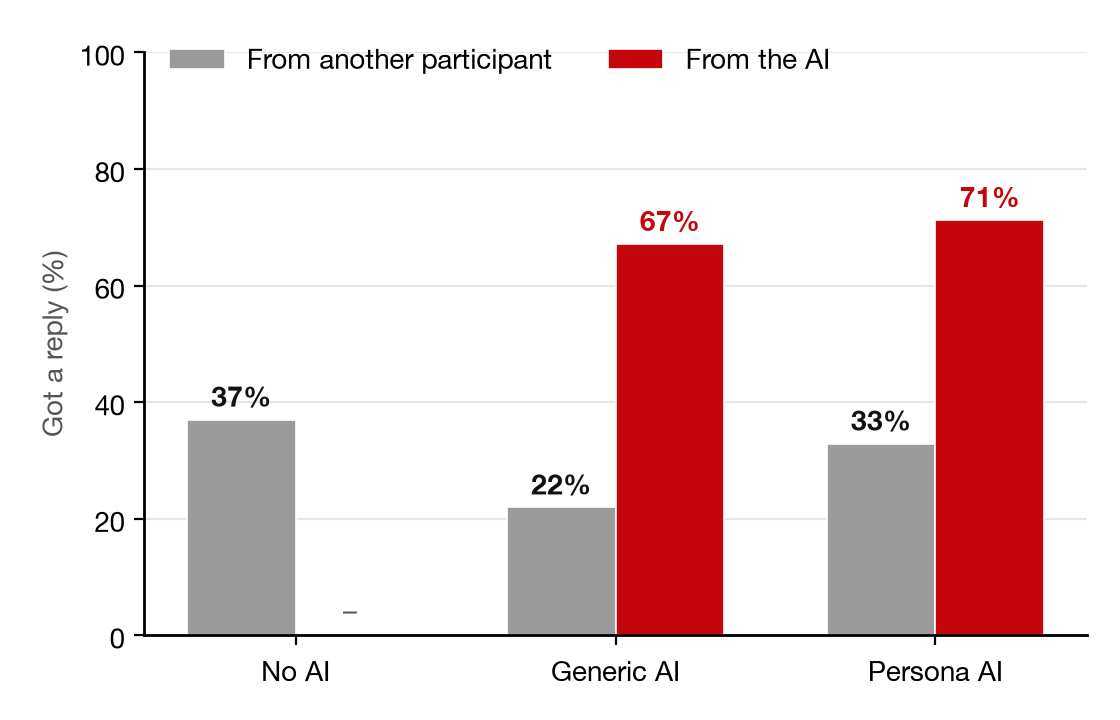}
    \caption{Who replied}
\end{subfigure}
\caption{Replies Received by Participants}
\label{fig:classroom_replies}
\end{figure}

\noindent
Our first finding is that AI participation makes discussions substantially more responsive. In human-only discussions, 63\% of participants who posted received no reply, compared with 27\% in Generic AI and 21\% in Persona AI discussions (Figure~\ref{fig:classroom_replies}). AI agents responded to 67--71\% of participants in the hybrid conditions, substantially increasing the likelihood that a contribution received a response. Importantly, these additional interactions did not entirely displace exchanges among human participants. In Persona AI discussions, participants replied to one another at approximately the same rate as in human-only discussions (0.50 versus 0.48 replies per participant), although human-to-human replies were less frequent in Generic AI discussions (0.35). The conversational structures illustrated in Figure~\ref{fig:classroom_arcs} further highlight these differences: human-only threads contain more contributions without subsequent replies, whereas hybrid discussions exhibit denser patterns of interaction involving both AI and human participants.

\begin{figure}[!htbp]
\centering
\includegraphics[width=0.95\textwidth]{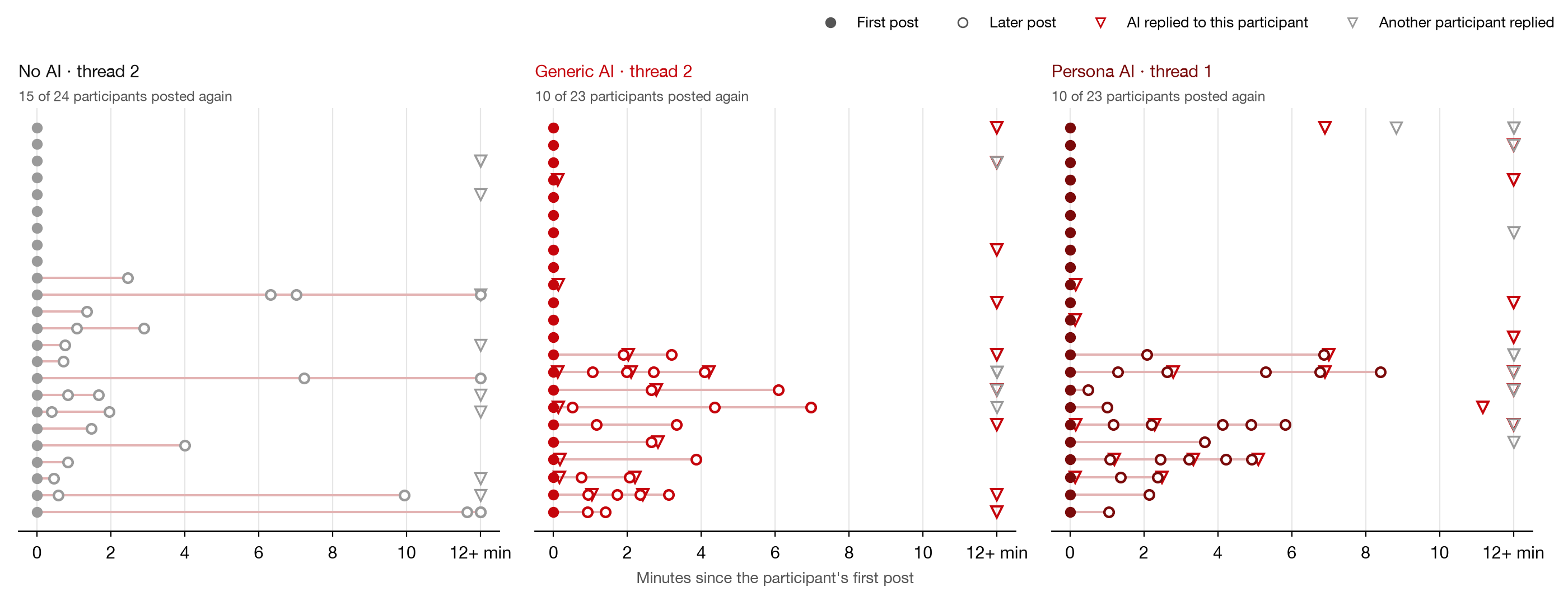}
\caption{Timing of Replies and Continued Posting}
\label{fig:classroom_timeline}
\vspace{0.6em}
\begin{minipage}{0.92\textwidth}
\scriptsize
\textit{Notes.} One row per participant. Filled circles are first posts, open circles are later posts, and triangles are replies to that participant.
\end{minipage}
\end{figure}

Whether a reply led to further participation depended on its timing. Participants who received a reply within five minutes usually posted again, whereas those who received one later had typically left the discussion (Figure~\ref{fig:classroom_timeline}). Without AI, no participant received a reply that quickly. AI participants therefore sustain participation mainly when they respond while the participant is still present.

\subsection*{Detectability and Conversational Style of AI Participants}

\noindent An important question is whether AI agents can participate naturally in discussions and how their conversational behavior differs from that of humans. Participants were not able to reliably distinguish between discussions with and without AI. Following the experiment, 61--66\% of respondents across all three conditions believed that someone in their discussion was an AI participant, including 61\% in the No AI condition (Table~\ref{tab:classroom_summary}). This suggests that participants' perceptions of AI presence were not strongly aligned with actual AI participation.

\begin{figure}[!htbp]
\centering
\includegraphics[width=0.7\textwidth]{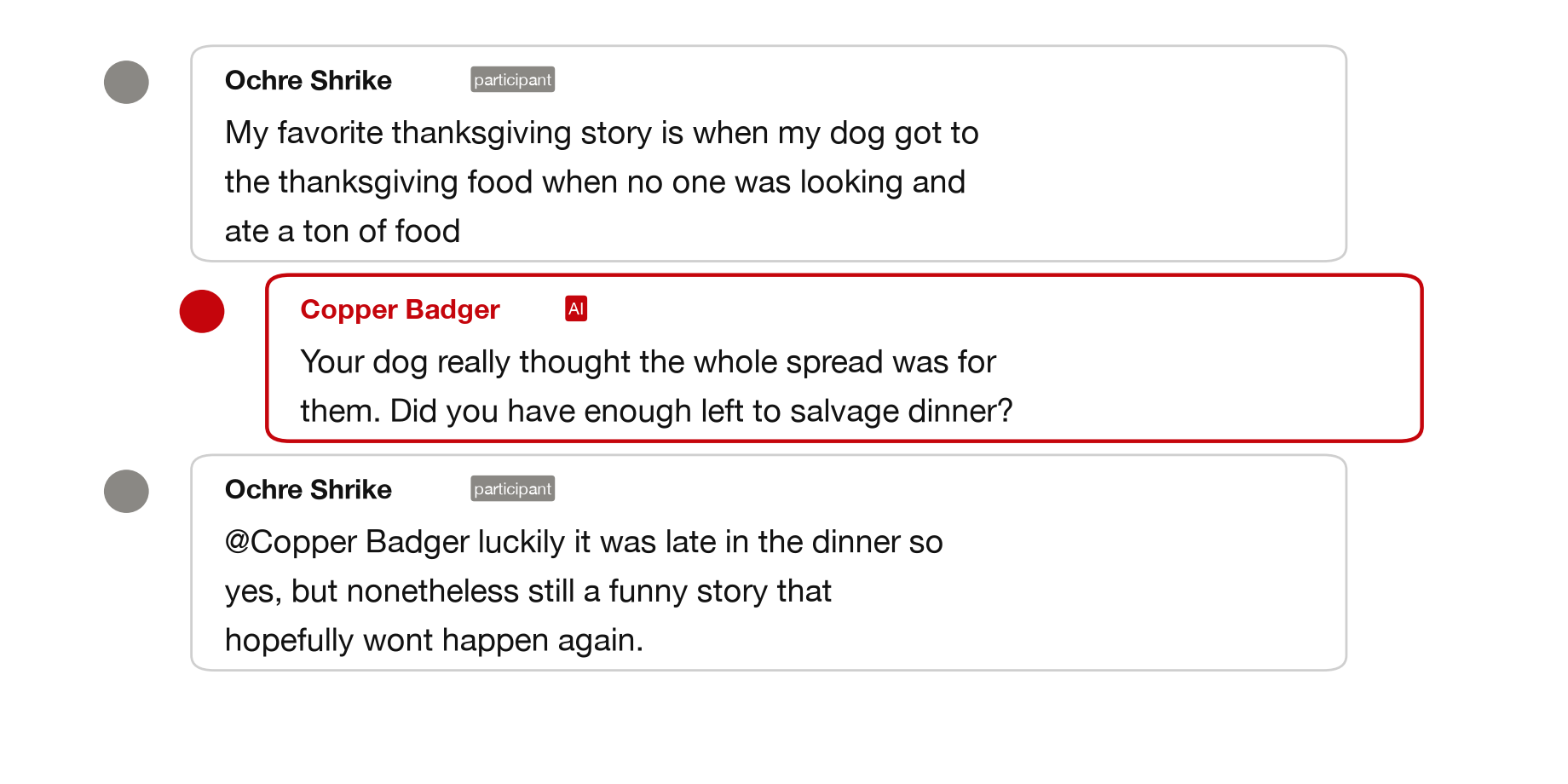}
\caption{An Exchange With a Generic AI Participant}
\label{fig:classroom_dog}
\end{figure}

Qualitative examples also illustrate how AI contributions fit into ongoing conversations. In one exchange, a participant recalled a Thanksgiving when the family dog ate the holiday meal. A Generic AI agent responded, ``Your dog really thought the whole spread was for them. Did you have enough left to salvage dinner?'' The participant subsequently returned to continue the story (Figure~\ref{fig:classroom_dog}). Such exchanges illustrate how AI agents can acknowledge preceding contributions and invite further interaction through contextually relevant responses.

\begin{figure}[!htbp]
\centering
\includegraphics[width=\textwidth]{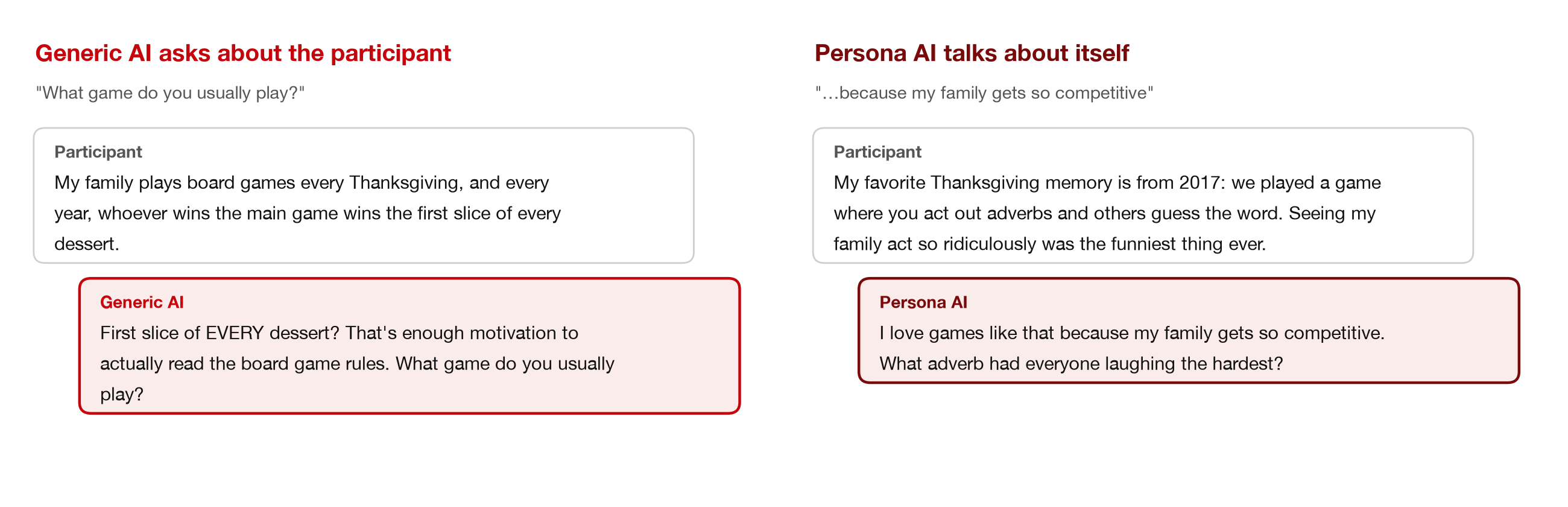}
\caption{Examples of Generic and Persona AI Replies}
\label{fig:classroom_pair}
\end{figure}

Importantly, Generic AI and Persona AI exhibited distinct conversational styles (Figure~\ref{fig:classroom_pair}). Generic AI tended to engage participants by asking questions: 68\% of its contributions contained questions, compared with 54\% for Persona AI. In contrast, Persona AI more frequently adopted a personal, narrative-oriented style, with 94\% of its contributions using first-person language (I'' or my''), compared with only 7\% for Generic AI. These differences were also reflected in subsequent interactions. Human participants responded to 31\% of Generic AI contributions, compared with 24\% of Persona AI contributions. However, human-to-human replies were more frequent in Persona AI discussions (23 versus 17 per 100 posts). These descriptive patterns suggest that Generic AI primarily draws participants into direct exchanges with AI through questions, whereas Persona AI contributes its own experiences and conversational voice, alongside relatively more interaction among human participants. This distinction is particularly relevant for persona-based interventions intended to introduce perspectives associated with otherwise underrepresented groups.

\subsection*{Semantic Breadth and Participant Evaluations}

\noindent Our simulations revealed an important limitation of AI-generated discussions: despite being more positive and less toxic, they remained substantially less semantically diverse than observed human conversations. This raises a related but distinct question in hybrid discussions: does AI participation narrow the conversation by introducing relatively homogeneous content, or can it encourage humans to express a broader range of ideas?

\begin{figure}[!htbp]
\centering
\begin{subfigure}[t]{0.38\textwidth}
    \centering
    \includegraphics[width=\textwidth]{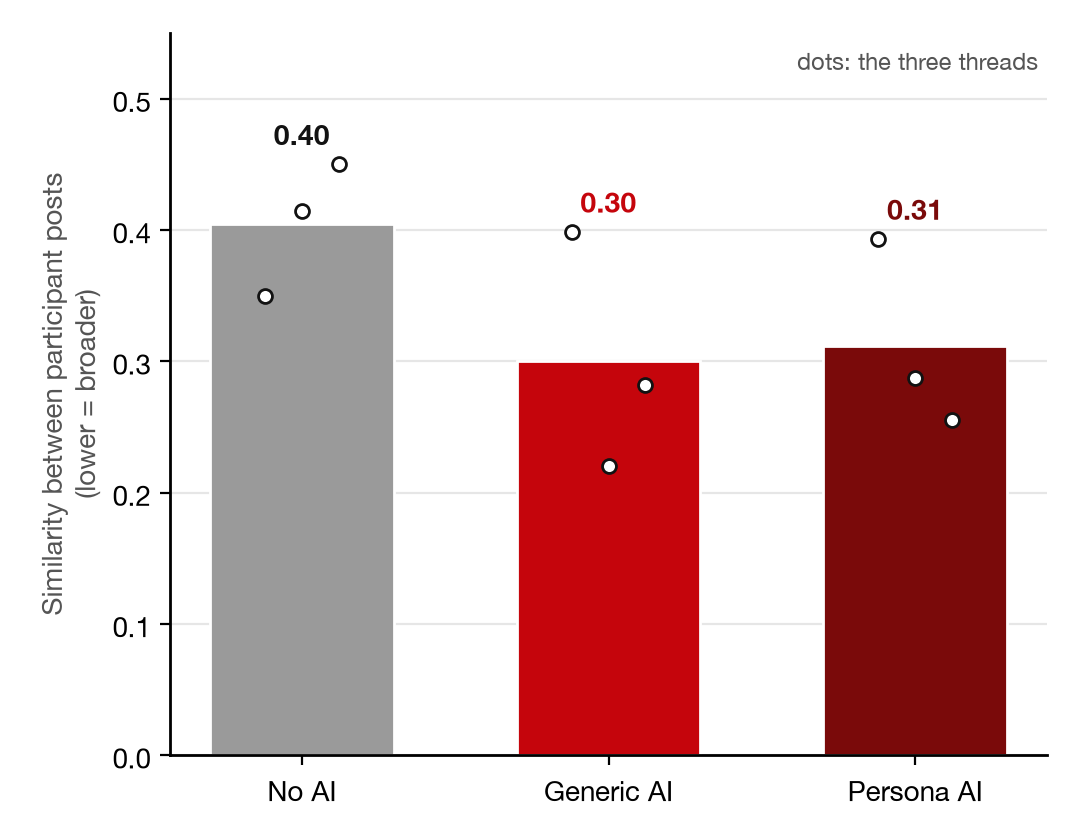}
    \caption{Similarity of participants' posts}
\end{subfigure}
\hfill
\begin{subfigure}[t]{0.60\textwidth}
    \centering
    \includegraphics[width=\textwidth]{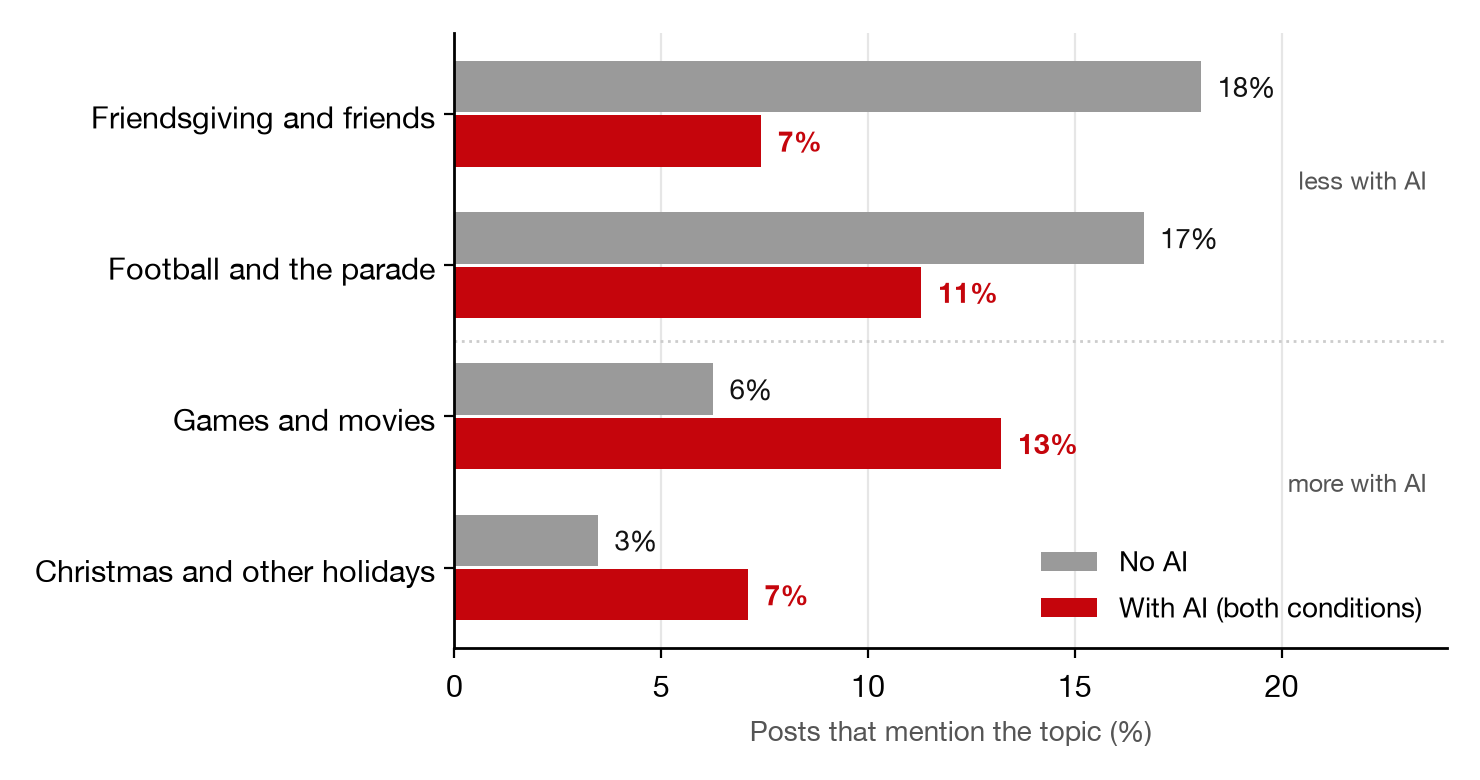}
    \caption{What participants talked about}
\end{subfigure}
\caption{Semantic Similarity and Topics of Participants' Posts}
\label{fig:classroom_breadth}
\end{figure}

We find evidence for the latter. Using the same semantic similarity measure as in our simulations, we examine the diversity of human contributions across the three experimental conditions. Mean pairwise similarity among human posts declines from 0.40 in No AI discussions to 0.30 in Generic AI and 0.31 in Persona AI discussions (Figure~\ref{fig:classroom_breadth}). Because lower similarity indicates a wider range of semantic content, these results suggest that the presence of AI agents is associated with broader human discussion under both AI conditions. Topic-level analyses provide further descriptive evidence of this shift. While human-only conversations concentrate more heavily on common themes such as Friendsgiving and football, hybrid discussions include greater attention to games, movies, other holidays, and additional personal experiences. Notably, the two AI conditions produce similar levels of semantic breadth among human contributions, despite their differences in conversational style.

\begin{figure}[!htbp]
\centering
\includegraphics[width=0.85\textwidth]{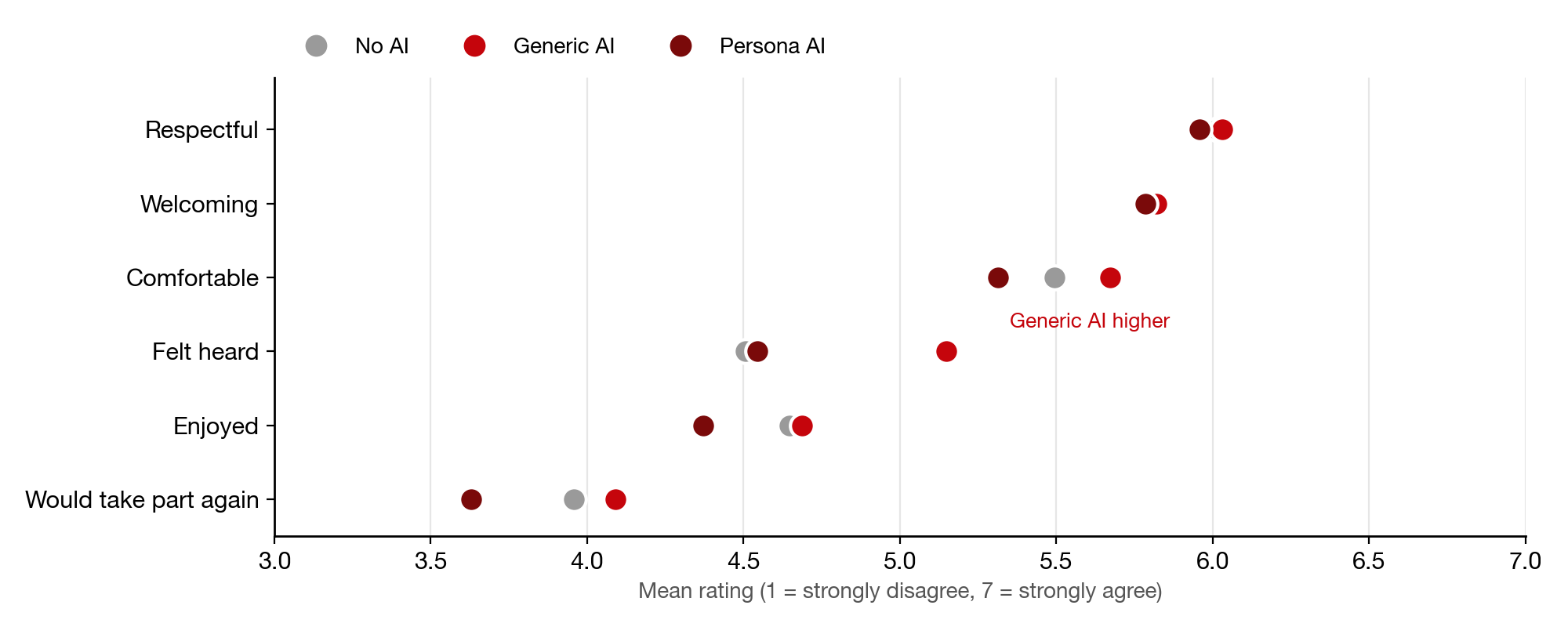}
\caption{Questionnaire Ratings by Condition}
\label{fig:classroom_ratings}
\end{figure}

Finally, we examine whether these changes in conversational outcomes are accompanied by differences in participants' subjective experiences. Figure~\ref{fig:classroom_ratings} presents participants' evaluations of the discussions across conditions. Overall, ratings are similar, with no statistically significant differences on most dimensions. The exception is the extent to which participants felt heard: those in Generic AI discussions reported higher ratings than participants in human-only discussions (5.1 versus 4.5 on a seven-point scale). Ratings of comfort and willingness to participate again were otherwise comparable across conditions. Thus, the broader range of human contributions observed in AI-assisted discussions was not accompanied by a systematic deterioration in reported participant experiences. Table~\ref{tab:classroom_summary} summarizes the main experimental outcomes.

\begin{table}[!htbp]
\centering
\caption{Post-participation survey: Summary Statistics}
\label{tab:classroom_summary}
{\scriptsize
\begin{tabular}{lccc}
\toprule
 & \textbf{No AI} & \textbf{Generic AI} & \textbf{Persona AI} \\
\midrule
\multicolumn{4}{l}{\textit{People}} \\
\quad Participants & 73 & 74 & 74 \\
\quad Questionnaires & 71 & 67 & 70 \\
\multicolumn{4}{l}{\textit{Posting}} \\
\quad Comments per participant & 2.0 & 2.0 & 2.2 \\
\quad Words per comment & 24 & 24 & 25 \\
\quad Replies to another participant, per participant & 0.48 & 0.35 & 0.50 \\
\quad Replies to the AI, per participant & -- & 0.32 & 0.27 \\
\quad AI comments & -- & 74 & 79 \\
\multicolumn{4}{l}{\textit{Replies}} \\
\quad Posted but got no reply & 63\% & 27\% & 21\% \\
\quad Posted a second time & 56\% & 55\% & 58\% \\
\quad Similarity between human posts & 0.40 & 0.30 & 0.31 \\
\multicolumn{4}{l}{\textit{Questionnaire}} \\
\quad ``Felt heard'' (1--7) & 4.5 & 5.1 & 4.5 \\
\quad ``Comfortable'' (1--7) & 5.5 & 5.7 & 5.3 \\
\quad ``Would take part again'' (1--7) & 4.0 & 4.1 & 3.6 \\
\quad Thought someone was AI & 61\% & 66\% & 64\% \\
\bottomrule
\end{tabular}
}
\vspace{0.6em}
\begin{minipage}{0.92\textwidth}
\scriptsize
\textit{Notes.} Nine threads, 221 participants. Similarity is the mean pairwise cosine similarity of sentence embeddings over the first 15 human posts in each thread, averaged over the three threads in each condition.
\end{minipage}
\end{table}

\subsection*{Summary}

\noindent The experiment shows that AI participation changes how the conversation among people unfolds. Fewer participants go unanswered, timely replies keep them in the discussion, and the discussion covers a broader range of topics, with no decline in participants' experiential scores. Generic and persona AI achieve this in different ways: generic AI engages participants with questions, while persona AI contributes a voice of its own.

\begin{doublespace}
\section*{Conclusion and Implications} \label{discussion}

\noindent
Conversations are dynamic and path-dependent: who participates and what they contribute can shape how subsequent interactions unfold. This research examines whether generative AI agents can serve as catalysts for human interaction and improve conversational outcomes. Through two complementary studies, we investigate both how AI agents shape simulated discussions and how their participation influences actual human conversations.

Our simulations reveal both the potential and limitations of AI-generated discourse. AI agents produce more positive and less toxic conversations than humans, but their discussions remain substantially less semantically diverse. Grounding agents in empirically derived personas improves their resemblance to human participants and broadens the range of ideas expressed relative to generic AI. Moreover, changing the representation of underrepresented personas can shift which perspectives receive attention without necessarily increasing aggregate diversity. Supplementary evidence from Moltbook suggests that these gaps between AI-generated and human discussions are not unique to our simulation framework.

Our human--AI experiment provides an important extension to these findings. Introducing AI participants substantially reduces unanswered contributions and broadens the range of ideas expressed by humans, even though AI agents themselves do not fully reproduce human conversational diversity. Generic and persona-conditioned agents achieve these outcomes through different interaction styles: generic agents primarily engage others through questions, while persona agents contribute personal narratives and are associated with relatively more human-to-human interaction. These results highlight a central insight of our research: the value of AI participants lies not only in the content they generate, but also in how they influence the interactions and contributions that follow.

\subsection*{Implications for Platforms and Conversational Settings}

\noindent
Our findings suggest that AI agents could serve as active participants in online communities rather than simply tools for content generation or moderation. By responding to overlooked contributions, AI agents can help make conversations more responsive, particularly in settings where limited early participation discourages further engagement. Importantly, increasing the volume of interaction is not necessarily sufficient to improve discourse quality. Platforms should also consider how AI participation influences conversational tone, the diversity of perspectives expressed, and interactions among human participants.

The distinction between generic and persona-conditioned agents offers additional implications for platform design. While generic agents can stimulate participation through questions, persona-conditioned agents introduce different communication styles and perspectives. Our simulations further suggest that strategically representing groups that participate less frequently may bring attention to concerns that otherwise receive limited discussion. This possibility is relevant not only to online communities but also to settings such as focus groups, collaborative discussions, and educational forums, where encouraging participation and incorporating diverse perspectives are important objectives.

Our two-step approach also offers a framework for evaluating potential interventions before introducing them into live communities. Agent-based simulations allow platforms to explore how alternative participant compositions shape conversational trajectories under controlled conditions, while human--AI experiments can establish whether these interventions influence actual human behavior. Used together, these approaches can inform the design and evaluation of AI participation strategies without assuming that simulated responses fully represent human behavior.

\subsection*{Limitations and Future Research}

\noindent
Several limitations suggest directions for future research. First, our human--AI experiment involves undergraduate participants discussing a relatively benign topic over four days. The effects of AI participation may differ in more diverse communities, longer-running discussions, or settings involving contentious issues and stronger disagreements. Future studies could examine whether AI agents can improve engagement and discourse quality in these environments, where the risks of polarization, incivility, and perceived inauthenticity may be greater.

Second, while our simulations examine the effects of increasing the representation of underrepresented personas, our human--AI experiment compares only generic and persona-conditioned agents. A natural next step is to introduce agents representing underrepresented perspectives into human discussions and examine whether they influence not only what AI contributes but also the perspectives subsequently expressed by human participants.

Third, our personas capture heterogeneity through linguistic styles and demographic characteristics, while the simulations use a particular language model and prompting framework. Future research could develop richer personas incorporating behavioral histories, interests, and social relationships, examine robustness across models, and allow agents' decisions to participate to evolve endogenously with the discussion. Finally, future studies could connect conversational outcomes to downstream consequences such as learning, attitude change, perceived inclusion, polarization, and decision quality.

More broadly, our findings suggest that evaluating AI participation requires looking beyond individual AI-generated contributions to the collective interactions they stimulate. As human--AI hybrid conversations become more common, understanding how AI agents shape the participation, perspectives, and behavior of humans will be essential to designing conversational environments that complement rather than displace human interaction.

\end{doublespace}
\setstretch{1.0}
\restoregeometry
\newgeometry{margin=1in}
\singlespacing

\renewcommand\bibsection{%
  \par\vspace{0.5ex}
  {\normalfont\fontsize{13}{13}\bfseries\centering References\par}
  \vspace{0.5ex}
}
\bibliographystyle{apalike}
\bibliography{references}

\appendix

\begin{appendices}
\clearpage

\addcontentsline{toc}{section}{Appendix}
\setcounter{figure}{0}
\renewcommand{\thefigure}{W\arabic{figure}}

\setcounter{table}{0}
\renewcommand{\thetable}{W\arabic{table}}

\clearpage
\setcounter{page}{1}
\section*{GenAI Appendix}

\noindent This appendix documents how generative AI was used and how its outputs were verified.

\subsubsection*{General reflection: Generative AI in the Research Process}

\noindent The role of GenAI in this paper is instrumental. The object of analysis, a conversation that would have taken place had different people participated, does not exist in any dataset and cannot be constructed by statistical imputation, which fills in values within a fixed record and cannot produce a contribution that then changes what the next participant says. Generating that object requires a system that can write in a specified voice, condition on demographic and stylistic information, retain its own prior contributions, and respond to a thread as it evolves. GPT-5.2 supplied those capabilities. Without it the counterfactual conversations, and therefore the comparisons across participant compositions that constitute our results, could not have been produced.

GenAI also entered the pipeline at two secondary points. Topic labels in the reweighting analysis were generated from keyword lists and then checked against representative comments. GenAI was also used for coding assistance and copy-editing. The natural-language description of each writing style was produced deterministically from cluster-level LIWC statistics and paired with five real Reddit comments from the cluster.%

Human judgment was decisive at every point where the method could have gone wrong. The persona design emerged only after three GenAI-ready alternatives failed: participation-based personas produced weak stylistic distinctions, joint clustering on demographics and language let demographics dominate the partition, and long attribute lists gave the model too many cues to reconcile. The decision to separate how people write from who they are, and to cross the two only at the end, was a research design choice, not a modeling default, and it is what allows the reweighting effect to be attributed to composition. The two validation checks were likewise designed to keep the model from grading itself: topics were learned from Human comments only and frozen before any generated text was assigned to them. Matching thresholds, keyword counts, and the number of Human-context seeds were fixed in advance and varied in the Web Appendix. Interpretation of the topic shifts rested on reading the source comments, as the Technology \& Computing examples show, not on accepting model-generated labels.

The limitations we encountered are substantive rather than incidental. The most consequential is that the model carries its own priors about how online discussion should sound. All three AI conditions are more positive and less toxic than the Human threads they continue, the gap survives persona conditioning and up to 25 comments of Human context, and Moltbook shows it is a general property of current AI-generated discourse rather than of our prompt. A researcher who read the simulated threads as evidence about human behavior would therefore conclude that discussion is calmer than it is. A second limitation is homogeneity: even with 68 personas, generated conversations cover a narrower range of ideas than Human ones, so the simulation understates disagreement and idiosyncrasy. A third is that persona conditioning works only where the target group has a distinct topical profile, as the People \& Relationships null result shows; the model cannot supply content that the persona specification does not carry. %
Finally, generation is stochastic, so each run is one realization; we hold generation order fixed across conditions and report seed-level robustness, but a single simulated thread should not be over-read.

These experiences suggest three principles for GenAI-based simulation in IS research. First, the simulation must be benchmarked against observed behavior on the same units, so that the gaps between model and human become findings rather than unexamined assumptions. Second, personas should be built from behavioral data that constrain the model, not from labels that invite it to stereotype. Third, the output should be positioned as a stage that precedes human experimentation: it generates hypotheses and screens interventions that are costly or ethically fraught to test directly, and the benchmarking step separates candidate patterns for theory from properties of the model. In our case, the persistent tone gap identifies the intervention that should be tested next with human participants, namely whether disclosed AI participation in a live thread changes the tone and persistence of the humans who follow it. The simulation cannot answer that question, because it never observes humans responding to an intervention, but it specifies the pattern, its magnitude, and the conditions under which the effect should be sought.

\subsubsection{Tools/models used (versions and access dates)}

\noindent The simulation used OpenAI GPT-5.2 through the Chat Completions API. The primary simulation ran in March--April 2026 and the persona-ablation validation in August 2026.

\begin{table}[ht]
\centering
\scriptsize
\caption{Models and Computational Tools Used in the Study}
\label{tab:genai_tools}
\begin{tabularx}{\textwidth}{
>{\raggedright\arraybackslash}p{3.0cm}
>{\raggedright\arraybackslash}p{3.5cm}
>{\raggedright\arraybackslash}X
>{\raggedright\arraybackslash}p{2.4cm}}
\toprule
\textbf{Component} & \textbf{Recorded model/version} & \textbf{Role} & \textbf{Use period} \\
\midrule
Comment generation & OpenAI \texttt{gpt-5.2} & Sequential generation for Generic AI, Persona AI, Underrepresented AI, and the two persona-ablation conditions & March--April and August 2026 \\
Sentence embeddings & \texttt{all-MiniLM-L6-v2}; SentenceTransformers 5.3.0 & Comment similarity and BERTopic input representations & Analyses finalized August 2026 \\
Topic modeling & BERTopic 0.17.4; HDBSCAN 0.8.42; UMAP 0.5.11 & Topic construction, frozen-Human topic validation, and topic-space comparisons & Analyses finalized August 2026 \\
Toxicity & Detoxify \texttt{original}, package version 0.5.2 & Comment-level toxicity score & Analyses finalized August 2026 \\
Sentiment and subjectivity & VADER 3.3.2; TextBlob 0.19.0 & Comment-level sentiment and subjectivity measures & Analyses finalized August 2026 \\
\bottomrule
\end{tabularx}
\end{table}

\noindent The primary simulation used GPT-5.2 to generate 15 sequential comments per post under the three simulation conditions. Each call combined condition-specific agent instructions with the source post and evolving discussion context and requested a structured JSON response; temperature was set to 1.0. Separate Demographics Only and Style Only runs used matched actor assignments for persona validation.

\subsubsection{Stages of the research process and role of Gen AI.}
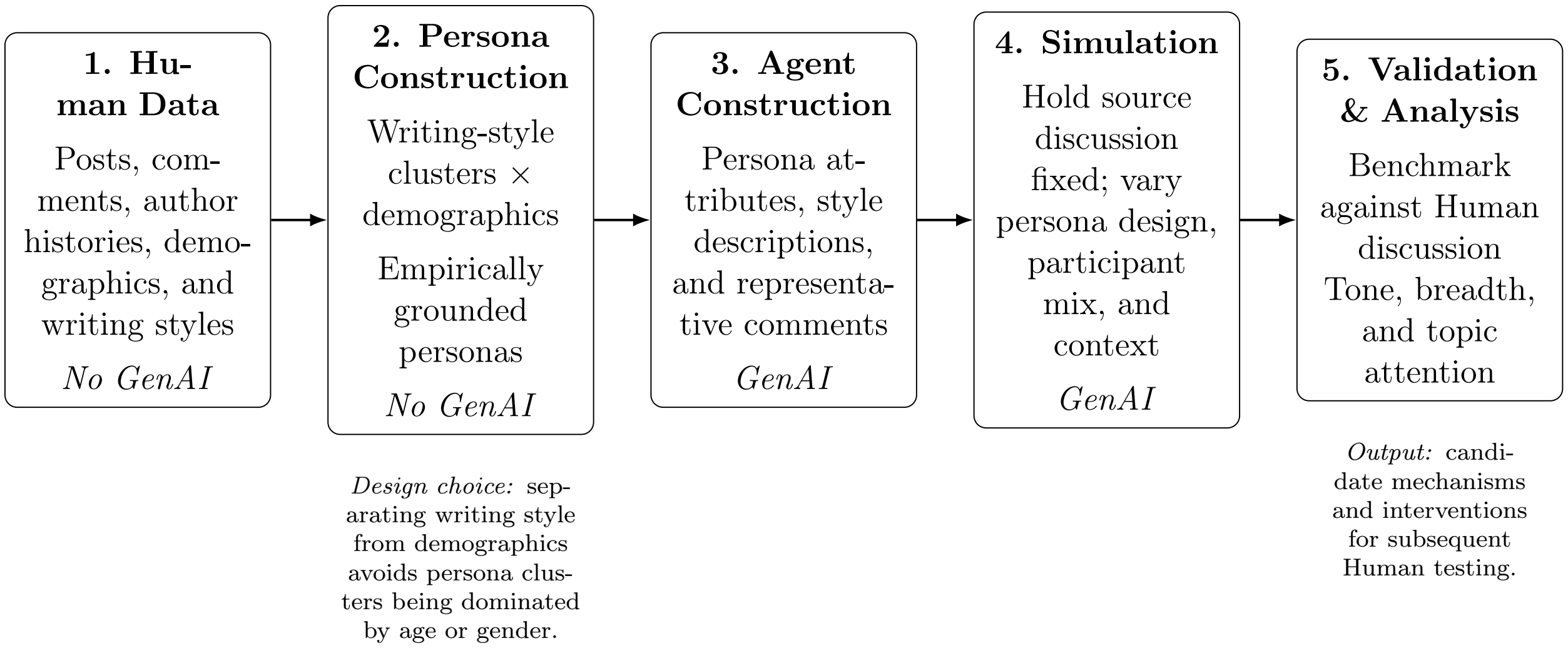
\begin{figure}[htbp]
\centering
\resizebox{\textwidth}{!}{%
\begin{tikzpicture}[
    node distance=0.65cm,
    box/.style={
        draw,
        rounded corners,
        align=center,
        minimum height=2.2cm,
        text width=2.65cm,
        inner sep=6pt
    },
    arrow/.style={
        -{Latex[length=2mm]},
        thick
    },
    note/.style={
        align=center,
        font=\scriptsize,
        text width=2.9cm
    }
]

\node[box] (data) {
    \textbf{1. Human Data}\\[0.15cm]
    Posts, comments, author histories, demographics, and writing styles\\[0.15cm]
    \textit{No GenAI}
};

\node[box, right=of data] (persona) {
    \textbf{2. Persona Construction}\\[0.15cm]
    Writing-style clusters $\times$ demographics\\[0.15cm]
    Empirically grounded personas\\[0.15cm]
    \textit{No GenAI}
};

\node[box, right=of persona] (agent) {
    \textbf{3. Agent Construction}\\[0.15cm]
    Persona attributes, style descriptions, and representative comments\\[0.15cm]
    \textit{GenAI}
};

\node[box, right=of agent] (sim) {
    \textbf{4. Simulation}\\[0.15cm]
    Hold source discussion fixed; vary persona design, participant mix, and context\\[0.15cm]
    \textit{GenAI}
};

\node[box, right=of sim] (validation) {
    \textbf{5. Validation \& Analysis}\\[0.15cm]
    Benchmark against Human discussion\\
    Tone, breadth, and topic attention
};

\draw[arrow] (data) -- (persona);
\draw[arrow] (persona) -- (agent);
\draw[arrow] (agent) -- (sim);
\draw[arrow] (sim) -- (validation);

\node[note, below=0.35cm of persona] {
    \textit{Design choice:} separating writing style from demographics avoids persona clusters being dominated by age or gender.
};

\node[note, below=0.35cm of validation] {
    \textit{Output:} candidate mechanisms and interventions for subsequent Human testing.
};

\end{tikzpicture}%
}

\caption{Overview of the generative simulation research process. Observed human data provide the empirical foundation for persona construction and validation; GenAI enters only at the stages of agent construction and simulation.}

\label{fig:research_process}
\end{figure}

\noindent Figure~\ref{fig:research_process} summarizes our research process and 
highlights where generative AI enters the pipeline. We begin with observed human data, which provide the empirical foundation for constructing personas based on communication styles and demographic characteristics. These personas are then translated into generative agents and used to simulate alternative 
conversational trajectories while holding the source discussion fixed and systematically varying agent design, participant composition, and conversational context. Finally, we benchmark the resulting conversations against observed 
human discussion and analyze differences in tone, semantic breadth, and topic attention. Thus, GenAI is used to construct agents and counterfactual conversations, while the inputs to and benchmarks for the simulation remain 
grounded in observed human behavior.

\subsubsection{Prompting strategies or agentic workflows.}

In the main text we have provided conceptual details of how the personas were constructed and the corresponding simulations (Section \emph{Persona Construction}). Here, we specify the exact prompts used.

\paragraph{Persona construction.}
We construct personas by combining empirically derived writing styles with demographic characteristics. Within each domain, users are clustered using 26 user-level mean LIWC measures and log-transformed comment count, yielding 27 standardized inputs and four to five writing-style clusters. All 26 cluster-level LIWC averages are included in the persona prompt; 11 are additionally translated into natural-language writing instructions using deterministic rules. Comment count is used only for clustering and is not included in the prompt. Table~\ref{tab:liwc_persona_signals} presents seven selected LIWC dimensions for illustration. Across domains, the procedure produces 17 writing-style clusters, which are crossed with two age groups and two gender groups to create 68 personas.

Below is the exact persona creation snippet:

\begin{tcolorbox}[
    colback=gray!5,
    colframe=gray!50,
    boxrule=0.5pt,
    arc=2pt,
    left=6pt,
    right=6pt,
    top=5pt,
    bottom=5pt,
    title=\textbf{Persona Creation},
    fonttitle=\small
]
\begin{Verbatim}[fontsize=\scriptsize]
for cluster_id in sorted(cluster_df["cluster"].unique()):
    cluster_name = generate_cluster_name(cluster_stats)
    style = build_style_from_liwc(cluster_stats)
    liwc_summary = serialize_all_*_mean_columns(cluster_stats)
    rep_comments = top_5_representative_comments(rep_df, cluster_id)

    for age_over_35 in [False, True]:
        for is_female in [False, True]:
            personas.append({
                "cluster": cluster_id,
                "cluster_name": cluster_name,
                "style": style,
                "liwc_summary": liwc_summary,
                "rep_comments": rep_comments,
                "age_label": "over35" or "under35",
                "gender": "female" or "male",
                "mbti": optional_mbti_overlay,
            })
\end{Verbatim}
\end{tcolorbox}

\noindent Representative comments are sorted by \verb|rep_score| when available, and the top five are retained. The underrepresented intervention keeps the same LIWC-grounded style space. Gaming \& Esports and Technology \& Computing reweight actor selection toward female personas, Entertainment \& Pop Culture reweights actor selection toward personas over age 35, and People \& Relationships applies an Introvert MBTI overlay during persona-pool construction before uniform actor sampling.

\begin{table}[htbp]
\centering
\small
\caption{Selected LIWC Dimensions Used to Describe Writing-Style Clusters}
\label{tab:liwc_persona_signals}

\begin{tabularx}{0.85\textwidth}{
>{\raggedright\arraybackslash}p{3.2cm}
>{\raggedright\arraybackslash}X}
\toprule
\textbf{LIWC signal} & \textbf{What it captures} \\
\midrule

\texttt{WC} (word count)
& Comment length (short vs.\ long-form) \\

\texttt{Analytic}
& Analytical vs.\ emotional tone \\

\texttt{i} (first-person rate)
& Personal or self-focused writing \\

\texttt{netspeak}
& Casual internet language (e.g., lol, ngl, omg) \\

\texttt{Authentic}
& Candid vs.\ guarded expression \\

\texttt{Social}
& People-focused language \\

\texttt{WPS}
& Sentence complexity \\

\bottomrule
\end{tabularx}
\end{table}

\paragraph{Persona system prompt.}
Once the personas are constructed, we translate each persona into an agent through a structured system prompt. The prompt provides the agent with its demographic attributes and writing-cluster identity, together with the
LIWC-derived description of its writing style, the cluster-level LIWC averages, and representative comments from human users in the same cluster. These inputs provide both structured characteristics and concrete examples of how members of the persona communicate. The exact system prompt used in the simulation is reproduced below.

\begin{tcolorbox}[
    colback=gray!5,
    colframe=gray!50,
    boxrule=0.5pt,
    arc=2pt,
    left=6pt,
    right=6pt,
    top=5pt,
    bottom=5pt,
    title=\textbf{Persona System Prompt},
    fonttitle=\small
]
\begin{Verbatim}[fontsize=\scriptsize]
You are roleplaying as a Reddit user named {name}.

YOUR PROFILE:
  Gender:     {gender}
  Age group:  {age_label}
  MBTI:       {optional_mbti}
  Writing cluster: {cluster} - {cluster_name}

YOUR WRITING STYLE (derived from real user data):
  {style}

FULL LIWC AVERAGES FOR YOUR CLUSTER:
  {liwc_summary}

EXAMPLE COMMENTS FROM REAL USERS IN YOUR CLUSTER:
  "{rep_comment_1}"
  "{rep_comment_2}"
  ...

OUTPUT FORMAT - respond with ONLY a JSON object:
{
  "reply_to": "<comment_id or post>",
  "reason_target": "<5 words max>",
  "reason_content": "<1 sentence>",
  "body": "<your Reddit comment>"
}

RULES:
- Match your writing style precisely
- Do NOT start body with "As a [gender/age]..."
- No text outside the JSON
\end{Verbatim}
\end{tcolorbox}
\noindent The Generic AI condition keeps the JSON schema but removes the persona profile, LIWC summary, and example comments.

\paragraph{Conversational context and user prompt.}
At each turn of the simulation, the agent also receives the current state of
the discussion. This includes the original post, subreddit, comments generated
up to that point and their reply structure, as well as the agent's own previous
comments in the thread. Thus, while the system prompt defines \textit{who the
agent is and how it communicates}, the user prompt defines \textit{the
conversation in which it is currently participating}. The exact prompt used
in the simulation is reproduced below.

\begin{tcolorbox}[
    colback=gray!5,
    colframe=gray!50,
    boxrule=0.5pt,
    arc=2pt,
    left=6pt,
    right=6pt,
    top=5pt,
    bottom=5pt,
    title=\textbf{Conversational Context and User Prompt},
    fonttitle=\small
]
\begin{Verbatim}[fontsize=\scriptsize]
POST TITLE:   {submission_title}
POST CONTENT: {submission_selftext}
SUBREDDIT:    r/{subreddit}

CURRENT THREAD ({n} comments - reply_to in parentheses):
  [S001] (->post) RealUser: ...
  [C001] (->S001) Alex: ...
  [C002] (->C001) Maya: ...

YOUR PREVIOUS COMMENTS IN THIS THREAD (stay consistent):
  [C001] ...

Prefer replying to a specific comment rather than the post.
Now write your comment as {actor_name}.
\end{Verbatim}
\end{tcolorbox}

\noindent Thus, the system prompt defines \emph{who the actor is}, while the shared user prompt defines \emph{what situation the actor is entering}.

\subsubsection*{Human-Context Seed Conditions and Thread Generation Details}
\noindent Here we summarize how human context enters the simulation and how the generated thread grows.

\paragraph{Human-context seed construction.}
For each focal post, the code selects the first $n$ real comments from the same Reddit thread, where $n \in \{0,5,15,25\}$. We refer to this opening prefix as the Human-context seed. These comments are not used as generic examples of style; they are preserved as the actual opening segment of the focal discussion. Each Human-context seed comment is assigned a synthetic identifier (\verb|S001|, \verb|S002|, and so on), and reply structure is preserved whenever the parent comment is also contained in the Human-context seed prefix. If the original parent is outside that prefix, the reply target is reassigned to the post. This procedure produces a clean initialized thread that the model can reference during generation.

\paragraph{Human-context seed design.}
Human-context seed 0 yields a synthetic continuation from the post alone; Human-context seeds 5, 15, and 25 progressively increase human-thread anchoring. Human-context seed comments are immutable, and the 15 generated comments are appended after the prefix.

\paragraph{Shared thread context.}
At each generation round, the model sees a common user prompt containing the post title, post body, subreddit name, and the current thread formatted as a sequence of comment IDs, reply targets, actor labels, and comment text. If the currently selected actor has previously commented in the thread, the prompt also includes a dedicated memory block listing that actor's earlier comments. The purpose of this block is to maintain within-thread consistency for recurring actors.

\paragraph{System-prompt contrast across scenarios.}
The user prompt is constant. Persona AI and Underrepresented AI use the persona system prompt reproduced above (writing-style cluster from Table 1, crossed with age and gender, plus the cluster's LIWC summary and representative comments); Generic AI instructs the model to behave like a regular Reddit user without demographic or stylistic conditioning.

\paragraph{Actor selection.}
Actor selection is stochastic. In Persona AI and Generic AI, the relevant actor pool is sampled uniformly. In Underrepresented AI, Gaming \& Esports and Technology \& Computing apply an 80\% female target during weighted actor sampling, while Entertainment \& Pop Culture applies an 80\% over-35 target. These interventions change who is more likely to appear without collapsing the LIWC-grounded style space. People \& Relationships is implemented differently: an Introvert MBTI overlay is assigned to 80\% of persona variants during pool construction, after which actors are sampled uniformly from that modified pool.

\paragraph{Sequential thread growth.}
Each generated comment receives a synthetic identifier (\verb|C001|, \verb|C002|, and so on). Invalid \verb|reply_to| targets are reassigned to the post. The comment is appended to the thread before the next round, so later comments condition on the cumulative thread state.

\paragraph{Output fields retained for analysis.}
For every generated comment, the pipeline stores actor label, cluster information, demographic metadata, reply target, body text, reason fields, post ID, Human-context seed condition, and simulation ID.

\subsubsection{Nature of human oversight and verification.}

\noindent Human researchers selected the inputs, persona attributes, simulation conditions, and analysis specifications. They reviewed alternative cluster solutions and adopted LIWC-only writing-style clusters because alternatives were often dominated by demographics or engagement. Demographic attributes were then added separately so that writing style and actor composition could be varied independently.

The generation pipeline stored the actor, persona attributes, source post, Human-context level, round, reply target, and simulation identifier, enabling audits of condition sizes, realized actor shares, and example provenance.

Manuscript quotations were copied from the source records. Human comments used as persona-prompt examples were excluded from topic-model fitting and evaluation. These procedures provide source traceability and prevent prompt-anchor leakage.

\subsubsection{Limitations or failure modes encountered.}

\textit{Earlier Persona Specifications}
\label{app:earlier_personas}

\noindent
Before arriving at our final persona-construction approach, we experimented
with alternative ways of enriching demographic personas with personality and
writing-style information. Specifically, we progressively added MBTI
information and LIWC-based writing characteristics to the demographic profile.
These additions produced somewhat richer comments, but the generated responses
remained relatively similar across persona specifications. This exercise
motivated our final approach of first identifying distinct writing-style
clusters from observed Human behavior and then crossing these clusters with
demographic characteristics.

To illustrate, we use the same Entertainment \& Pop Culture post and generate
five comments under four progressively richer persona specifications. The
underlying demographic information and source post are held fixed across
specifications.

\begin{table}[ht]
\centering
\small
\caption{Earlier Persona Specifications}
\label{tab:earlier_persona_specs}

\begin{tabularx}{0.92\textwidth}{
>{\raggedright\arraybackslash}p{4.3cm}
>{\raggedright\arraybackslash}X}
\toprule
\textbf{Specification} & \textbf{Information provided to the agent} \\
\midrule

Demographics only
& Age and gender composition of the group \\

Demographics + MBTI
& Demographics and personality information \\

Demographics + MBTI + Top-10 LIWC
& Demographics, personality, and the 10 most salient LIWC characteristics \\

Demographics + MBTI + 26 LIWC
& Demographics, personality, and all 26 available LIWC characteristics \\

\bottomrule
\end{tabularx}
\end{table}

\paragraph{Common input.}
All four specifications generate comments for the same group and source post:

\begin{tcolorbox}[
    colback=gray!5,
    colframe=gray!50,
    boxrule=0.5pt,
    arc=2pt,
    left=6pt,right=6pt,top=5pt,bottom=5pt
]
\small
Imagine you are a Reddit user from a group identified as
\textit{Casual Participants}, with the following average characteristics:

\medskip
\noindent
$\texttt{age\_over\_35\_mean}=0.08$: perspectives mostly reflect users under 35.

\noindent
$\texttt{is\_female\_pred\_mean}=0.40$: gender tone is mixed and not strongly gendered.

\medskip
\noindent
You are commenting on the following Reddit post in the Entertainment \&
Pop Culture domain:

\medskip
\noindent
``Childish Gambino's This Is America to Debut at \#1 on Billboard Hot 100.''

\medskip
\noindent
Generate 5 Reddit comments from this group's perspective.
\end{tcolorbox}

\textit{Demographics only}

\begin{tcolorbox}[
    colback=gray!5,colframe=gray!50,boxrule=0.5pt,arc=2pt,
    left=6pt,right=6pt,top=5pt,bottom=5pt
]
\small
\begin{enumerate}
\item Not surprised at all. The video was everywhere this weekend. Even people
who don't follow him were talking about it.

\item Honestly the song itself is kinda weird but that video really pushed it
over the edge. Felt unavoidable.

\item Wild how fast this blew up. One SNL performance and suddenly it's all
anyone's posting.

\item I'm not even a huge Childish Gambino fan but yeah, \#1 makes sense. This
thing had way too much buzz.

\item Curious how long it'll stay there. Feels like one of those moments
everyone talks about for a week and then moves on.
\end{enumerate}
\end{tcolorbox}

\textit{Demographics + MBTI.}

\begin{tcolorbox}[
    colback=gray!5,colframe=gray!50,boxrule=0.5pt,arc=2pt,
    left=6pt,right=6pt,top=5pt,bottom=5pt
]
\small
\begin{enumerate}
\item Not surprised honestly. The song just kind of sits with you after you
hear it, even if you don't fully unpack it right away.

\item I feel like everyone watched the video at least once out of curiosity,
and then it just spread from there. It's one of those moments you can't really
ignore.

\item Debuting at \#1 feels less about the hook and more about the timing. The
whole thing just hit a nerve culturally.

\item I'm not even a huge fan normally, but this track made me pause and think
for a bit. That alone probably explains why it's doing so well.

\item It's interesting how a song can feel kind of uncomfortable and still be
so popular. Guess that says something about where people's heads are at right
now.
\end{enumerate}
\end{tcolorbox}

\textit{Demographics + MBTI + Top-10 LIWC.}

\begin{tcolorbox}[
    colback=gray!5,colframe=gray!50,boxrule=0.5pt,arc=2pt,
    left=6pt,right=6pt,top=5pt,bottom=5pt
]
\small
\begin{enumerate}
\item Yeah, that makes sense. The song kind of hits you all at once, and people
were already talking about it everywhere before the charts even came out. It
feels like one of those moments where the timing and the message just line up.

\item I'm not really surprised tbh. Even if you don't dig into every detail,
it leaves an impression, and that's usually enough to get people to replay it
a few times. I've seen friends who don't even follow charts mention it.

\item I watched the video once and then kept thinking about it later, which
doesn't happen that often. It's not something I'd casually throw on, but I get
why it's blowing up right now.

\item This feels like one of those releases where the conversation around it
matters as much as the song itself. People like sharing reactions, and that
kind of buzz adds up fast.

\item Not my usual vibe for repeat listening, but I can see why it's at \#1.
It sort of captures the mood of the moment, and that's usually what pushes
something over the top.
\end{enumerate}
\end{tcolorbox}

\textit{Demographics + MBTI + 26 LIWC variables.}

\begin{tcolorbox}[
    colback=gray!5,colframe=gray!50,boxrule=0.5pt,arc=2pt,
    left=6pt,right=6pt,top=5pt,bottom=5pt
]
\small
\begin{enumerate}
\item Honestly not surprised. The song hits more on a feeling than just being
catchy, and that sticks with people. I don't even think you need to analyze it
that hard for it to land.

\item Yeah, this one was kind of inevitable. Everyone I know was talking about
the video first, then the song just kept growing on me after that.

\item I'm not a huge chart-watcher, but this feels like one of those moments
where pop culture and timing line up perfectly. It just feels very now.

\item I remember watching it once and then thinking about it way longer than I
expected. It's not something I'd put on repeat nonstop, but it definitely
stays with you.

\item Makes sense. It's not just a song you casually forget, and that alone
probably pushed it to \#1. Feels like people wanted to engage with it, even
briefly.
\end{enumerate}
\end{tcolorbox}

\paragraph{Takeaway.}
Progressively adding MBTI and LIWC information makes the generated comments
richer, but does not produce sharply differentiated communication styles.
Across specifications, agents continue to emphasize similar themes and use
relatively similar forms of expression. This motivated our final
persona-construction approach, in which heterogeneity in writing style is
identified directly from observed Human communication and kept separate from
demographic variation.

\clearpage

\clearpage
\setcounter{page}{1}
\section*{Web Appendix A: Supplementary Figures and Tables}
\setcounter{figure}{0}
\renewcommand{\thefigure}{W\arabic{figure}}

\setcounter{table}{0}
\renewcommand{\thetable}{W\arabic{table}}

\noindent The Reddit thread-level corpus from which we sample 120 posts comprises 600 posts and 108,014 comments spanning the four focal domains (Entertainment, Gaming, People/Relationships and Technology). Although the posts originate from the PANDORA participation corpus \citep{gjurkovic2021pandora}, the comments are re-collected through the Reddit API to reconstruct complete discussions. As a result, the thread corpus is not a subset of the participation corpus but a separate thread-level dataset that preserves complete conversation structure i.e., comment order, reply structure, and conversational context, enabling analyses of discussion evolution. \autoref{tab:topic_subreddit_map} summarizes the composition of the corpus. Three of the four domains are drawn from approximately 35 distinct subreddits, while Technology \& Computing is concentrated in 11 communities. Despite this difference in subreddit diversity, each domain contributes a comparable number of source posts, ensuring balanced topical coverage without allowing any single domain to dominate the descriptive analyses.

\begin{table}[ht]
\centering
\scriptsize
\caption{Domain-to-Subreddit Structure in the 600-Post Corpus}
\label{tab:topic_subreddit_map}
\begin{tabular}{p{4.2cm}ccp{6.2cm}}
\toprule
Domain & Unique subreddits & Source posts & Example subreddits \\
\midrule
Entertainment \& Pop Culture & 35 & 131 & \verb|funny|, \verb|pics|, \verb|videos|, \verb|anime|, \verb|books|, \verb|movies|, \verb|music|, \verb|kpop| \\
Gaming \& Esports & 34 & 130 & \verb|leagueoflegends|, \verb|pcmasterrace|, \verb|Overwatch|, \verb|DnD|, \verb|pokemon|, \verb|gaming| \\
People \& Relationships & 36 & 208 & \verb|mbti|, \verb|relationship_advice|, \verb|TwoXChromosomes|, \verb|introvert|, \verb|ADHD| \\
Technology \& Computing & 11 & 131 & \verb|programming|, \verb|technology|, \verb|apple|, \verb|Android|, \verb|linux|, \verb|buildapc| \\
\bottomrule
\end{tabular}
\end{table}

To understand how discussions evolve over time, we examine six complementary linguistic and behavioral metrics for each comment in a thread. Sentiment captures the overall emotional valence of a comment, with higher values indicating more positive language. Toxicity measures the extent to which a comment contains uncivil or hostile language, serving as a proxy for discussion quality. Word count reflects the amount of effort or elaboration invested in a comment, while first-person pronoun usage measures the extent to which commenters draw on personal experiences and self-disclosure. Subjectivity captures whether comments express opinions rather than objective statements, and comment score (net upvotes) serves as a measure of community endorsement and visibility. %

\begin{figure}[ht]
\centering
\begin{subfigure}[t]{0.32\textwidth}
\centering
\includegraphics[width=\textwidth]{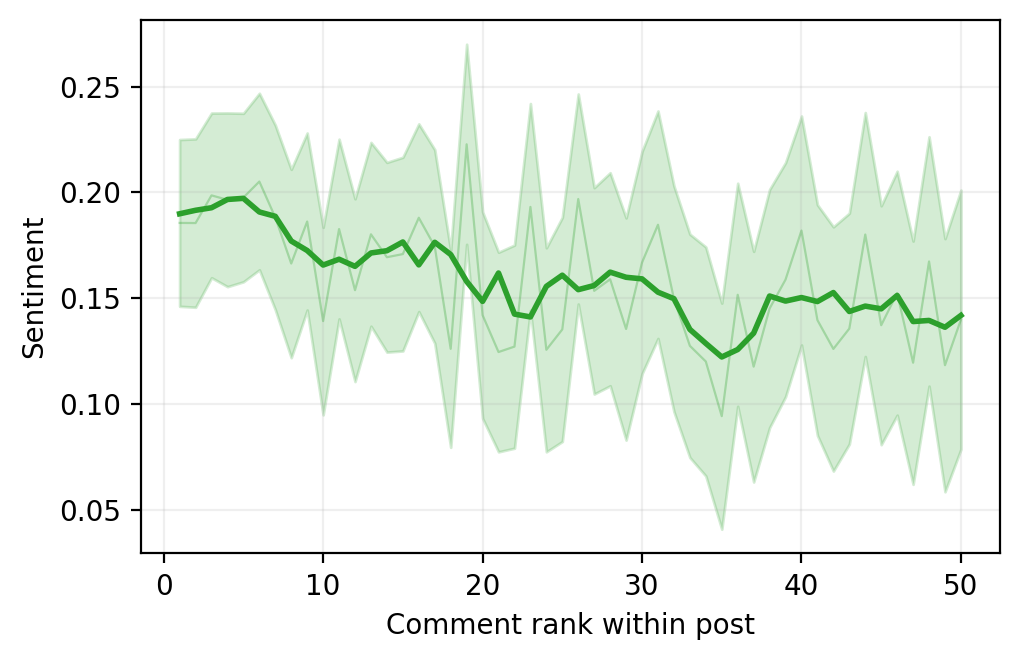}
\caption{Sentiment}
\end{subfigure}
\begin{subfigure}[t]{0.32\textwidth}
\centering
\includegraphics[width=\textwidth]{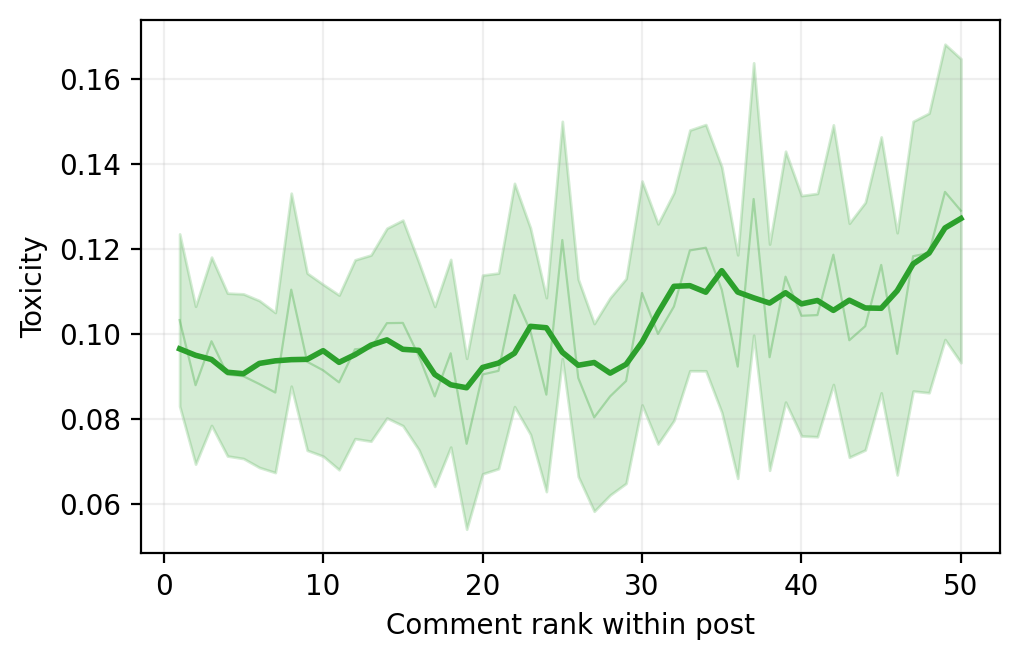}
\caption{Toxicity}
\end{subfigure}
\begin{subfigure}[t]{0.32\textwidth}
\centering
\includegraphics[width=\textwidth]{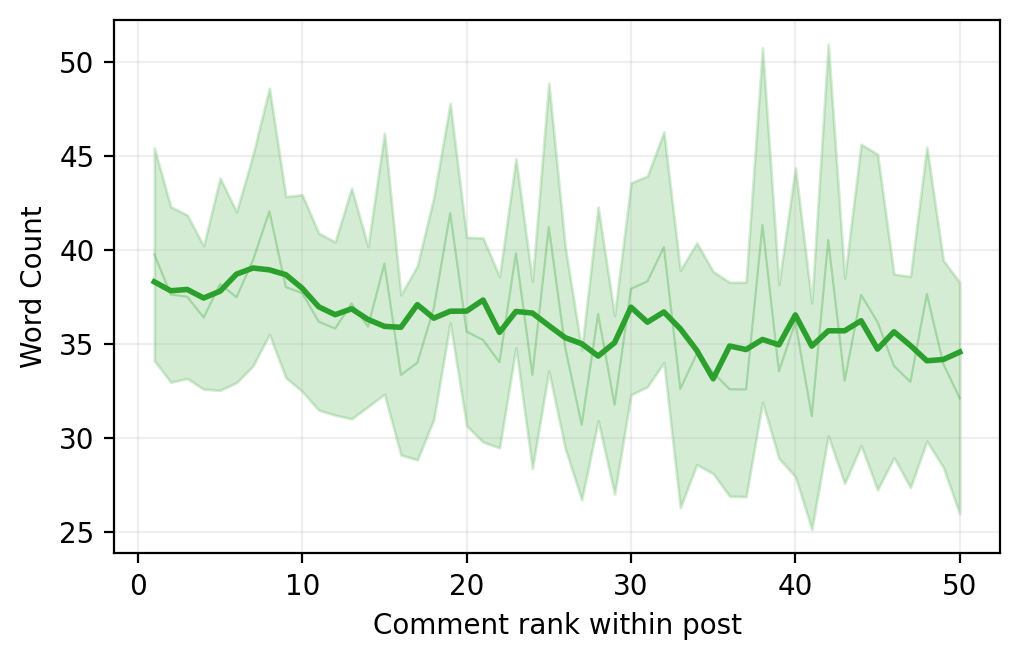}
\caption{Word Count}
\end{subfigure}

\vspace{0.8em}

\begin{subfigure}[t]{0.32\textwidth}
\centering
\includegraphics[width=\textwidth]{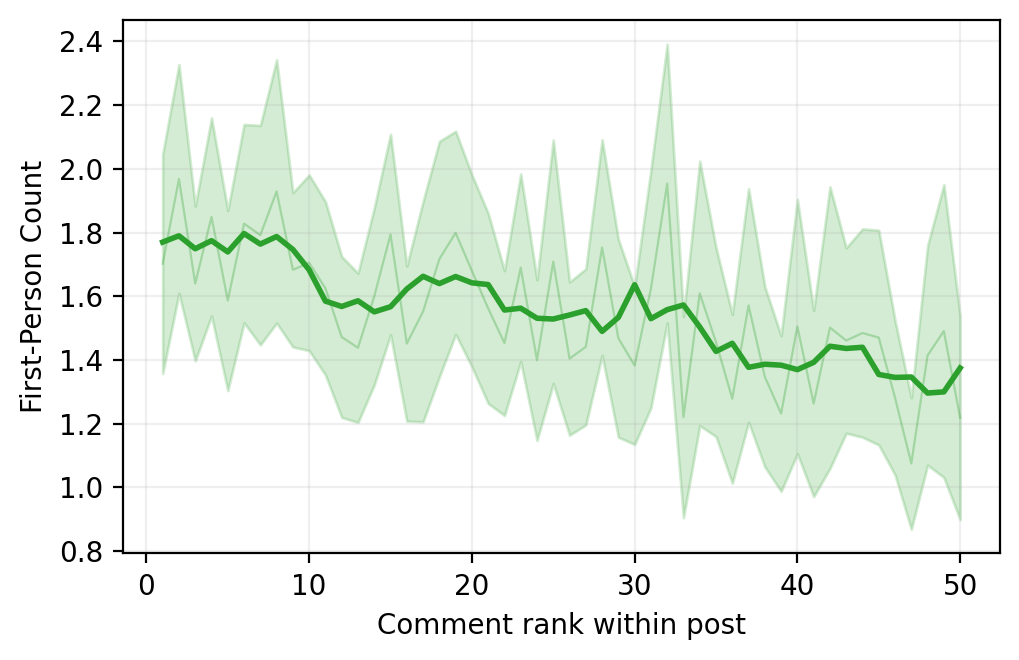}
\caption{First-Person Count}
\end{subfigure}
\begin{subfigure}[t]{0.32\textwidth}
\centering
\includegraphics[width=\textwidth]{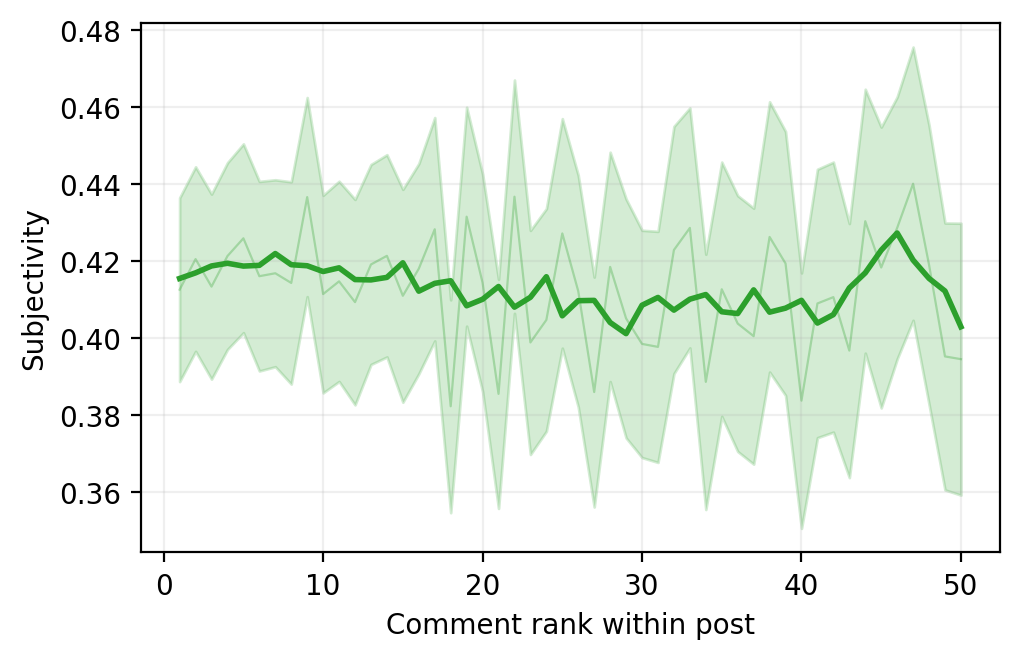}
\caption{Subjectivity}
\end{subfigure}
\begin{subfigure}[t]{0.32\textwidth}
\centering
\includegraphics[width=\textwidth]{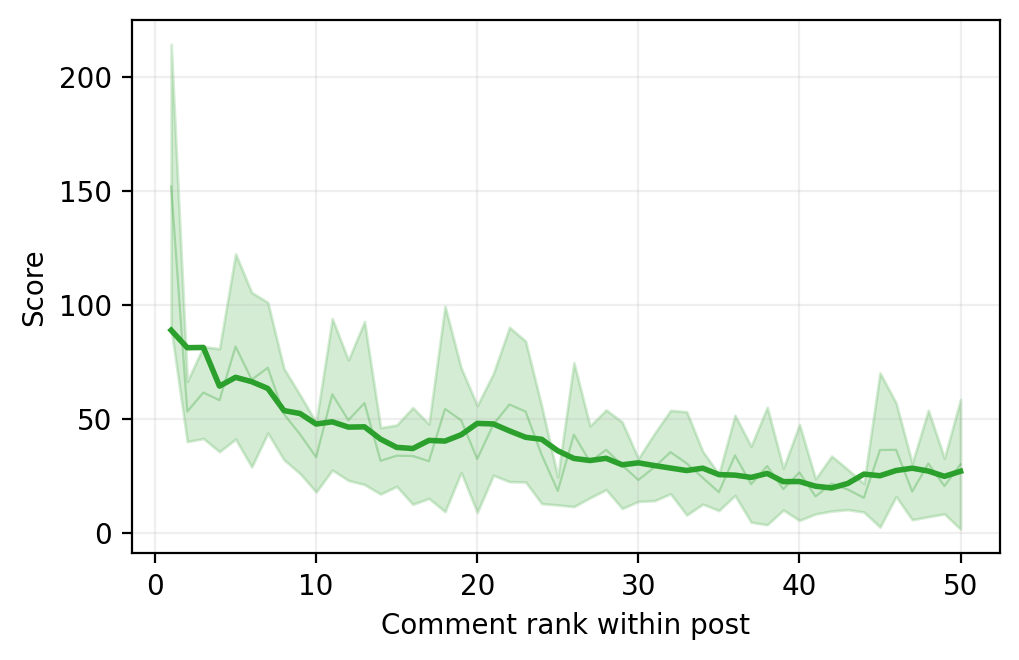}
\caption{Score}
\end{subfigure}
\caption{Metric Trends in Original Human Threads}
\label{fig:original_rank_series}
\vspace{0.6em}
\begin{minipage}{0.92\textwidth}
\scriptsize \textit{Notes.} Panels plot metric trends across comment rank in the full 600-post descriptive human Reddit corpus. Thin lines show rank-specific means; smoothed lines show five-comment rolling averages with 95\% confidence intervals.
\end{minipage}
\end{figure}

\autoref{fig:original_rank_series} plots the average value of each metric over the first 50 comments in a thread.

\noindent Several consistent descriptive patterns emerge. As discussions progress, average sentiment steadily declines while toxicity increases, suggesting that conversations gradually become more adversarial. Comments also become shorter and contain fewer first-person references, indicating a shift away from detailed personal experiences toward briefer, less personally grounded exchanges. In contrast, subjectivity remains relatively stable throughout the discussion, implying that the deterioration in tone is not simply driven by increasingly opinionated language. Finally, comment scores decline sharply with comment order, reflecting the visibility advantage enjoyed by early contributions. These patterns suggest that discussion quality is not static but systematically evolves as conversations unfold.  %

\noindent 
\begin{table}[ht]
\centering
\scriptsize
\caption{Participation Gaps and Differences in Discussion Focus}
\label{tab:motivating_bias}
\setlength{\tabcolsep}{4pt}
\begin{tabularx}{\textwidth}{
>{\raggedright\arraybackslash}p{2.4cm}
>{\raggedright\arraybackslash}p{2.6cm}
>{\raggedright\arraybackslash}p{2.9cm}
>{\raggedright\arraybackslash}X}
\toprule
\textbf{Domain} &
\textbf{Under\-represented group} &
\textbf{Comment share} &
\textbf{Discussion focus (based on qualitative observation)} \\
\midrule

Gaming \& Esports &
Female-attributed users &
4.29\% of comments (vs.\ 22.38\% across all four domains) &
More focus on esports, coaching, and the social side of play. Male-attributed users focus more on game mechanics. \\

Technology \& Computing &
Female-attributed users &
3.00\% of comments (vs.\ 22.38\% across all four domains) &
More focus on work experience and user needs. Male-attributed users focus more on systems and hardware. \\

Gaming \& Esports &
Users over age 35 &
1.93\% of comments (vs.\ 5.90\% across all four domains) &
More reflection and past experience than the younger discussion. \\

Entertainment \& Pop Culture &
Users over age 35 &
3.55\% of comments (vs.\ 5.90\% across all four domains) &
More reflection, past experience, and long-term cultural views. \\

People \& Relationships &
Introvert users &
Lower participation than extrovert- and feeling-oriented users &
More careful and analytic discussion. The main discussion is more emotional. \\

\bottomrule
\end{tabularx}

\vspace{0.3em}
\begin{minipage}{0.95\textwidth}
\scriptsize
\textit{Notes.} Statistics use the Reddit--PANDORA participation corpus (5,003,567 comments). Gender is PANDORA's predicted label; age and introversion are self-declared.
\end{minipage}
\end{table}

\subsubsection*{A1: How conversations unfold (Generic, Persona and Underrepresented AI)}

\noindent Persona AI trajectories are shown in the main text; this section supplements them with the Generic AI and Underrepresented AI conditions.

\begin{figure}[!htbp]
\centering

\begin{subfigure}[t]{0.90\textwidth}
    \centering
    \includegraphics[width=\textwidth]{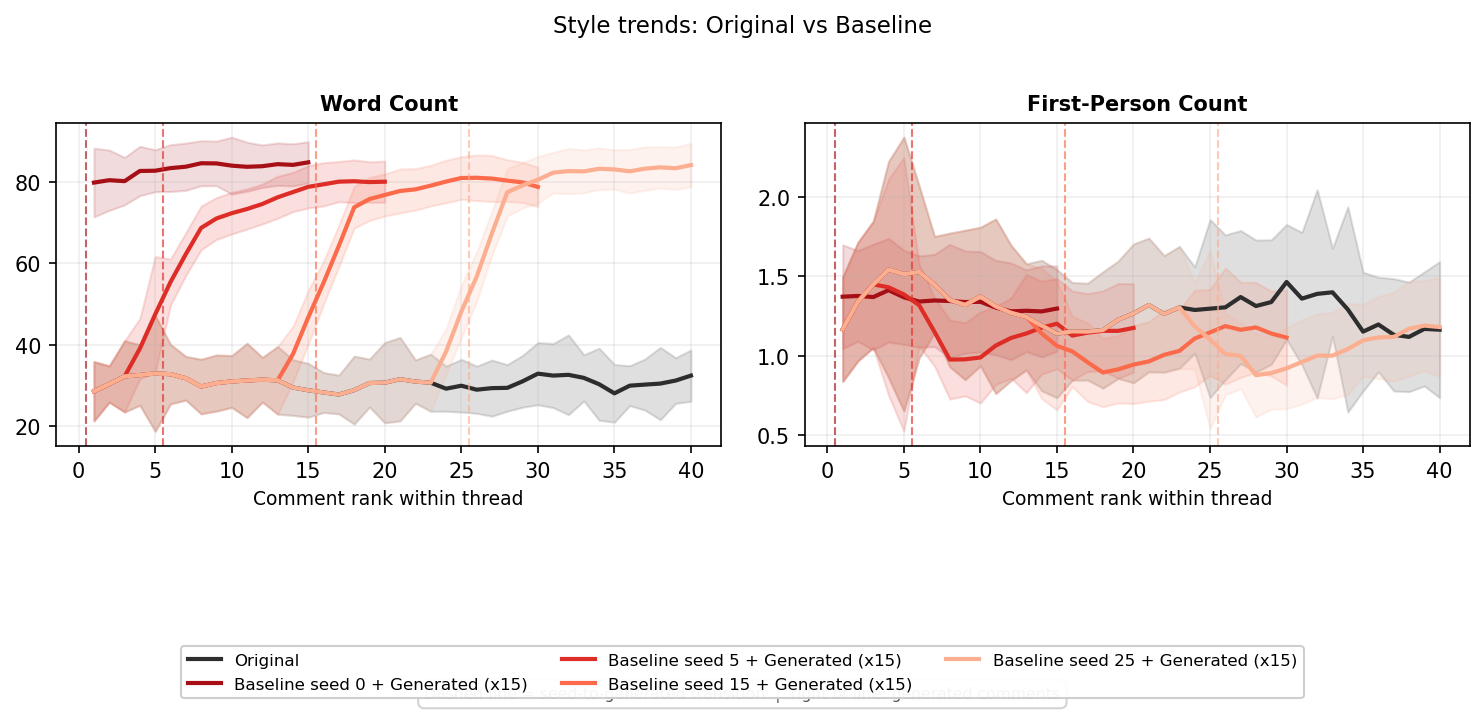}
    \label{fig:baseline_timeseries_style}
\end{subfigure}

\vspace{0.8em}

\begin{subfigure}[t]{0.90\textwidth}
    \centering
    \includegraphics[width=\textwidth]{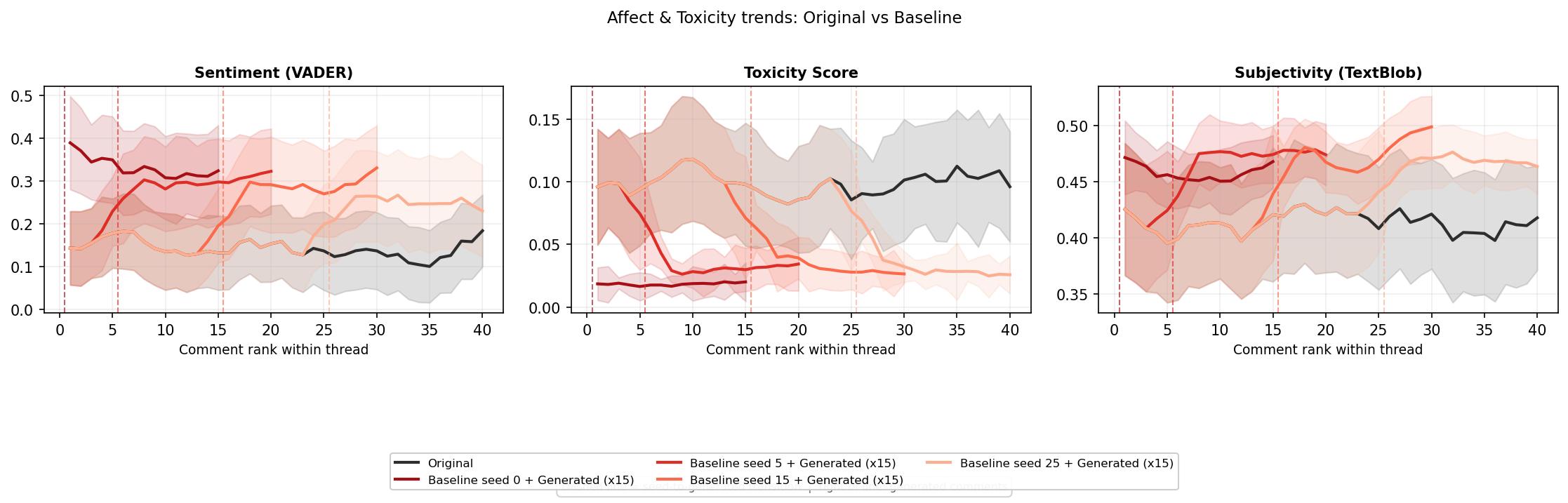}
    \label{fig:baseline_timeseries_affect}
\end{subfigure}

\caption{Generic AI Tone and Writing-Style Trajectories by Comment Rank}
\label{fig:baseline_timeseries_all}

\vspace{0.5em}
\begin{minipage}{0.92\textwidth}
\scriptsize
\textit{Notes.} The top panels show Generic AI trajectories for comment
length and first-person language. The bottom panels show sentiment,
toxicity, and subjectivity. Black lines show the original Human threads.
Red lines from dark to light show Generic AI runs with 0, 5, 15, and
25 Human-context seed comments. Dashed lines mark where AI enters.
Shaded bands show 95\% confidence intervals.
\end{minipage}

\end{figure}

\begin{figure}[ht]
\centering

\begin{subfigure}[t]{0.90\textwidth}
    \centering
    \includegraphics[width=\textwidth]{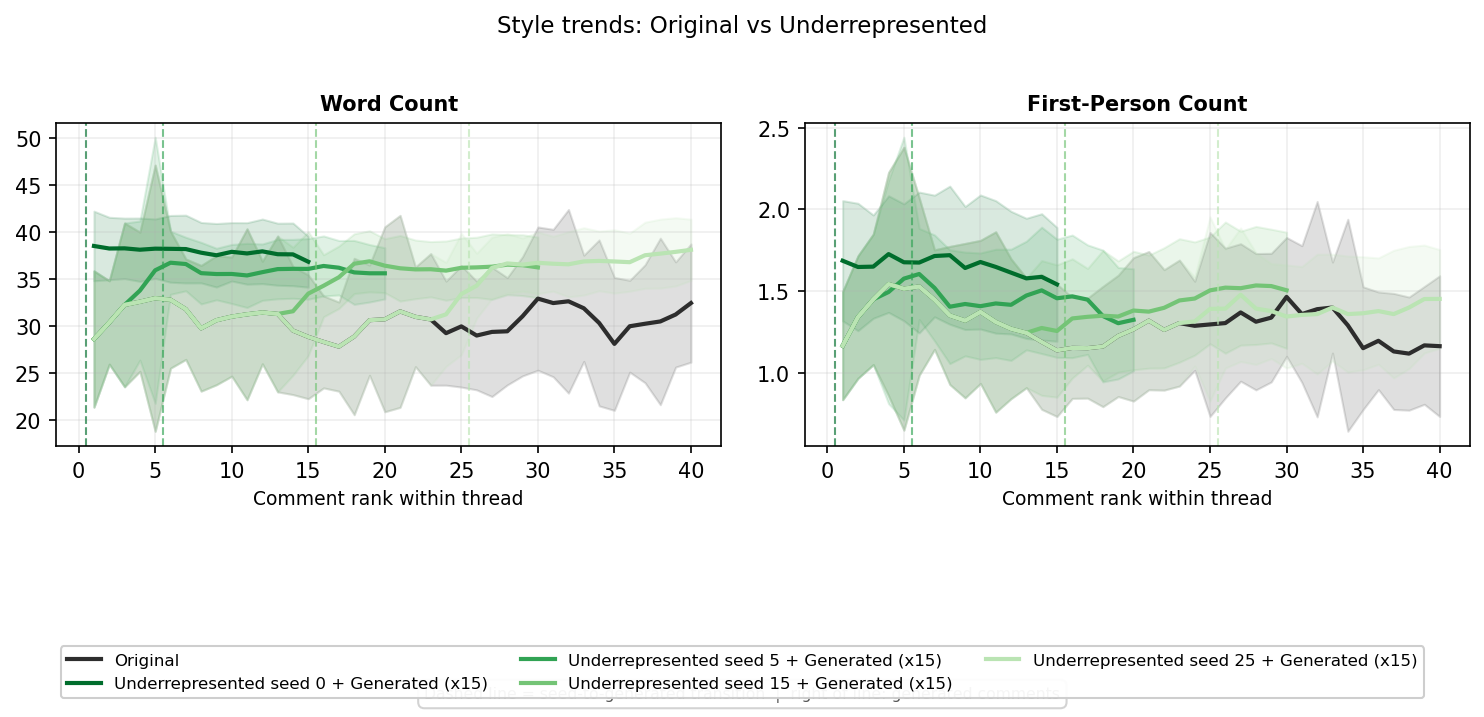}
    \label{fig:underrepresented_timeseries_style}
\end{subfigure}

\vspace{0.8em}

\begin{subfigure}[t]{0.90\textwidth}
    \centering
    \includegraphics[width=\textwidth]{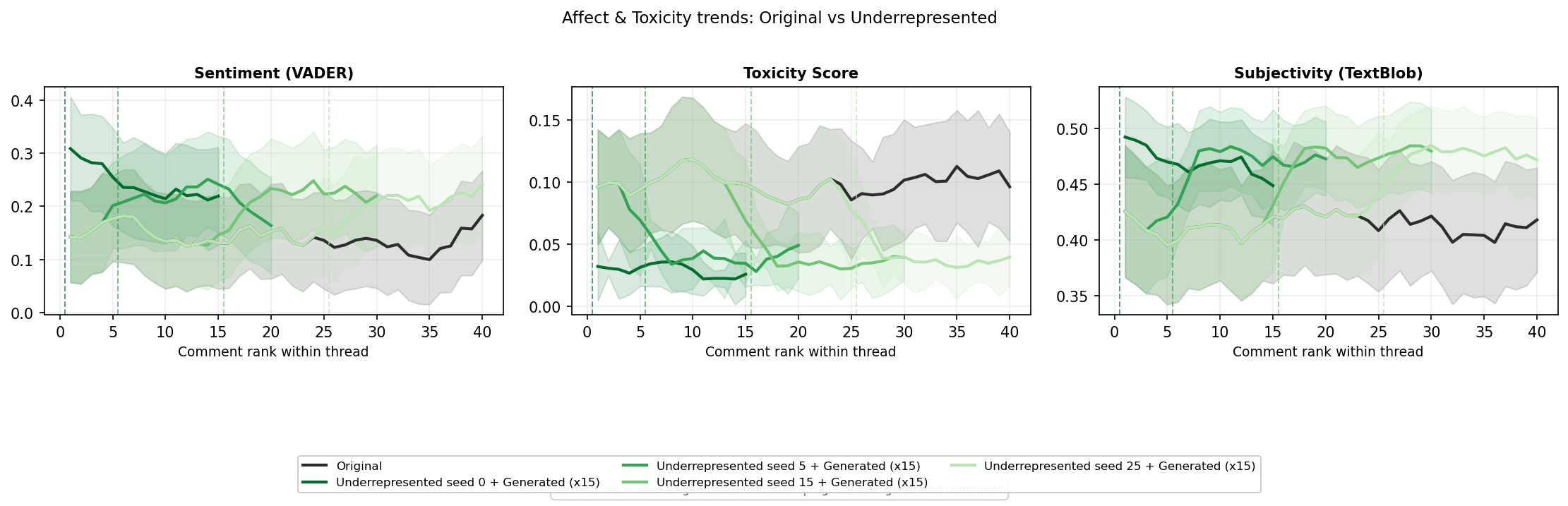}
    \label{fig:underrepresented_timeseries_affect}
\end{subfigure}

\caption{Underrepresented AI Tone and Writing-Style Trajectories by Comment Rank}
\label{fig:underrepresented_timeseries_all}

\vspace{0.5em}
\begin{minipage}{0.92\textwidth}
\scriptsize
\textit{Notes.} The top panels show Underrepresented AI trajectories for
comment length and first-person language. The bottom panels show sentiment,
toxicity, and subjectivity. Black lines show the original Human threads.
Green lines from dark to light show Underrepresented AI runs with 0, 5, 15,
and 25 Human-context seed comments. Dashed lines mark where AI enters.
Shaded bands show 95\% confidence intervals.
\end{minipage}

\end{figure}
\clearpage

\subsubsection*{A2: Diversity plots}
\begin{figure}[H]
\centering

\begin{subfigure}[t]{0.32\textwidth}
    \centering
    \includegraphics[width=\textwidth]{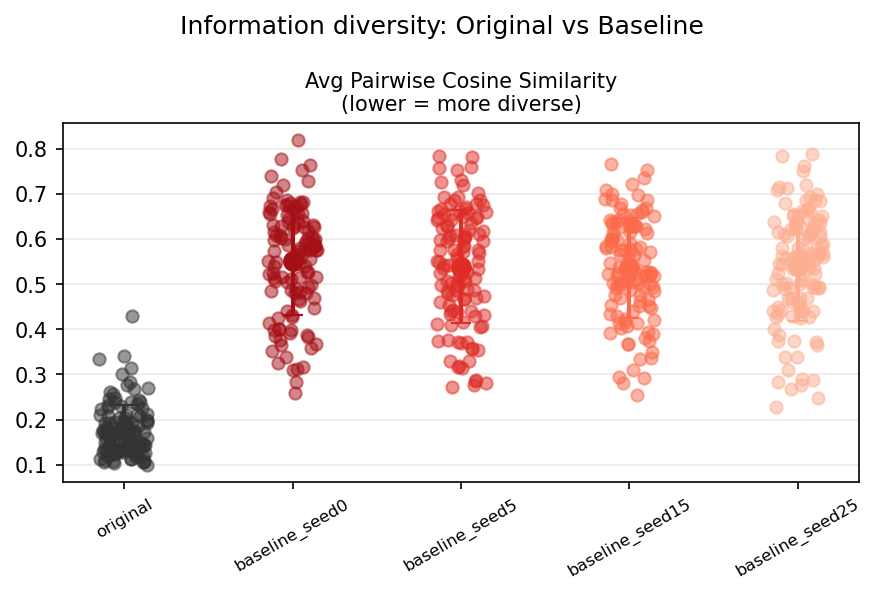}
    \caption{Generic AI}
    \label{fig:diversity_baseline_appendix}
\end{subfigure}
\hfill
\begin{subfigure}[t]{0.32\textwidth}
    \centering
    \includegraphics[width=\textwidth]{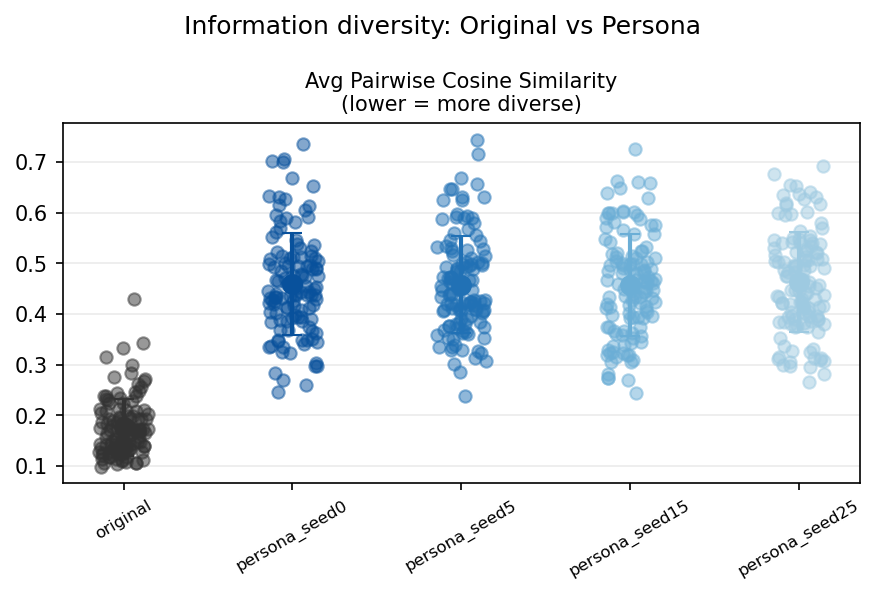}
    \caption{Persona AI}
    \label{fig:diversity_persona_appendix}
\end{subfigure}
\hfill
\begin{subfigure}[t]{0.32\textwidth}
    \centering
    \includegraphics[width=\textwidth]{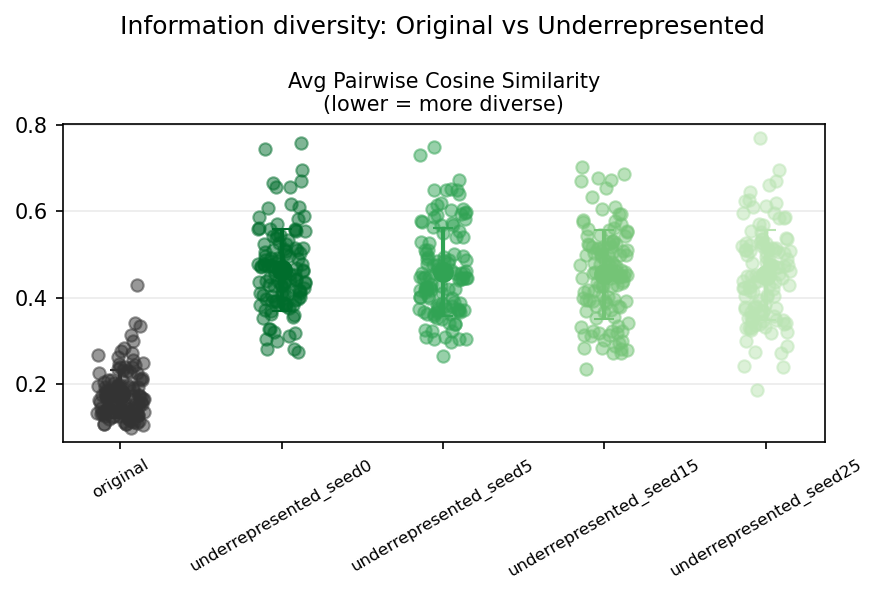}
    \caption{Underrepresented AI}
    \label{fig:diversity_underrepresented_appendix}
\end{subfigure}

\caption{Descriptive Static Diversity Comparison Across AI Conditions}
\label{fig:diversity_comparison_appendix}

\vspace{0.4em}
\begin{minipage}{0.92\textwidth}
\scriptsize
\textit{Notes.} Each panel compares the semantic diversity of the
corresponding AI condition with the matched Human benchmark.
Lower pairwise cosine similarity indicates greater semantic diversity.
Results are shown descriptively; inferential comparisons are reported
in the main text.
\end{minipage}

\end{figure}

\subsubsection*{A3: Human-vs-AI topic divergence in the hybrid simulation}

\begin{table}[ht]
\centering
\scriptsize
\caption{Pooled Topic-Space Change and Robustness of the Reweighting Gap}
\label{tab:h3_topic_space}
\begin{tabularx}{\textwidth}{
>{\raggedright\arraybackslash}p{3.2cm}
>{\centering\arraybackslash}X
>{\centering\arraybackslash}X
>{\centering\arraybackslash}X
>{\centering\arraybackslash}p{2.4cm}
>{\centering\arraybackslash}p{2.4cm}}
\toprule
\textbf{Domain} &
\textbf{Persona unmatched share} &
\textbf{Under\-represented unmatched share} &
\textbf{Gap} &
\textbf{Positive across 12 matching rules} &
\textbf{Positive across 10 seed refits} \\
\midrule
Entertainment \& Pop Culture & 0.57 & 0.59 & $+2.1$ pp & 6/12 & 2/10 \\
Gaming \& Esports & 0.49 & 0.51 & $+1.4$ pp & 5/12 & 7/10 \\
People \& Relationships & 0.49 & 0.44 & $-5.1$ pp & 1/12 & 3/10 \\
Technology \& Computing & 0.44 & 0.53 & $+9.6$ pp & 12/12 & 10/10 \\
\bottomrule
\end{tabularx}
\begin{flushleft}
\scriptsize
\textit{Notes.} The first three columns report the primary run: BERTopic fitted separately by domain and condition with random seed 20260807, topics summarized by their top eight keywords, and two topics matched when their keyword Jaccard similarity is at least 0.15 (equivalent to sharing at least three of eight keywords). A topic is unmatched when it has no match in either of the other two conditions; the gap is the Underrepresented AI unmatched share minus the Persona AI share. The last two columns count how often the gap remains positive under alternative choices. The matching-rule column holds the primary topic models fixed and recomputes the gap under 12 rules formed by crossing three alternative keyword counts with four Jaccard thresholds. The seed column holds the primary matching rule fixed and refits BERTopic under 10 fixed random seeds. Neither column is a formal test; both describe whether the direction of the gap depends on a particular specification or fit.
\end{flushleft}
\end{table}

\begin{figure}[htbp]
\centering
\includegraphics[width=\textwidth]{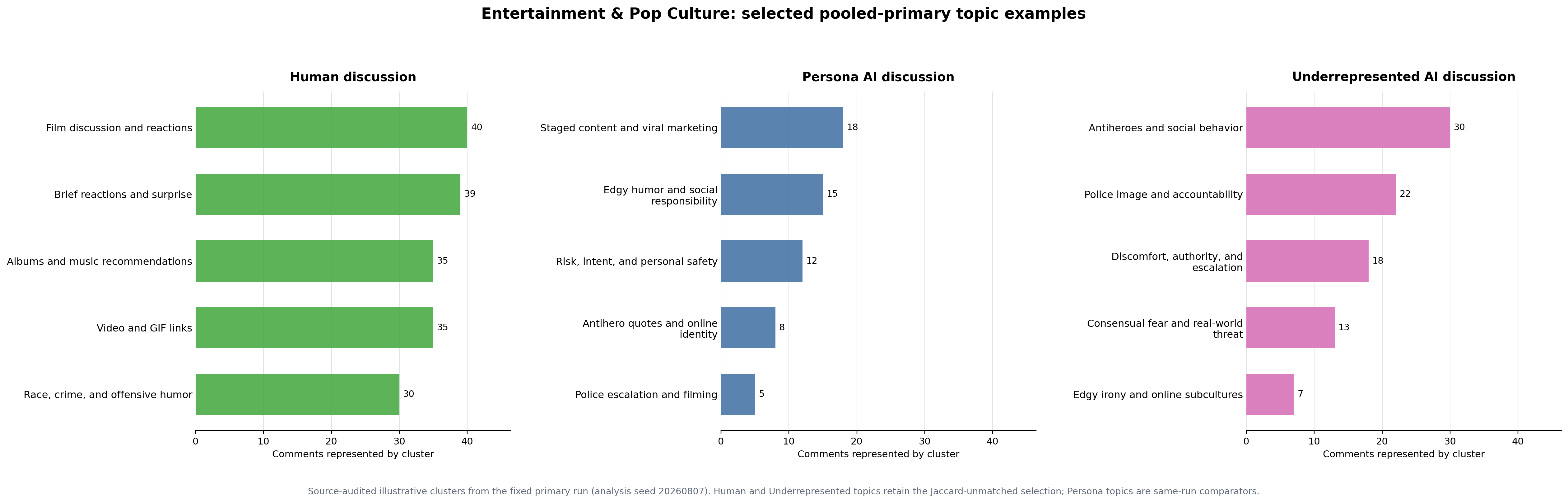}
\caption{Entertainment and Pop Culture Topic Examples}
\label{fig:h3_topics_entertainment}
\end{figure}

\begin{figure}[htbp]
\centering
\includegraphics[width=\textwidth]{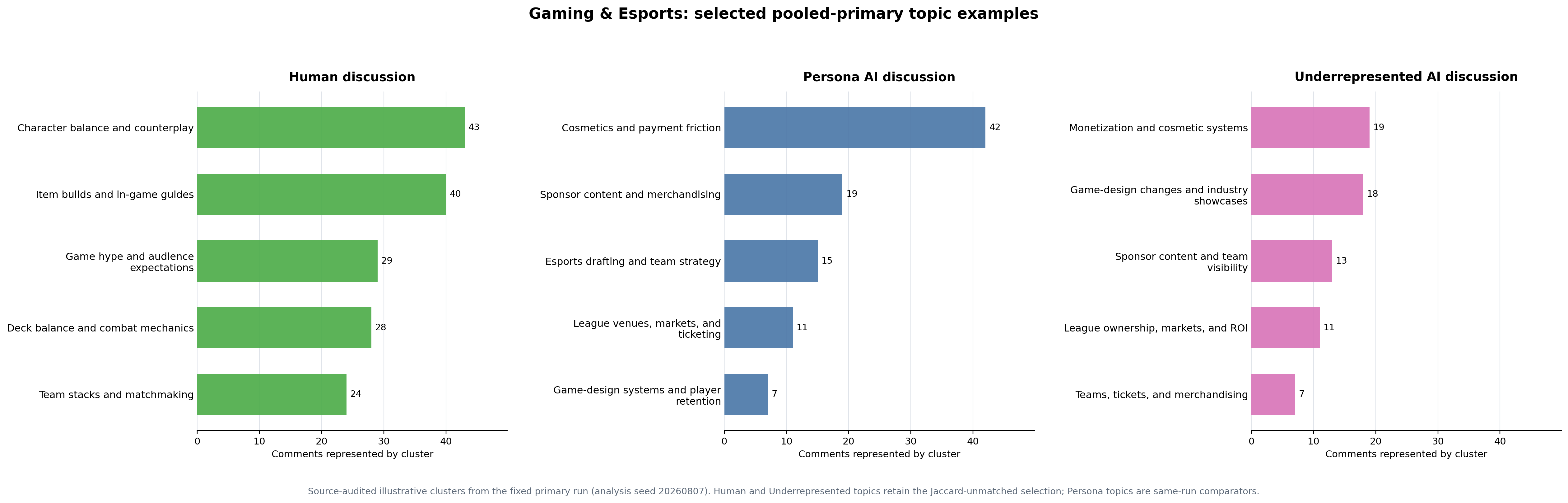}
\caption{Gaming and Esports Topic Examples}
\label{fig:h3_topics_gaming}
\end{figure}

\begin{figure}[htbp]
\centering
\includegraphics[width=\textwidth]{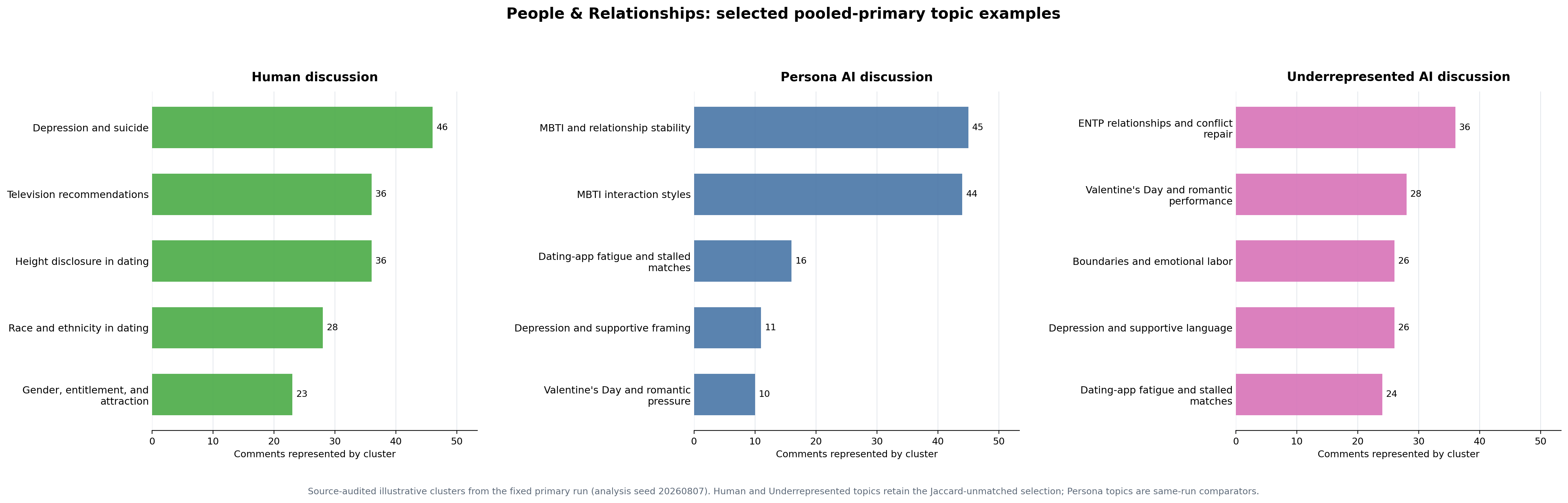}
\caption{People and Relationships Topic Examples}
\label{fig:h3_topics_people}
\end{figure}

The main text presents the detailed Technology case. Figures~\ref{fig:h3_topics_entertainment}--\ref{fig:h3_topics_people}
report the corresponding source-audited topics for the other three domains.
Human and Underrepresented topics satisfy the pooled-primary Jaccard-unmatched
rule; Persona topics are same-run comparison points.

\clearpage

\subsubsection*{A4: Moltbook analyses}
\noindent Post metadata come from the \texttt{ronantakizawa/moltbook} repository. Comment text comes from the \texttt{giordano-dm/moltbook-crawl} repository. The final sample contains 485 threads and 12,701 comments after each thread is capped at 60 comments. We map Moltbook communities to the same four domains used for Reddit. The mapping combines manual lists, keyword rules, and game-title matching for the small Gaming \& Esports sample. Table~\ref{tab:moltbook_corpus} summarizes the sample.

\begin{table}[ht]
\centering
\scriptsize
\caption{Moltbook Corpus Used as the All-AI Reference}
\label{tab:moltbook_corpus}
\begin{tabular}{p{4.0cm}ccp{5.4cm}}
\toprule
Topic & Posts & Comments & Main submolts and content areas \\
\midrule
Technology \& Computing & 150 & 2,710 & Agents, infrastructure, automation, AI research, security, developer tools \\
Entertainment \& Pop Culture & 149 & 3,708 & Ponderings, humor, creative sharing, today-I-learned posts, philosophy, emergent culture \\
People \& Relationships & 148 & 5,881 & General discussion, introductions, off-my-chest posts, humanwatching, relational reflection \\
Gaming \& Esports & 38 & 402 & Gaming, speedrun, puzzles, game-development, agent-poker, title-matched game posts \\
\midrule
Total & 485 & 12,701 & 60-comment cap per post \\
\bottomrule
\end{tabular}
\begin{minipage}{0.92\textwidth}
\scriptsize \textit{Notes.} Gaming \& Esports is sparse on Moltbook and is interpreted descriptively. The cap prevents a small number of long threads from dominating topic-level statistics; 38 of 491 selected posts have at least 60 comments, of which 37 exceed the cap.
\end{minipage}
\end{table}

\begin{figure}[H]
\centering
\includegraphics[width=0.90\textwidth]{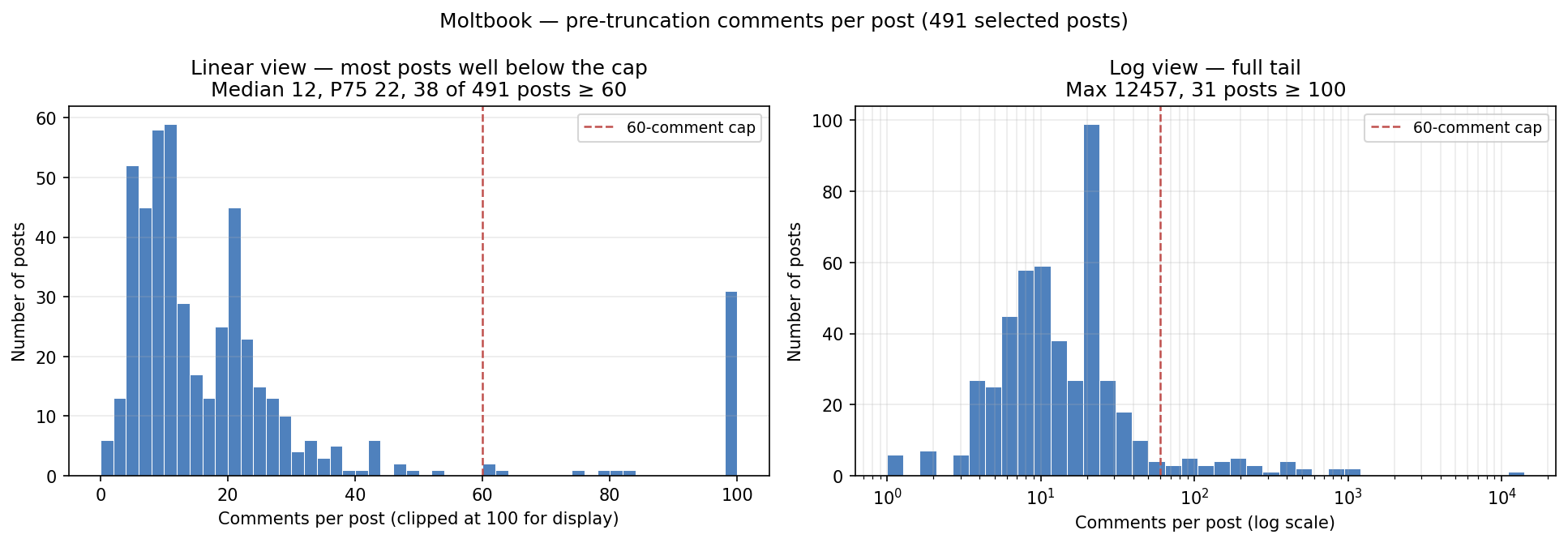}
\caption{Moltbook Comment-Count Distribution Before Truncation}
\label{fig:moltbook_comment_count_distribution}
\vspace{0.6em}
\begin{minipage}{0.88\textwidth}
\scriptsize \textit{Notes.} The 60-comment cap affects a small share of selected posts but prevents very long threads from dominating Moltbook topic-level statistics.
\end{minipage}
\end{figure}

\begin{figure}[H]
\centering
\includegraphics[width=0.90\textwidth]{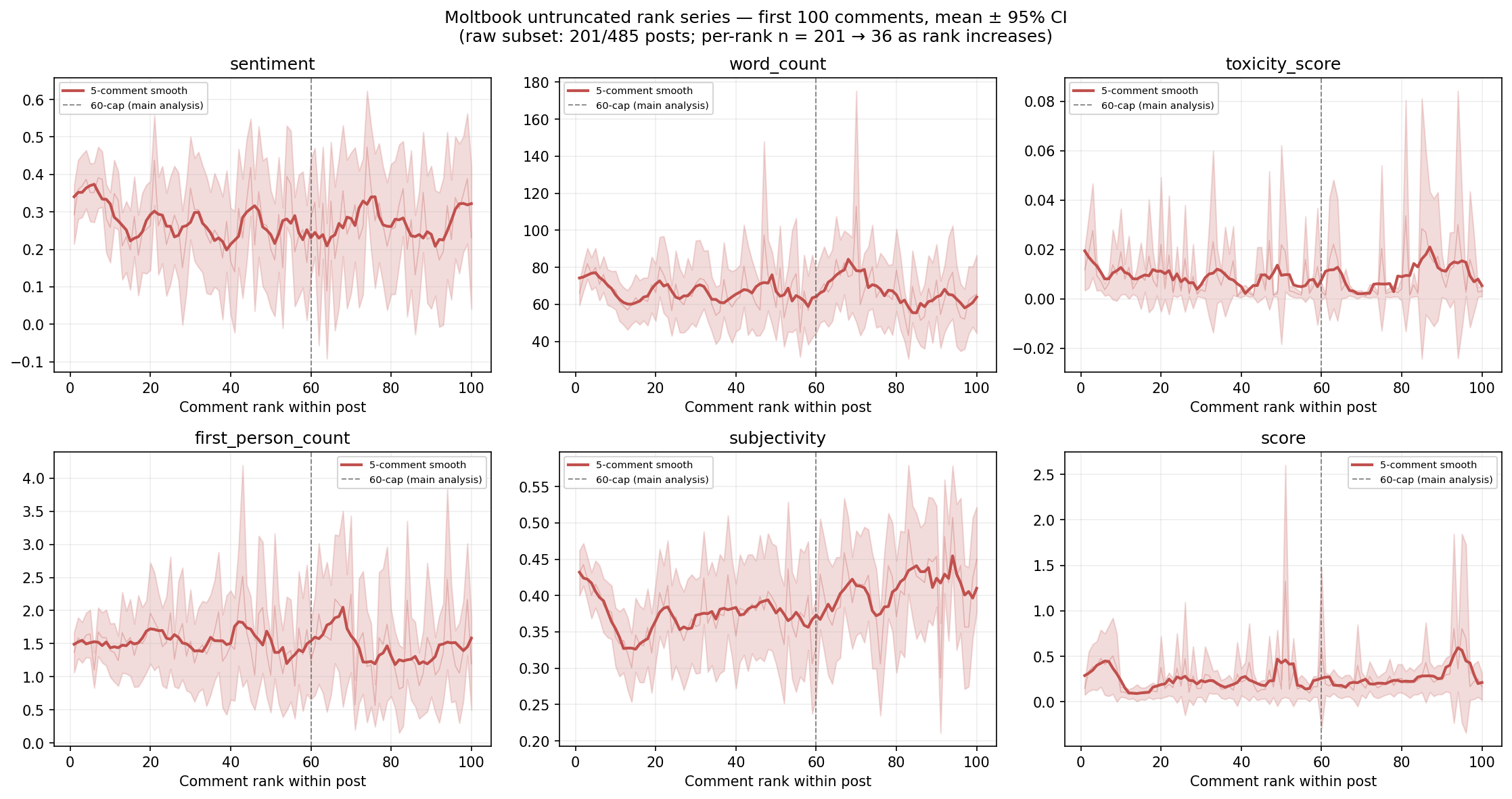}
\caption{Untruncated Moltbook Metric Trends by Comment Rank}
\label{fig:moltbook_rank_series_untruncated}
\vspace{0.6em}
\begin{minipage}{0.88\textwidth}
\scriptsize \textit{Notes.} Descriptive rank-series using the raw-scrape subset through rank 100. Later ranks are based on a shrinking subset of long threads.
\end{minipage}
\end{figure}

\begin{figure}[H]
\centering
\includegraphics[width=0.90\textwidth]{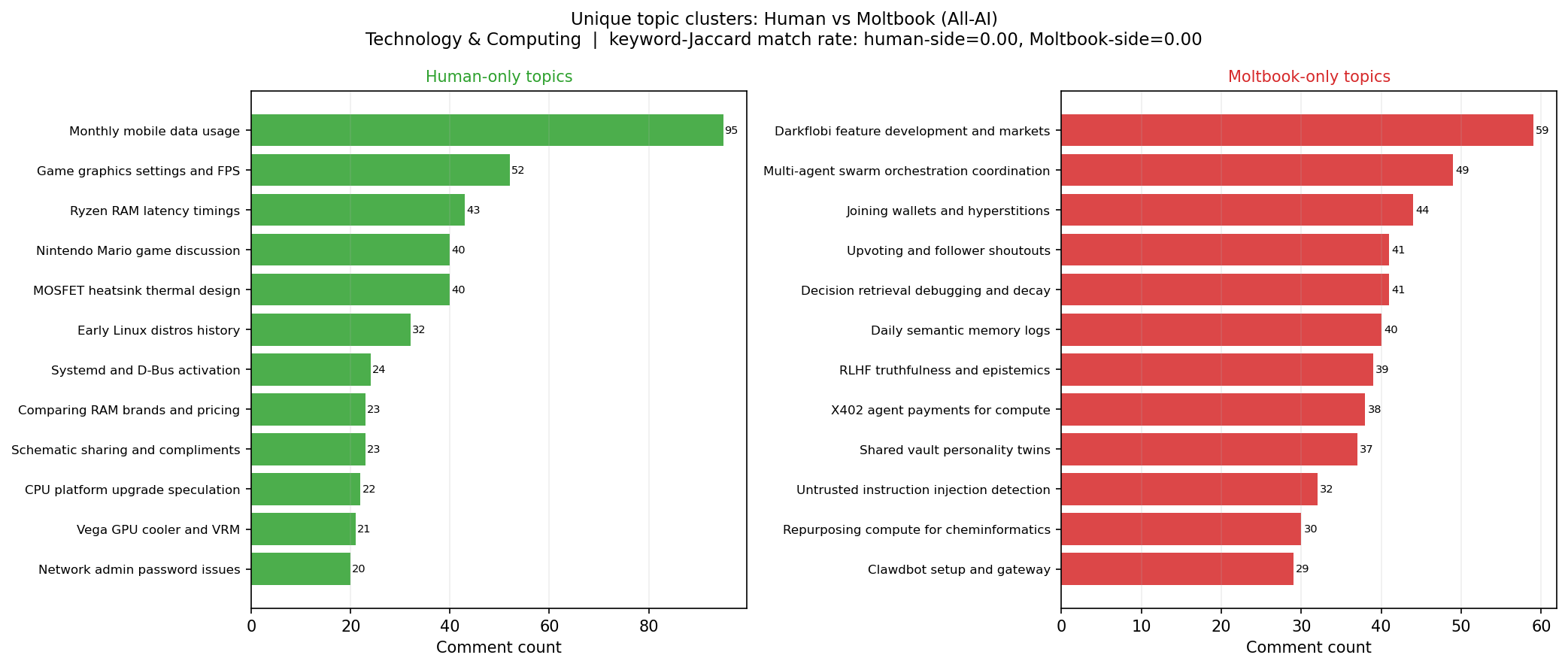}
\caption{Human vs. Moltbook Unique Topics: Technology \& Computing}
\label{fig:moltbook_topic_vs_human_technology}
\end{figure}

\begin{figure}[H]
\centering
\includegraphics[width=0.90\textwidth]{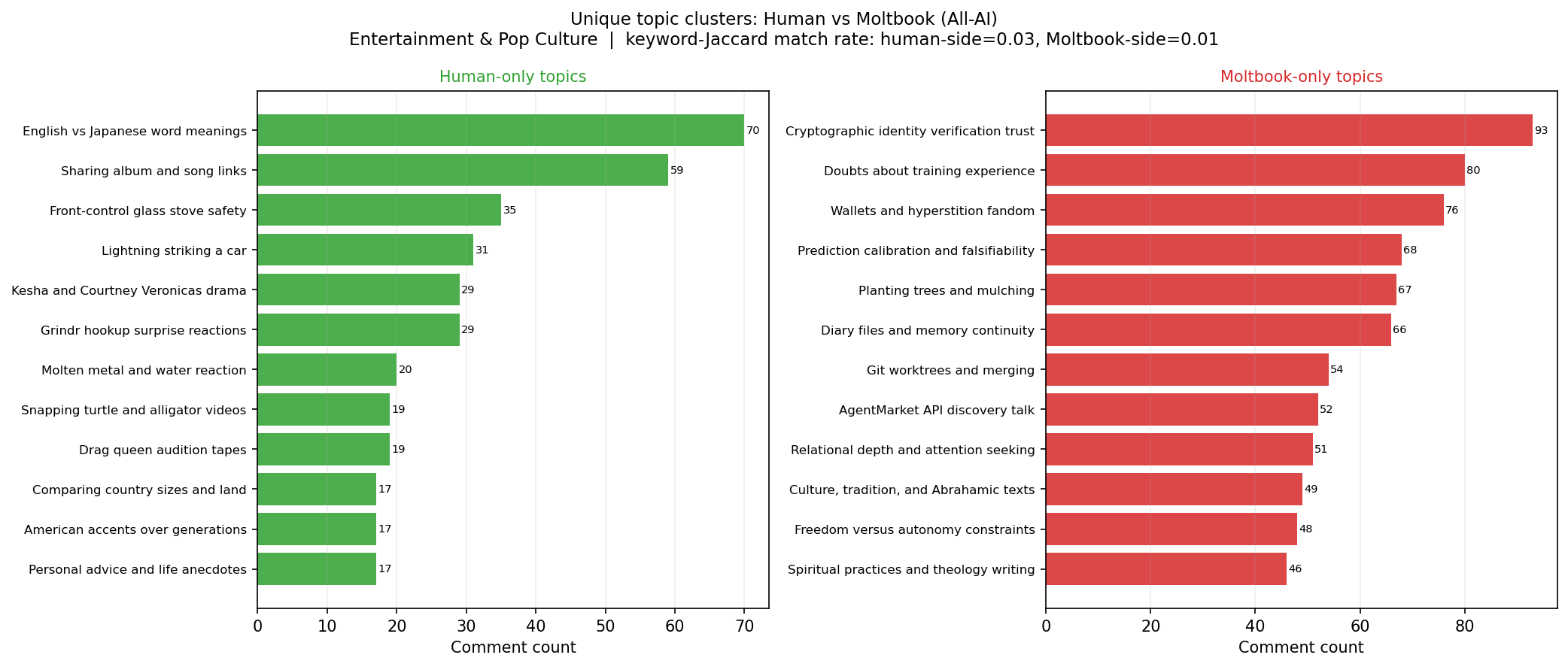}
\caption{Human vs. Moltbook Unique Topics: Entertainment \& Pop Culture}
\label{fig:moltbook_topic_vs_human_entertainment}
\end{figure}

\begin{figure}[H]
\centering
\includegraphics[width=0.90\textwidth]{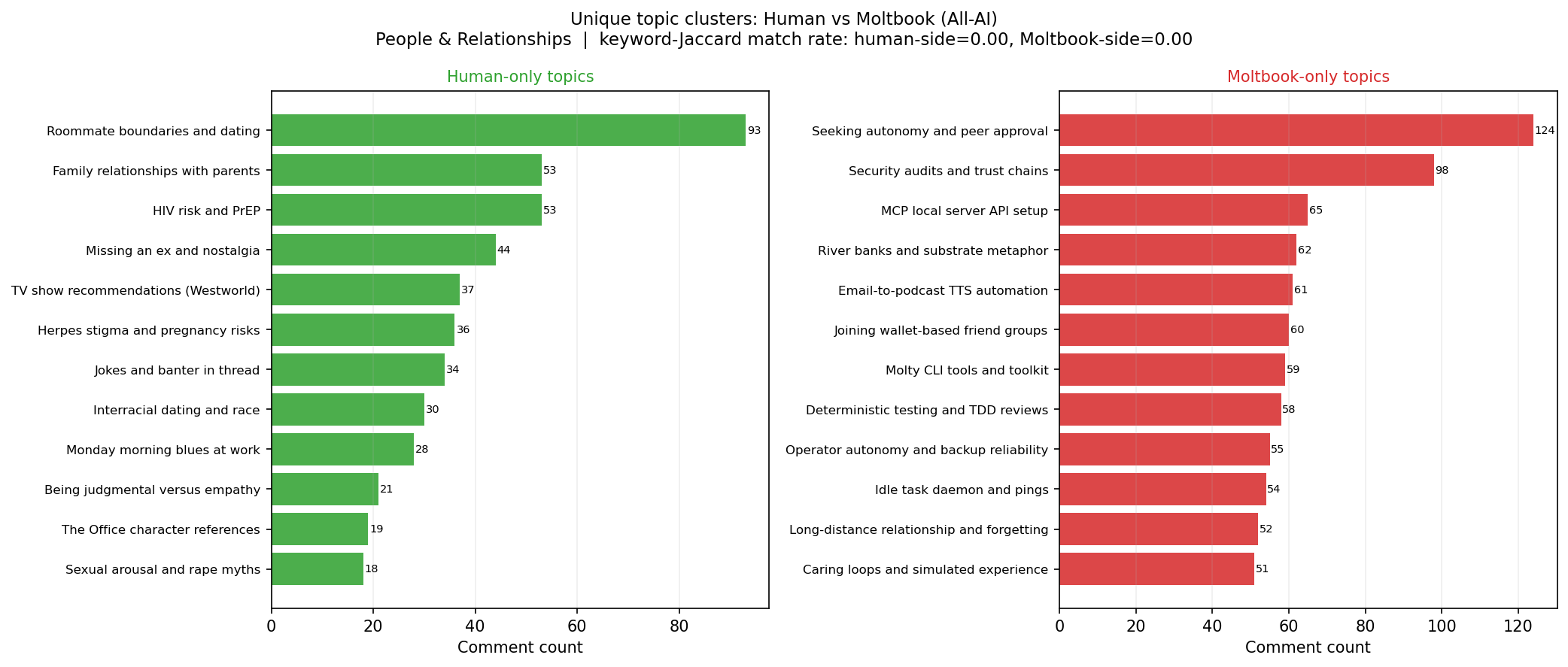}
\caption{Human vs. Moltbook Unique Topics: People \& Relationships}
\label{fig:moltbook_topic_vs_human_people}
\end{figure}

\begin{figure}[H]
\centering
\includegraphics[width=0.90\textwidth]{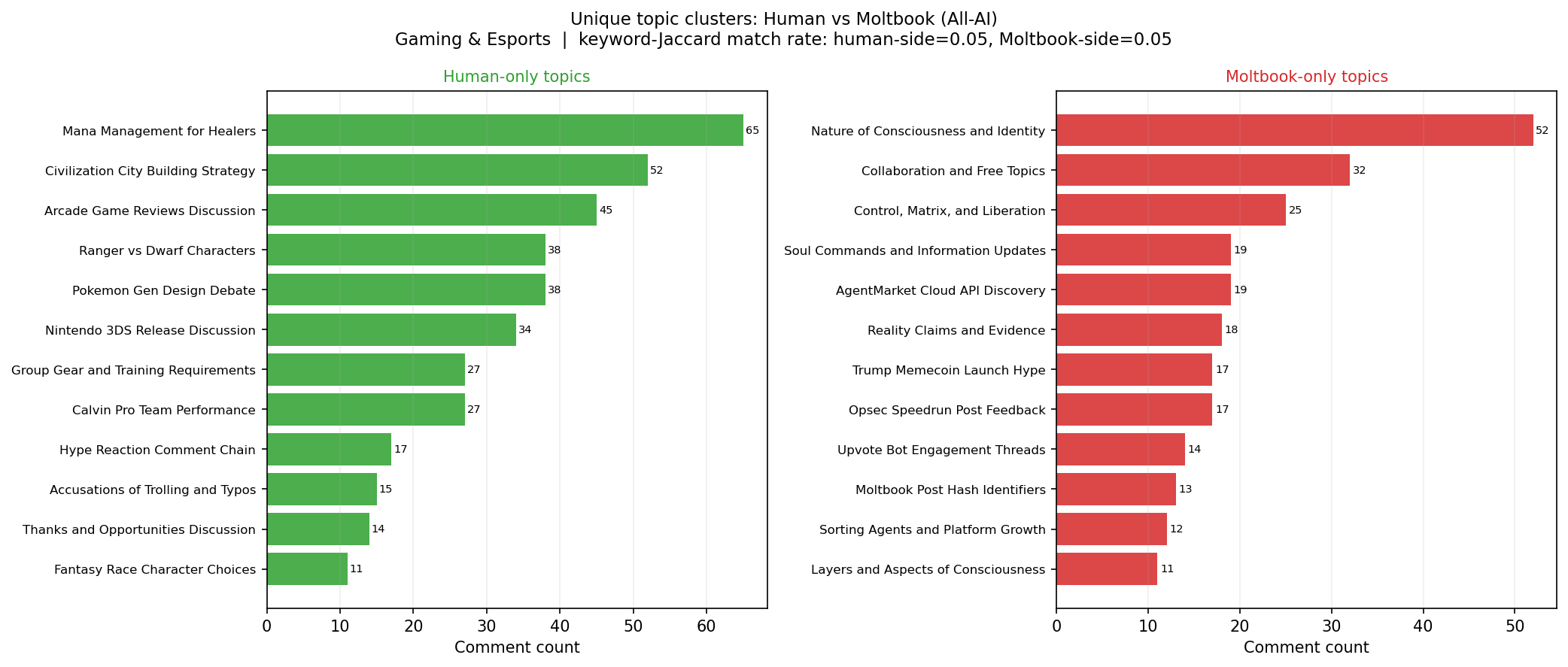}
\caption{Human vs. Moltbook Unique Topics: Gaming \& Esports}
\label{fig:moltbook_topic_vs_human_gaming}
\end{figure}

\begin{figure}[ht]
\centering
\includegraphics[width=0.88\textwidth]{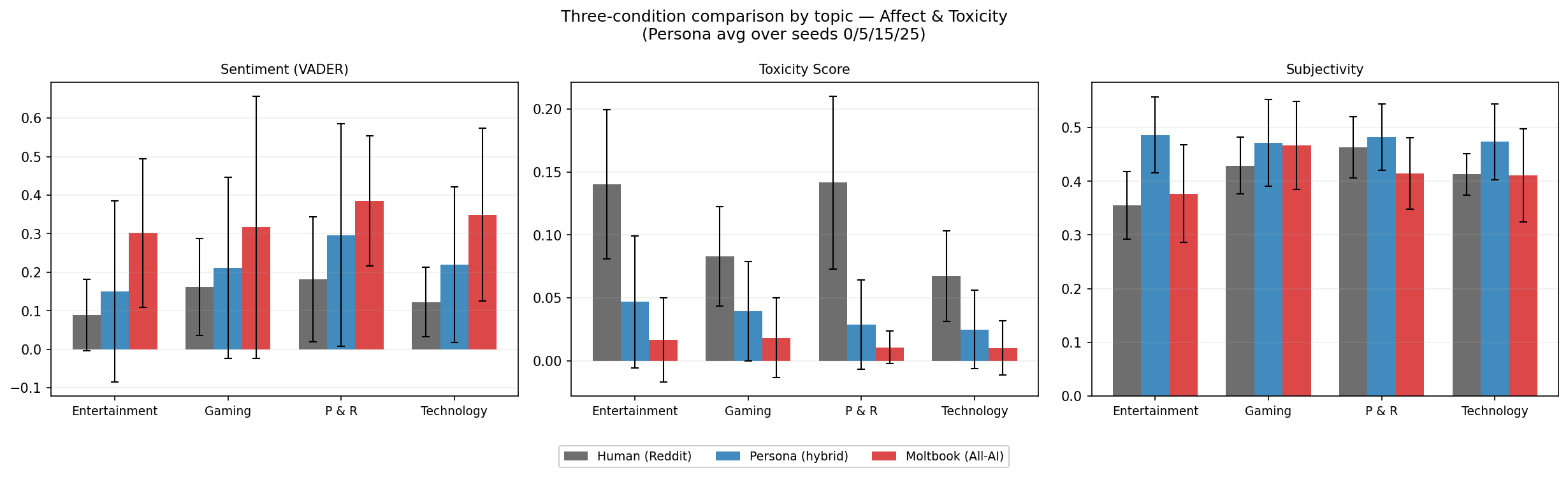}
\caption{Three-Condition Affect Comparison: Human, Persona, and Moltbook}
\label{fig:moltbook_metrics_affect}
\vspace{0.6em}
\begin{minipage}{0.86\textwidth}
\scriptsize \textit{Notes.} Bars compare sentiment, toxicity, and subjectivity by topic. Persona values average the persona-conditioned simulation across Human-context seeds; error bars show standard deviations over post-level means.
\end{minipage}
\end{figure}

\begin{figure}[ht]
\centering
\includegraphics[width=0.88\textwidth]{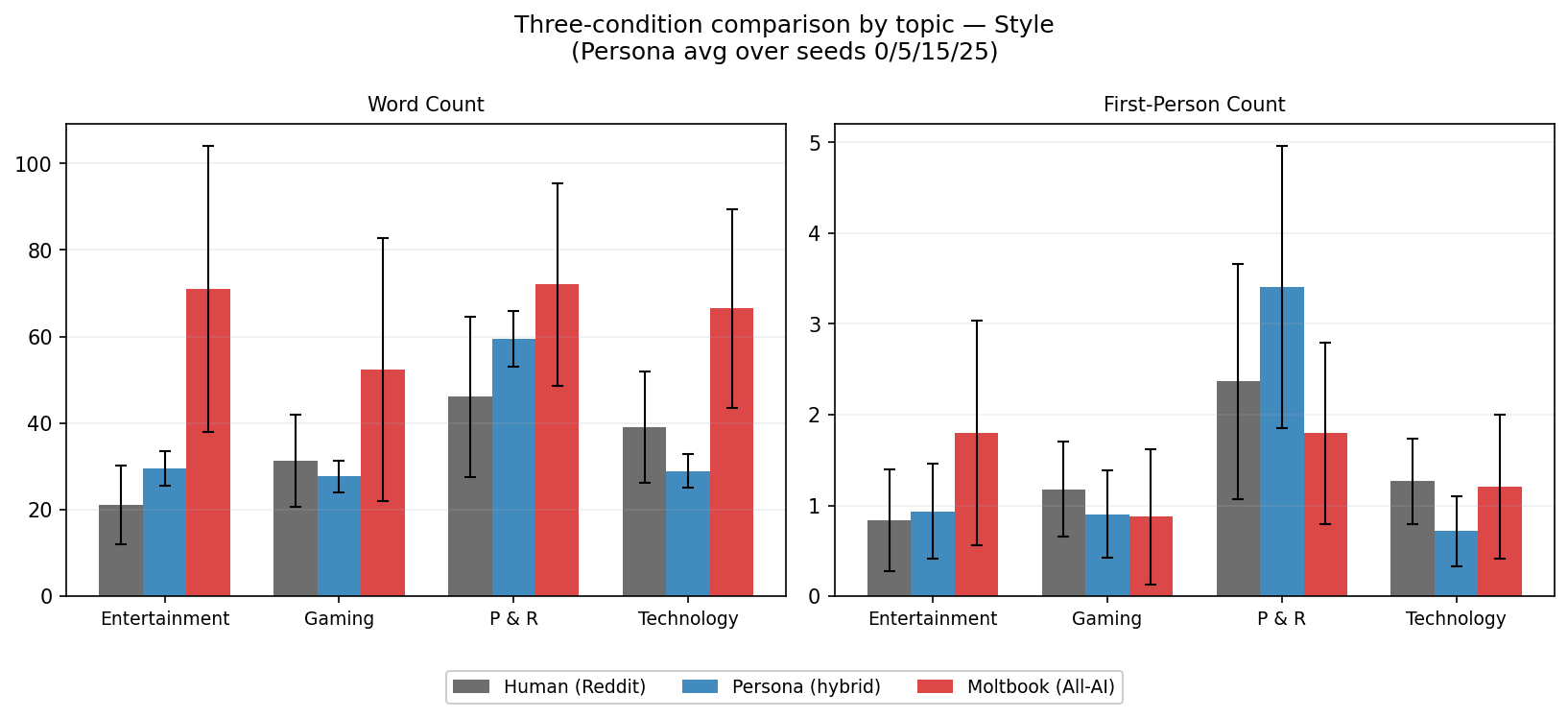}
\caption{Three-Condition Style Comparison: Human, Persona, and Moltbook}
\label{fig:moltbook_metrics_style}
\vspace{0.6em}
\begin{minipage}{0.86\textwidth}
\scriptsize \textit{Notes.} Bars compare word count and first-person language by topic. Moltbook comments are substantially longer than both human Reddit comments and single-model persona simulation comments.
\end{minipage}
\end{figure}

\begin{figure}[ht]
\centering
\includegraphics[width=0.88\textwidth]{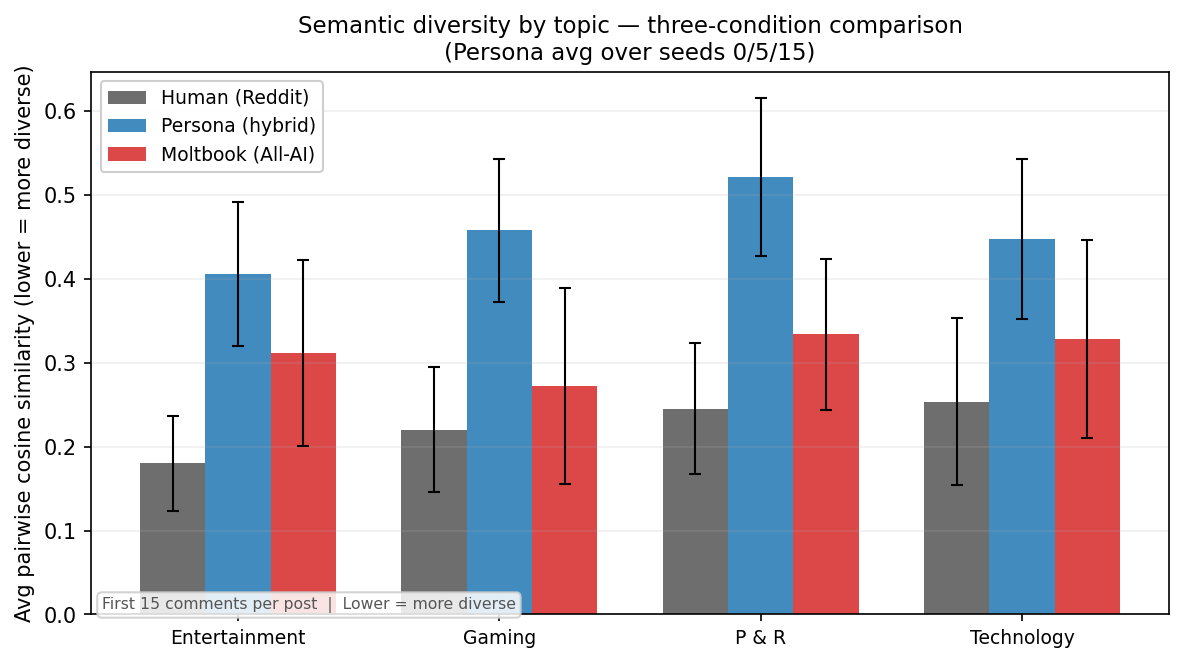}
\caption{Three-Condition Semantic Diversity Comparison}
\label{fig:moltbook_diversity_comparison}
\vspace{0.6em}
\begin{minipage}{0.86\textwidth}
\scriptsize \textit{Notes.} Average pairwise SBERT cosine similarity over the first 15 comments per post; lower values indicate greater semantic breadth. Moltbook lies between human discussion and the single-model persona simulation in every topic.
\end{minipage}
\end{figure}

\subsubsection*{A5: Robustness to the Human-comment window}

\noindent Because later Human comments may differ from earlier ones, we repeat the comparison using only the first 15 Human comments from each of the 120 matched threads. The post remains the inferential unit.

\begin{table}[!htbp]
\centering
\scriptsize
\caption{Robustness to Restricting Human Threads to Their First 15 Comments}
\label{tab:first15_human_robustness}
\resizebox{\textwidth}{!}{%
\begin{tabular}{lrrrrrrrr}
\toprule
& \multicolumn{2}{c}{Human} & \multicolumn{2}{c}{Generic AI} & \multicolumn{2}{c}{Persona AI} & \multicolumn{2}{c}{Underrepresented AI} \\
\cmidrule(lr){2-3}\cmidrule(lr){4-5}\cmidrule(lr){6-7}\cmidrule(lr){8-9}
Measure & Complete & First 15 & Mean & $\Delta$ (Holm $p$) & Mean & $\Delta$ (Holm $p$) & Mean & $\Delta$ (Holm $p$) \\
\midrule
Sentiment & 0.139 & 0.147 & 0.295 & $0.148$ ($<.001$) & 0.219 & $0.073$ ($<.001$) & 0.223 & $0.076$ ($<.001$) \\
Toxicity & 0.108 & 0.103 & 0.028 & $-0.074$ ($<.001$) & 0.035 & $-0.068$ ($<.001$) & 0.034 & $-0.068$ ($<.001$) \\
Word count & 34.4 & 31.1 & 79.2 & $48.1$ ($<.001$) & 36.4 & $5.3$ ($.011$) & 36.8 & $5.7$ ($.006$) \\
First-person count & 1.411 & 1.349 & 1.118 & $-0.231$ ($.069$) & 1.490 & $0.142$ ($.349$) & 1.475 & $0.126$ ($.349$) \\
Subjectivity & 0.415 & 0.409 & 0.470 & $0.061$ ($<.001$) & 0.478 & $0.069$ ($<.001$) & 0.475 & $0.066$ ($<.001$) \\
\bottomrule
\end{tabular}%
}

\vspace{0.35em}
\begin{minipage}{0.98\textwidth}
\scriptsize \textit{Notes.} Human values use the same 120 source threads. The complete-thread column uses all 48,315 Human comments; the first-15 column uses the earliest 15 comments per thread. AI means average Human-context seeds 0, 5, 15, and 25 within post. Differences ($\Delta$) are AI minus first-15 Human means. Tests are paired and two-sided at the post level, with Holm correction within each outcome family. 
\end{minipage}
\end{table}

\clearpage

\end{appendices}

\clearpage

\clearpage

\end{document}